\documentclass{article}

\usepackage{microtype}
\usepackage{graphicx}
\usepackage{subcaption}
\usepackage{booktabs} 

\usepackage{hyperref}

\usepackage[preprint]{icml2026}

\usepackage{amsmath}
\usepackage{amssymb}
\usepackage{mathtools}
\usepackage{amsthm}

\usepackage[capitalize,noabbrev]{cleveref}
\usepackage{multirow}
\theoremstyle{plain}

\theoremstyle{definition}

\theoremstyle{remark}

\usepackage[textsize=tiny]{todonotes}

\icmltitlerunning{Reason and Restore: Improving Universal Image Restoration with Chain-of-Thought Reasoning Framework}

\begin{document}

\twocolumn[
  \icmltitle{Reason and Restore: Improving Universal Image Restoration with Chain-of-Thought Reasoning Framework}



  \icmlsetsymbol{equal}{*}

  \begin{icmlauthorlist}
    \icmlauthor{Wending Yan}{swjtu}
    \icmlauthor{Rongkai Zhang}{ntu}
    \icmlauthor{Kaihua Tang}{tongji}
    \icmlauthor{Yu Cheng}{nus}
    \icmlauthor{Qiankun Liu}{comp}
  \end{icmlauthorlist}

  \icmlaffiliation{swjtu}{Southwest Jiaotong University, Chengdu, China}
  \icmlaffiliation{ntu}{Nanyang Technological University, Singapore}
  \icmlaffiliation{tongji}{Tongji University, Shanghai, China}
  \icmlaffiliation{nus}{National University of Singapore, Singapore}
  \icmlaffiliation{comp}{Tsinghua University}

  \icmlcorrespondingauthor{Rongkai Zhang}{rongkai002@e.ntu.edu.sg}

  \icmlkeywords{Machine Learning, ICML}

  \vskip 0.3in
]



\printAffiliationsAndNotice{}  


\begin{abstract}
Universal image restoration (UIR) aims to recover clean images from diverse and unknown degradations using a unified model. 
Existing UIR methods primarily focus on pixel reconstruction and often lack explicit diagnostic reasoning over degradation composition, severity, and scene semantics prior to restoration.
We propose Reason and Restore (R\&R), a novel framework that integrates structured Chain-of-Thought (CoT) reasoning into the image restoration pipeline. 
R\&R introduces an explicit reasoner, implemented by fine-tuning Qwen3-VL, to diagnose degradation types, quantify degradation severity, infer key degradation-related factors, and describe relevant scene and object semantics. 
The resulting structured reasoning provides interpretable and fine-grained diagnostic priors for the restorer. 
To further improve restoration quality, the quantified degradation severity produced by the reasoner is leveraged as reinforcement learning (RL) signals to guide and strengthen the restorer.
Unlike existing multimodal LLM-based agentic systems that decouple reasoning from low-level vision tasks, R\&R tightly couples semantic diagnostic reasoning with pixel-level restoration in a unified framework. 
Extensive experiments across diverse UIR benchmarks demonstrate that R\&R achieves state-of-the-art performance while offering unique interpretability into the restoration process.
\end{abstract}
\section{Introduction}
\label{sec:intro}


Image restoration~\cite{jiang2025survey, liang2021swinir} is a fundamental problem in computer vision and plays a critical role in safety-critical applications~\cite{zhang2023perception, freeman2021aerial, yan2020optical, galshetwar2025clear} such as autonomous driving, construction robotics, aerial inspection, and outdoor surveillance, where visual perception often operates under adverse environmental conditions. Early image restoration frameworks primarily relied on handcrafted priors and task-specific optimization pipelines~\cite{zhang2021plug, schmidt2014shrinkage}, while recent advances have been driven by the emergence of deep generative models~\cite{yan2020nighttime,luo2025taming, luo2023refusion}, \textit{e.g.}, diffusion models that have significantly improved restoration quality across a wide range of degradation types. 

\begin{figure}[t]
    \centering
    \includegraphics[width=\columnwidth]{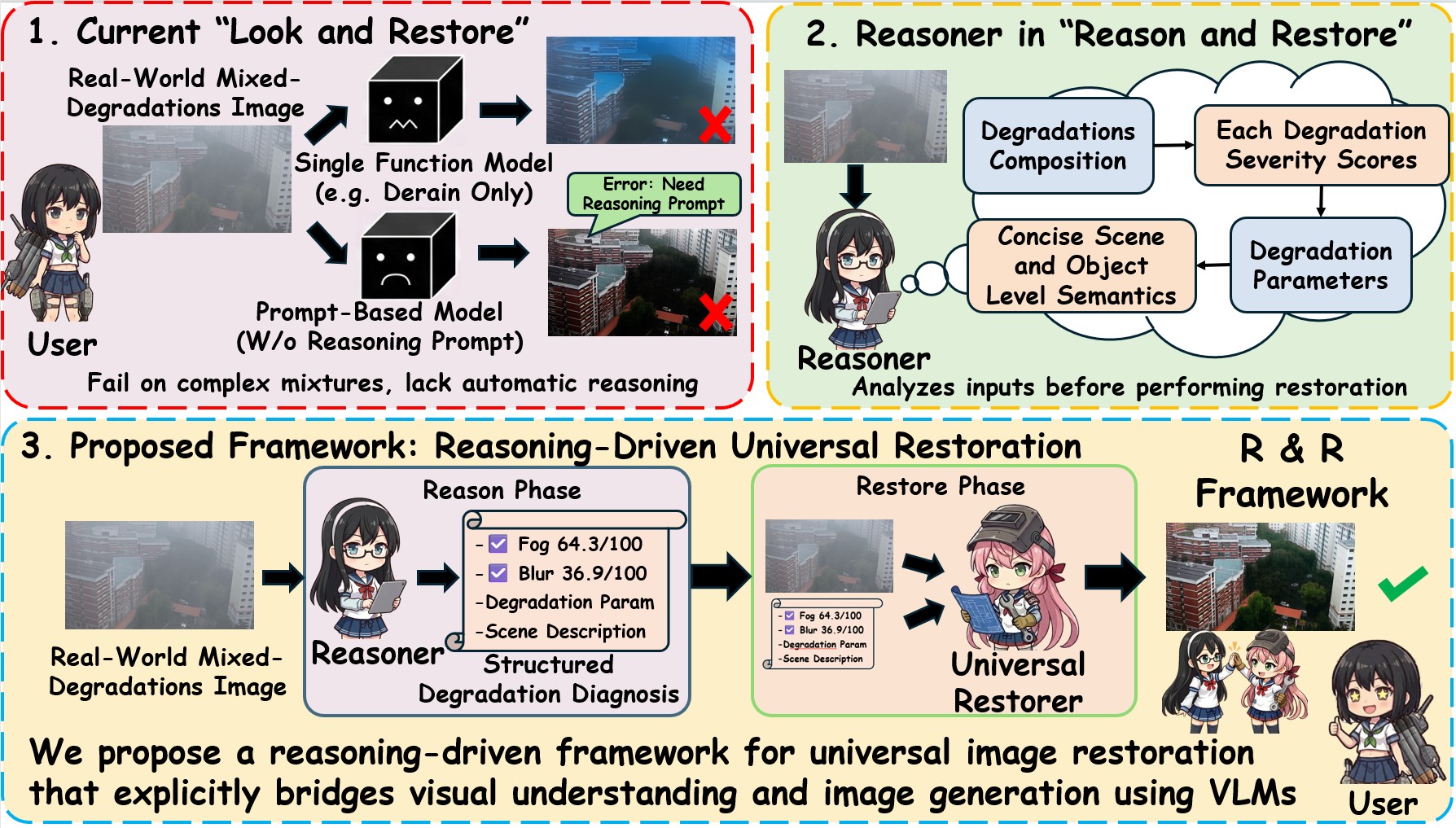}
    \caption{The proposed Reason and Restore (R\&R) framework first performs structured diagnostic reasoning to analyze degradation composition, severity, and parameters, and then guides universal image restoration.}
    \label{fig:framework}
\end{figure}

Despite these advances, most existing approaches still follow a direct 
\textbf{Look-and-Restore} paradigm: they either employ single-function models specialized for a specific degradation~\cite{wang2020model, gui2023comprehensive, yan2022feature} (deraining, dehazing, or denoising), or require users to manually specify prompts, degradation types, or hyper-parameters at inference time~\cite{vaishnav2023promptir, conde2024instructir}. However, real-world visual environments are inherently complex and rarely exhibit a single isolated degradation, \textit{e.g.}, rain and fog frequently co-occur~\cite{yan2021self}, motion blur is often coupled with low-light noise, besides, atmospheric scattering, sensor noise, and camera shake may jointly affect image quality. We argue that a principled mechanism for automatically analyzing the degradation composition of an input image is essential for robust restoration in realistic scenarios, and can substantially improve the adaptability and reliability of modern restoration systems.



Inspired by recent advances in reasoning-enabled Large Language Models (LLMs)~\cite{floridi2020gpt}, we posit that introducing explicit reasoning and analysis into the image restoration pipeline can lead to more reliable and effective restoration performance. In LLMs, allowing intermediate reasoning steps before producing a final answer, \textit{i.e.}, the Chain-of-Thought framework~\cite{wei2022chain}, induces a test-time scaling law, under which the model’s reasoning performance improves with increased computational effort (tokens) allocated to the intermediate thinking stage. Analogously, we argue that image restoration should follow a \textbf{Reason-and-Restore (R\&R)} principle, where the model first analyzes the input image to infer the underlying degradation factors and their severity before performing restoration. Figure~\ref{fig:framework} demonstrates the main difference between \textbf{Look-and-Restore } and \textbf{Reason-and-Restore}.

Early works such as RestoreAgent \cite{he2024restoreagent} and Q-Agent \cite{qagent2025} only employ Chain-of-Thought as a mechanism for tool invocation, without integrating the intermediate reasoning process and its generated outputs into the downstream pixel-level restoration models. In contrast, this paper is the first to propose a unified R\&R framework. Specifically, R\&R framework formulates universal image restoration as a two-stage process consisting of a structured diagnostic \emph{Reason} phase and a pixel-level \emph{Restore} phase. The Reason phase infers degradation types, severities, parameters, and scene semantics, which are then encoded as diagnostic priors to guide restoration and further optimized via reinforcement learning with severity-based rewards, enabling degradation-aware and scene-adaptive restoration.

We evaluate our method on both synthetic benchmarks and a newly collected real-world dataset to assess restoration performance and generalization. Experiments are conducted on the OTS~\cite{li2018benchmarking} and RESIDE~\cite{ zhao2020dehazing} datasets under standard protocols, using PSNR and SSIM as evaluation metrics, and on a real-world outdoor dataset comprising over 700 images from 15 representative scenes. Compared with conventional restoration pipelines and recent diffusion- and foundation-model-based baselines, our method consistently achieves the best quantitative results on both synthetic benchmarks and produces more faithful structures and colors in qualitative comparisons. Ablation studies further demonstrate that structured diagnostics, semantic guidance, and reinforcement learning jointly contribute to the observed performance gains.


Our contributions are threefold:
\textbf{1) Degradation reasoning via Chain-of-Thought learning.}
We introduce a \emph{Reason} phase that leverages the strong visual understanding prior of large vision-Language Models  (VLMs), \textit{e.g.}, Qwen3-VL~\cite{Qwen3-VL} to explicitly reason about image degradations.
Using a data generation pipeline grounded in realistic degradation models, we construct large-scale training data and perform supervised fine-tuning to enable the VLM to produce structured Chain-of-Thought reasoning, including degradation types, severity, perceptual quality, and scene context.
\emph{This goes beyond prompt engineering by training the model to explicitly reason about degradations rather than implicitly conditioning on text.}
\textbf{2) Degradation reasoning guidance for VLM-based restoration.}
We introduce a \emph{Restore}  phase by retraining a VLM-based image restorer (\textit{e.g.}, Qwen-Image-Edit~\cite{wu2025qwen}) and, for the first time, explicitly feed the reasoning outputs from the think stage into the restoration model as structured priors.
Unlike prior language-guided or agentic restoration methods, our approach decouples reasoning from generation and does not rely on action selection or heuristic control policies.
\emph{This formulation enables universal restoration under unknown and mixed degradations.}
\textbf{3) Reinforcement learning alignment with diagnostic rewards.}
To align reasoning and restoration outcomes, we further fine-tune the model using reinforcement learning, where scores produced by the \emph{Reason} phase serve as reward signals.
This alignment encourages restorations that simultaneously improve quantitative metrics and conform to realistic degradation-aware priors.
\emph{As a result, the model achieves both perceptual fidelity and strong numerical performance.}


\section{Related Work}
\label{sec:related}

\begin{table*}[ht]
\centering
\small
\begin{tabular}{lcccc}
\toprule
Method & Language Input & Explicit Reasoning & Trainable Restoration & Universal IR \\
\midrule
Prompt-based IR & \checkmark & $\times$ & \checkmark & $\triangle$ \\
Agentic IR & \checkmark & $\triangle$ & $\times$ & $\triangle$ \\
\textbf{Ours} & \checkmark & \checkmark & \checkmark & \checkmark \\
\bottomrule
\end{tabular}
\caption{Comparison between prompt-based methods, agentic image restoration, and our method. $\checkmark$ indicates that the property is explicitly modeled as a core component, $\triangle$ indicates partial or implicit support, and $\times$ indicates that the property is not modeled.
Our approach uniquely leverages explicit Chain-of-Thought reasoning as structured priors for universal image restoration.}
\label{tab:comparison}
\end{table*}

\noindent\textbf{Universal Image Restoration.} 
Universal image restoration (UIR) aims to restore degraded images with diverse and unknown degradations within a single framework. Formally, given a degraded image $y$, UIR aims to estimate the clean latent image $x$ through a mapping $\mathcal{F}_{\theta}(y, c)$, where $c$ represents optional degradation cues. Early attempts primarily relied on deep Convolutional Neural Networks (CNNs) \cite{zhang2017beyond} and later moves toward Vision Transformers (ViTs) such as SwinIR \cite{liang2021swinir} and Restormer \cite{zamir2022restormer} to capture long-range dependencies. However, they often struggle with ``in-the-wild'' scenarios where multiple degradations, such as blur, noise, and haze are coupled in complex ways, leading to over-smoothed textures or failure in artifact removal. Diffusion Models (DMs) based methods like StableSR \cite{wang2024exploiting} ResShift \cite{yue2023resshift}, and FoundIR \cite{foundir2025} leverage the rich structural priors of pre-trained text-to-image models to hallucinate realistic details. These Prompt-based generative approaches are computationally expensive and prone to ``hallucination,'' where the model generates plausible but contextually incorrect details. 

The emergence of Multimodal Large Language Models (MLLMs) has inspired a new branch of ``agentic IR.'' Frameworks like RestoreAgent \cite{he2024restoreagent} and Q-Agent \cite{qagent2025} employ VLMs as high-level controllers to perceive image quality and sequence a set of external, specialized tools. While this ``dispatching'' mechanism allows for human-like task decomposition, it suffers from two major drawbacks. First, the restoration quality is strictly bounded by the performance of the external toolset. Second, the disconnect between the ``reasoner'' (VLM) and the ``restorer'' (IR models) prevents the model from utilizing high-level semantic insights during the actual pixel-level reconstruction process, resulting in a suboptimal information flow. Therefore, we propose R\&R to \emph{fully utilize the strong reasoning and editing ability of VLM for a better universal image restoration.} Table 1 summarizes the differences among our proposed R\&R and other counterparts.


\noindent\textbf{ Chain-of-Thought Reasoning in Vision-Language Models.} Chain-of-Thought (CoT) reasoning rooted on large language models has been extended to vision-language models (VLMs)~\cite{chen2024cot,zhang2025cot} to improve multimodal reasoning and interpretability. Recent VLM based IR approaches ~\cite{agenticIR,qagent2025} usually use CoT to decompose
multi-degradation perception into single-degradation, and restore images with pretrained restoration models selected by the VLM accordingly. Although such agentic systems demonstrate strong flexibility for complex restoration scenarios, we argue that the performance is largely limited by the pretrained restoration models, and reasoning should guide the restoration model instead of simply selecting. 
To this end, the CoT of reasoning outputs in R\&R are designed into interpretable and structured priors, which are then provided as inputs to a separate VLM-based restoration network. This design enables the reasoning process to be explicit, reusable, and directly consumable by the restorer, \emph{i.e.} Qwen-image-edit~\cite{wu2025qwen}.   

\section{Method}

\begin{figure*}[t]
    \centering
    \includegraphics[width=\textwidth]{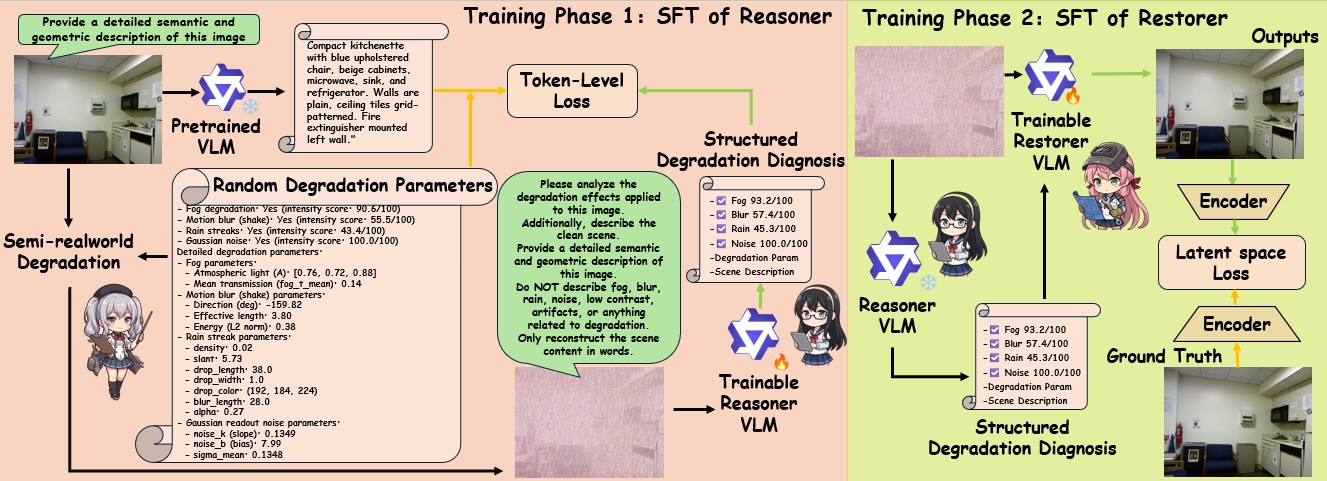}
    \caption{
    R\&R formulates universal image restoration as a two-stage process consisting of a \emph{Reason} phase and a \emph{Restore} phase, and is trained in three stages.
    In Training Phase~1, a VLM-based reasoner is supervised to perform structured degradation diagnosis under a semi-realworld degradation model.
    In Training Phase~2, a universal image restorer is fine-tuned with paired degraded and clean images, conditioned on the diagnostic outputs from the reasoner.
    }
    \label{fig:phase1}
\end{figure*}

\begin{figure}[t]
    \centering
\includegraphics[width=\columnwidth]{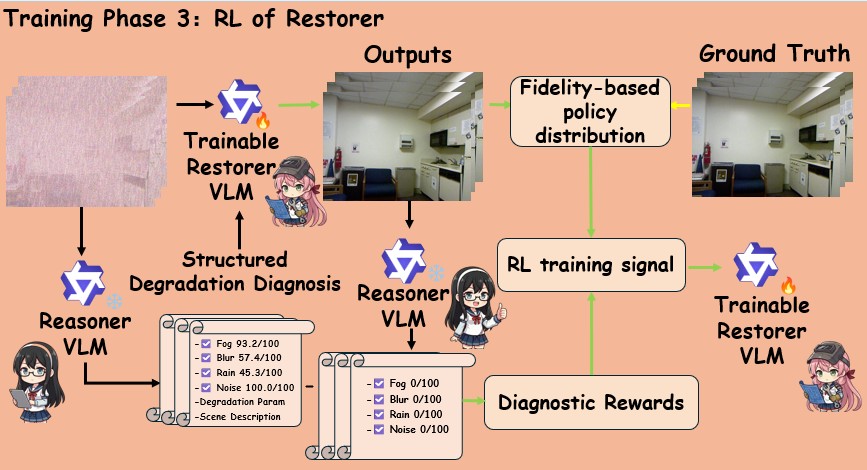}
    \caption{ In Training Phase~3, the restorer is further optimized via reinforcement learning using diagnostic rewards derived from severity reduction.
    }
    \label{fig:phase3}
\end{figure}

The proposed \textbf{Reason and Restore (R\&R)} framework formulates universal image restoration as a two-stage process which are a \emph{Reason} phase and a \emph{Restore} phase, as illustrated in Figure.~\ref{fig:framework}. We train the whole framework in three phases, which are supervised fine-tuning of the reasoner, supervised fine-tuning of the restorer, and reinforcement learning of the restorer, as illustrated in Figure~\ref{fig:phase1} and Figure~\ref{fig:phase3}.
We next describe the overall R\&R framework, followed by detailed discussions of the \emph{Reason} phase, the \emph{Restore} phase, and the reinforcement learning strategy.
Given a degraded input image $I_d$, in the \emph{Reason} phase, R\&R first performs a high-level analysis to understand it via VLM-based reasoner, \emph{i.e.}, Qwen3-VL in a CoT, such as \emph{what degradations are present, how severe they are, and under what scene context they occur}, before executing low-level restoration. 
%
%

In the \emph{Restore} phase, a VLM based universal image restorer, \emph{i.e.}, Qwen-image-edit, leverages the diagnostic representation generated by the reasoner to guide pixel reconstruction. 
Rather than treating restoration as a purely feed-forward generation problem, R\&R conditions the restoration process on degradation-aware and scene-adaptive cues, enabling the model to better handle complex real-world scenarios involving multiple, entangled degradations. 
Furthermore, the quantified degradation severity inferred during the \emph{Reason} phase is utilized as reinforcement learning signals to strengthen the restoration process, explicitly aligning diagnostic understanding with restoration quality. Overall, the \textbf{Reason-and-Restore (R\&R)} framework establishes a unified and extensible paradigm that tightly couples structured reasoning with image restoration. 
The following subsections describe the \emph{Reason} phase, the \emph{Restore} phase, and the reinforcement learning strategy in detail, respectively.
\subsection{Reason Phase: Structured Degradation Diagnosis}
\paragraph{Semi-realworld Degradation
Model}
The objective of the \emph{Reason} phase is to explicitly diagnose the underlying causes of image degradation before performing pixel-level restoration. Usually, the degraded observation $I_d$ is the result of a composition of physically
and statistically motivated degradation processes applied
to an unknown clean image $J$. In this work, we consider degradations commonly encountered in real-world outdoor environments and propose a \textbf{semi-realworld degradation model}~\footnote{Since degradations are usually forward process with known parameters, this model can be easily extended to include others for more scenarios.} as Eq. \ref{eq:degradation_model}, including fog, motion blur, rain streaks, and sensor noise.
%
%
\begin{equation}
I_d \;=\; \mathcal{B}_{\mathbf{k}}\!\Big( \, t\,J \;+\; (1-t)\,A \,\Big)\;+\;\mathcal{S}_{\boldsymbol{\theta}_r}(J)\;+\;\mathcal{N}_{\boldsymbol{\theta}_n},
\label{eq:degradation_model}
\end{equation}
where $J$ denotes the latent clean image, $t \in [0,1]$ is the mean transmission coefficient characterizing fog severity, $A \in \mathbb{R}^3$ represents atmospheric light, and $\mathcal{B}_{\mathbf{k}}(\cdot)$ denotes a motion blur operator parameterized by kernel parameters $\mathbf{k}$ such as direction and effective length. The term $\mathcal{S}_{\boldsymbol{\theta}_r}(J)$ models rain streaks with parameters $\boldsymbol{\theta}_r$ controlling properties such as density, slant, and streak geometry, while $\mathcal{N}_{\boldsymbol{\theta}_n}$ denotes additive sensor noise parameterized by noise statistics $\boldsymbol{\theta}_n$.

This formulation highlights that real-world image degradation is inherently compositional: multiple degradation factors jointly contribute to the observed image, and their effects are governed by distinct physical or statistical parameters.
\paragraph{Structured Diagnostic Reasoning.}
The \emph{Reason} phase is designed to infer the latent structure underlying Eq.~\eqref{eq:degradation_model} from the degraded observation $I_d$. 
The diagnostic outputs produced by the reasoner are intentionally structured into four complementary components in the order of degradation presence, severity scores, degradation-specific parameters, and scene-level semantics. This design is motivated by the observation that effective restoration requires not only identifying \emph{what} degradations exist, but also understanding \emph{how strongly} they manifest, \emph{how} they are formed, and \emph{where} they occur within a semantic context. The detailed CoT is elaborated as follow:

First, explicit identification of degradation presence (e.g., fog, motion blur, rain streaks, and sensor noise) allows the restoration model to reason over \emph{mixed degradation compositions}, which are common in real-world scenarios but rarely addressed by single-degradation assumptions. 

Second, continuous severity scores provide a relative measure of degradation intensity (e.g., fog intensity $55.9/100$, motion blur $70.9/100$), enabling the restoration process to prioritize dominant degradations and avoid over- or under-correction. 

Third, degradation-specific parameters offer fine-grained, physically or statistically grounded cues that directly inform restoration strategies. For example, atmospheric light and transmission statistics characterize the strength and color bias of fog and noise statistics describe sensor readout characteristics. These parameters correspond to the latent variables in the degradation formation model of Eq.~\eqref{eq:degradation_model} and provide actionable guidance for restoration beyond coarse severity estimation.

Finally, concise scene- and object-level semantic descriptions supply high-level contextual information about scene structure, materials, and illumination conditions (e.g., urban street scenes with buildings and vehicles). Such semantics help disambiguate visually similar degradation patterns and encourage scene-consistent restoration, particularly in regions where low-level cues alone are insufficient.

Together, these four diagnostic components form a unified and interpretable representation of image degradation that bridges physical modeling, statistical characterization, and semantic understanding. This structured diagnosis serves as a critical intermediary between the degraded observation and the restoration process, enabling degradation-aware, scene-adaptive, and robust image restoration.

\paragraph{VLM Fine-tune.}
To realize this diagnostic capability, we fine-tune a reasoning-enabled VLM based on Qwen3-VL to perform task-oriented analysis rather than generic image captioning. Massive training data are generated by randomly degrading clean images according to the proposed semi-realworld degradation model. The model is trained to output corresponding diagnosis following the aforementioned predefined diagnostic format. Such Chain-of-Thought (CoT) decomposes complex inference into intermediate reasoning steps leading to more reliable and consistent decisions. Importantly, the outputs of the reason phase are not optimized for visual fidelity by themselves. Instead, they serve as interpretable, high-level priors that capture the underlying degradation structure and guide the subsequent restoration process. 
\subsection{Restore Phase: Universal Image Restoration}
The \emph{Restore} phase is responsible for translating the structured diagnostic outputs produced by the reasoner into pixel-level reconstruction. Rather than performing restoration in isolation, the restorer should explicitly condition the reconstruction process on degradation-aware and scene-adaptive cues inferred during diagnostic reasoning. This design enables the restorer to adapt its behavior to diverse and mixed degradation scenarios without requiring manual intervention or task-specific configurations.
\paragraph{Restorer Model.}
We instantiate the Restorer using Qwen-Image-Edit as the backbone, a powerful image-editing foundation model capable of high-fidelity and controllable image generation. Qwen-Image-Edit takes the degraded image $I_d$ as input and produces a restored image $\hat{J}$, while being guided by the diagnostic information inferred in the Reason phase. Importantly, the backbone remains task-agnostic and does not rely on separate models for individual degradations, allowing R\&R to support universal image restoration within a single unified framework.
\paragraph{Conditioning with Diagnostic Priors.}
Rather than treating degradation diagnostics as free-form textual prompts, R\&R regularizes them into an interpretable and structured natural-language format and injects them into Qwen-Image-Edit through its image-editing and conditional generation interfaces. Therefore, besides the paired clean/degraded observation, the degradation diagnosis is also recorded as an input in the training data. This design ensures that diagnostic reasoning directly influences pixel reconstruction, instead of merely serving as post-hoc explanations.
Moreover, by conditioning restoration on explicit diagnostic priors, the Restore phase naturally supports mixed-degradation scenarios in which multiple degradation types coexist. Severity-aware conditioning allows the model to balance competing restoration objectives, such as suppressing dominant degradations while preserving fine details affected by weaker ones. As a result, the restoration process becomes adaptive to the degradation composition of each input image, rather than being constrained by assumptions of single-degradation dominance.

Overall, the \emph{Restore} phase operationalizes the \textbf{Reason and Restore} principle by tightly coupling structured diagnostic reasoning with pixel-level reconstruction. This coupling enables universal image restoration to be degradation-aware, scene-adaptive, and robust to the complex degradation patterns encountered in real-world visual environments.
\subsection{Reinforcement Learning with Diagnostic Rewards}
\label{sec:method_rl}
While the reasoner provides structured diagnostic priors $\mathcal{T}$, standard supervised fine-tuning (SFT) of the restorer $\mathcal{F}_{R}$ primarily focuses on pixel-level reconstruction. SFT alone does not guarantee that the restorer explicitly follows the diagnostic ``plan" to mitigate specific degradations. To bridge this gap, we introduce a reinforcement learning stage with diagnostic rewards utilizing group relative policy optimization (GRPO)\citep{shao2024deepseekmath}. 

Traditional RL methods rely on the model's internal log-probabilities, which are computationally expensive to track for diffusion models. Inspired by recent work~\citep{black2024training}, our approach constructs a \textit{fidelity-based policy distribution} as a proxy and optimizes it against \emph{diagnostic rewards}. This creates a closed-loop system where restoration quality is jointly defined by pixel fidelity (MSE) and diagnostic consistency (Severity Reduction).
\paragraph{Group Sampling and Fidelity-Based Policy Distribution.}
Given a degraded input $I_d$ and the diagnostic context $\mathcal{T}$, we sample a group of $N$ candidate restorations $\{\hat{J}_1, \hat{J}_2, \dots, \hat{J}_N\}$ from the current policy $\pi_\theta$, namely the restorer. 
To incorporate ground-truth supervision into the RL framework, we define the probability of selecting a candidate $\hat{J}_k$ based on its reconstruction fidelity. Specifically, we compute the mean squared error (MSE) between each candidate and the ground truth $J_{gt}$ as $\mathcal{E}_k = \| \hat{J}_k - J_{gt} \|^2_2$. We then model the policy's action distribution $P_\theta(\hat{J}_k | I_d, \mathcal{T})$ as a categorical distribution derived from the softmax-normalized negative MSE, $\log P_\theta(\hat{J}_k) = \log \left( \frac{\exp(-\mathcal{E}_k / \tau)}{\sum_{j=1}^N \exp(-\mathcal{E}_j / \tau)} \right)$, where $\tau$ is a temperature hyper-parameter controlling the sharpness of the distribution. This formulation is theoretically grounded in energy-based models (EBMs), and a detailed mathematical derivation of the proxy is given in Appendix.
\paragraph{Diagnostic Reward Calculation.}
The reward signal is provided by the frozen diagnostic model $\Phi$, namely the reasoner, which serves as the environment feedback. For each candidate $\hat{J}_k$, $\Phi$ predicts the residual severity scores $s(\hat{J}_k)$. The reward $r_k$ is defined as the reduction in severity compared to the input $I_d$, encouraging the removal of diagnosed degradations,$r_k = \sum_{i \in N} \left( s_i(I_d) - s_i(\hat{J}_k) \right)$.
\paragraph{Optimization via GRPO.}
To optimize the model, we employ GRPO, which estimates the baseline for the advantage function using the group average of rewards. Here, the gradient descent updates the model parameters $\theta$ such that trajectories yielding high diagnostic rewards ($A_k > 0$) are assigned lower MSE (higher probability). This mechanism essentially aligns the "restoration intent" (diagnosis) with the "restoration execution" (pixels). The overall RL training procedure is summarized in Algorithm \ref{alg:rl_training}.
\begin{algorithm}[t]
   \caption{RL Fine-tuning with GRPO for R\&R} \label{alg:rl_training}
\begin{algorithmic}
   \STATE {\bfseries Input:} \small
   Dataset $\mathcal{D}$, Restorer  $\pi_\theta$,Reasoner $\Phi$, Group Size $N$.
   \FOR{each training step}
   \STATE Sample batch $(I_d, J_{gt})$ from $\mathcal{D}$.
   \STATE Generate diagnostic priors $\mathcal{T} = \Phi(I_d)$.
   \STATE \textbf{Rollout:} \STATE
   Generate $N$ candidates $\{\hat{J}_1, \dots, \hat{J}_N\}$ via $\pi_\theta(I_d, \mathcal{T})$.
   \STATE \textbf{Fidelity Evaluation:} 
   \STATE \quad Compute $\mathcal{E}_k = \| \hat{J}_k - J_{gt} \|^2$ for $k=1 \dots N$.
   \STATE \quad Compute log-probs $\log P_\theta(\hat{J}_k)$ via Softmax($-\mathcal{E}$).
   \STATE \textbf{Diagnostic Evaluation:}
   \STATE \quad Predict severities $s(\hat{J}_k) = \Phi(\hat{J}_k)$.
   \STATE \quad Compute rewards $r_k = \text{Sum}(s(I_d) - s(\hat{J}_k))$.
   \STATE \textbf{Optimization:}
   \STATE \quad Compute Advantages $A_k = (r_k - \text{mean}(r)) / \text{std}(r)$.
   \STATE \quad Update $\theta$ minimizing
$\mathcal{L}_{GRPO} = -\frac{1}{N}\sum A_k \log P_{\theta}(\hat{J}_k)$.
   \ENDFOR
\end{algorithmic}
\end{algorithm}

\label{sec:method}

\section{Experiments}
\subsection{Experimental Settings}
We conduct experiments on both synthetic and real-world datasets to comprehensively evaluate the proposed method. For synthetic benchmarks, we follow standard evaluation protocols and use the OTS and RESIDE datasets~\cite{li2018benchmarking, zhao2020dehazing}, where degradations are artificially generated using the proposed semi-realworld degradation model. The RESIDE dataset mainly consists of \emph{indoor} scenes, with 1,399 images used for training and 50 images for testing, while the OTS dataset focuses on \emph{outdoor} scenes, containing 1,861 training images and 200 test images. To ensure fair comparison, our method and all end-to-end baseline approaches are trained and evaluated on the same training and test splits of OTS and RESIDE. 

All models are trained using the same synthetic training data and default settings unless otherwise specified. Training is performed on NVIDIA H200 GPUs. In addition, to assess generalization to real-world scenarios, we construct a new real-world test set collected under diverse outdoor conditions. The dataset consists of 15 representative scenes covering a wide range of environmental settings and degradation patterns, with more than 700 real images. This real-world dataset is used exclusively for evaluation and will be publicly released.
\begin{table}[ht]
\small
\centering
\caption{Quantitative comparison on OTS and RESIDE test sets. Higher PSNR and SSIM indicate better restoration quality.}
\label{tab:main_results}
\begin{tabular}{l|cc|cc}
\hline
\multirow{2}{*}{Method} 
& \multicolumn{2}{c|}{OTS} 
& \multicolumn{2}{c}{RESIDE} \\
& PSNR$\uparrow$ & SSIM$\uparrow$ & PSNR$\uparrow$ & SSIM$\uparrow$ \\
\hline
Input              & 13.4610 & 0.3258 & 13.0450 & 0.3182 \\
3D                 & 13.0786 & 0.3739 & 13.9362 & 0.4539 \\
FoundIR            & 14.4094 & 0.3579 & 14.0800 & 0.3586 \\
Img2Img-Turbo      & 17.8744 & 0.5715 & 16.7320 & 0.5479 \\
Stable Diffusion3  & 18.8597 & 0.5918 & 15.4911 & 0.5102 \\
Qwen-Image         & 18.5255 & 0.5788 & 16.1400 & 0.6100 \\
\hline
\hline (Ours)R\&R   & \textbf{19.564} & \textbf{0.6214} & \textbf{17.0036} & \textbf{0.6188} \\
\hline
\end{tabular}
\end{table}
\begin{figure*}[ht]
    \centering
    \begin{subfigure}{0.12\textwidth}
        \centering
        \textbf{Input}
    \end{subfigure}
    \begin{subfigure}{0.12\textwidth}
        \centering
        \textbf{3D}
    \end{subfigure}
    \begin{subfigure}{0.12\textwidth}
        \centering
        \textbf{FoundIR}
    \end{subfigure}
    \begin{subfigure}{0.12\textwidth}
        \centering
        \textbf{Img2Img-Turbo}
    \end{subfigure}
    \begin{subfigure}{0.12\textwidth}
        \centering
        \textbf{Stable Diffusion3}
    \end{subfigure}
    \begin{subfigure}{0.12\textwidth}
        \centering
        \textbf{Qwen3-Image}
    \end{subfigure}
    \begin{subfigure}{0.12\textwidth}
        \centering
        \textbf{(Ours) R\&R}
    \end{subfigure}
    \begin{subfigure}{0.12\textwidth}
        \centering
        \textbf{GT}
    \end{subfigure}

    \includegraphics[width=0.12\textwidth]{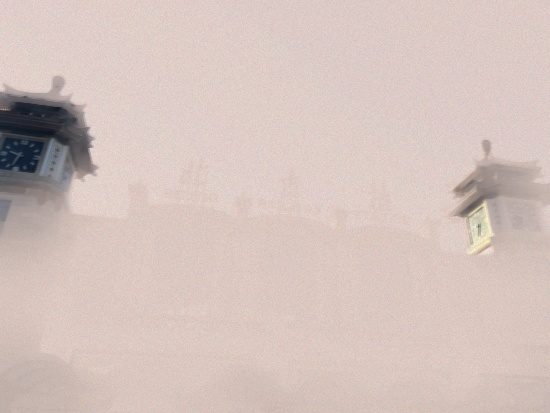}
    \includegraphics[width=0.12\textwidth]{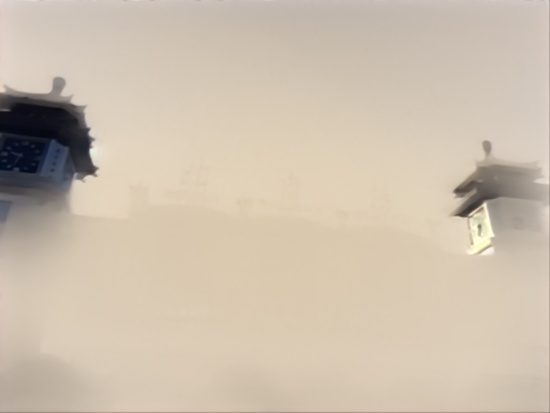}
    \includegraphics[width=0.12\textwidth]{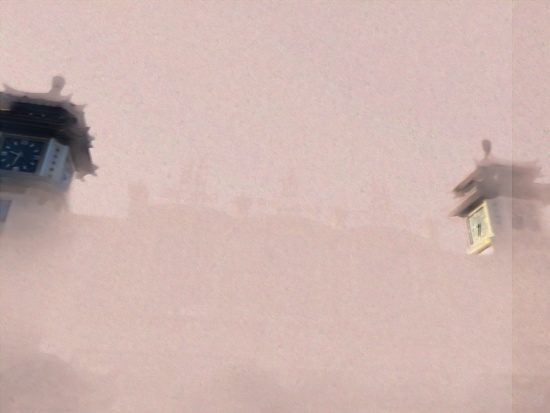}
    \includegraphics[width=0.12\textwidth]{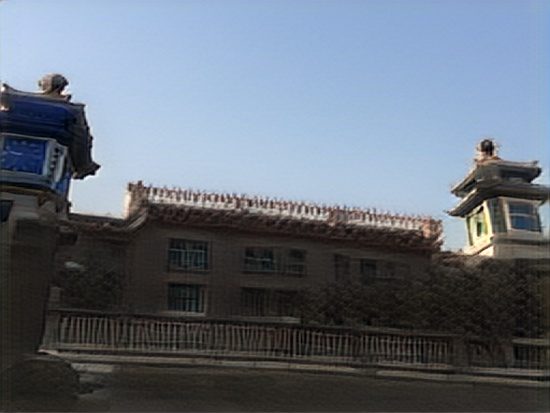}
    \includegraphics[width=0.12\textwidth]{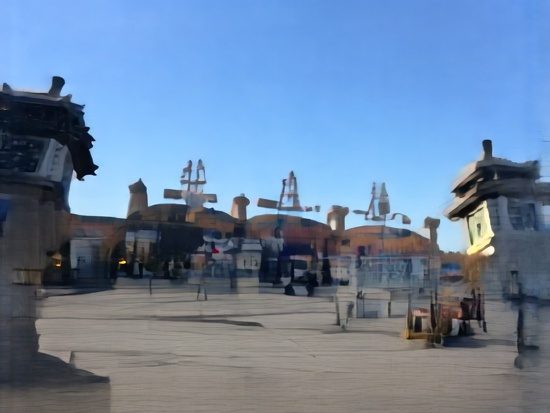}
    \includegraphics[width=0.12\textwidth]{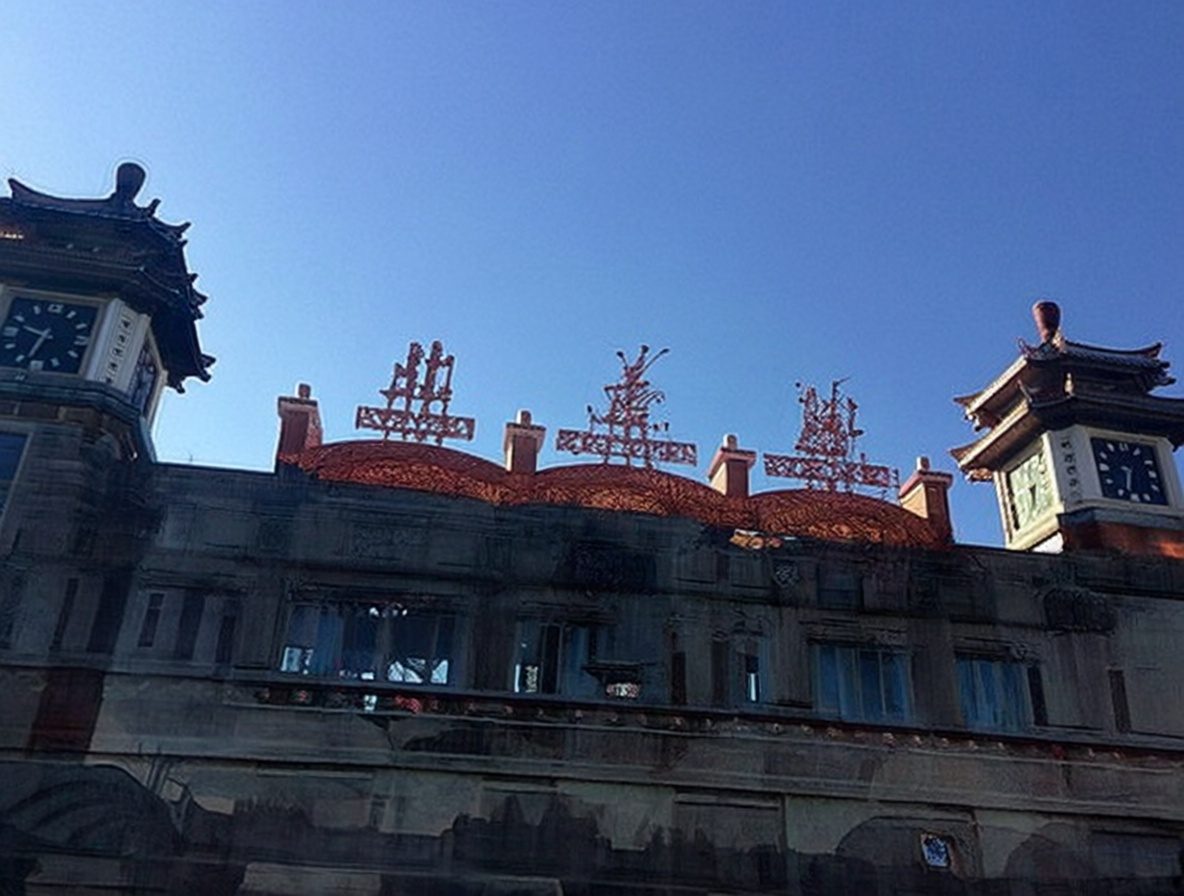}
    \includegraphics[width=0.12\textwidth]{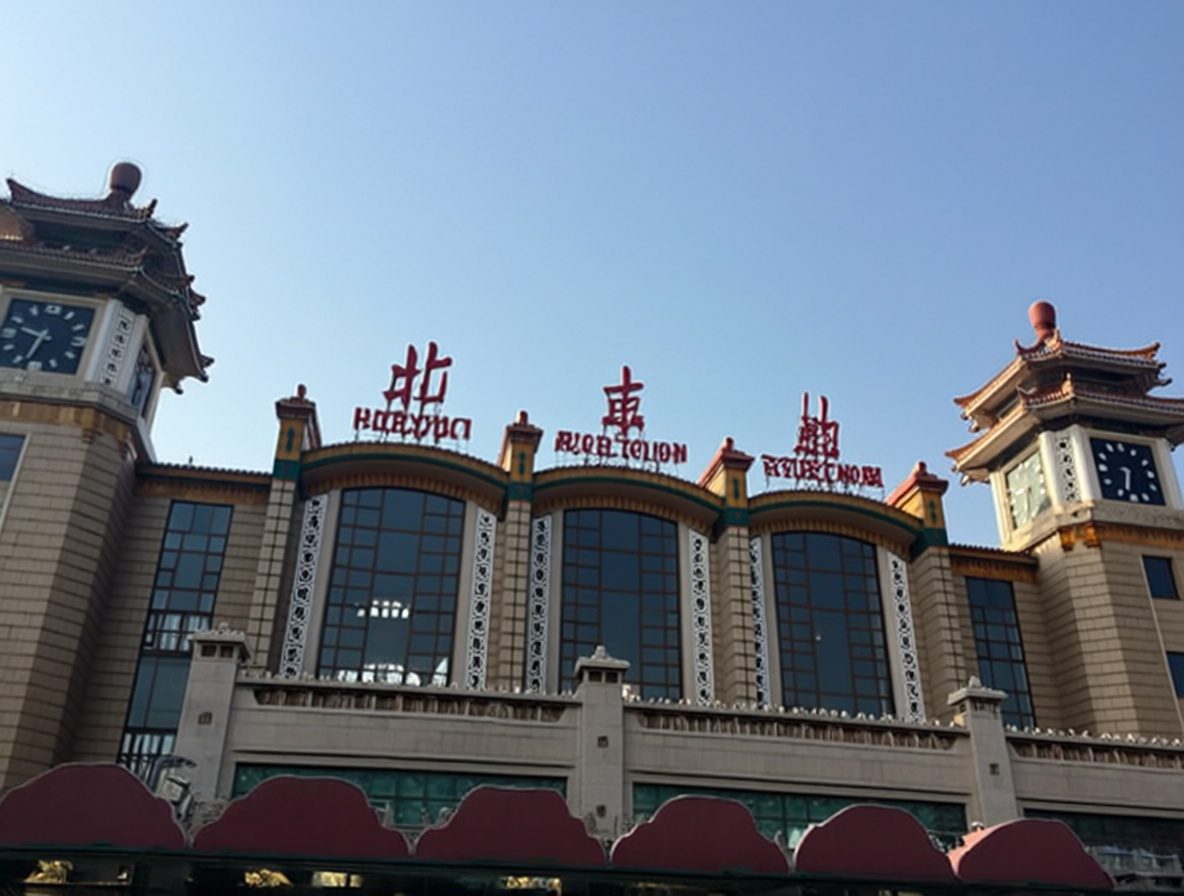}
    \includegraphics[width=0.12\textwidth]{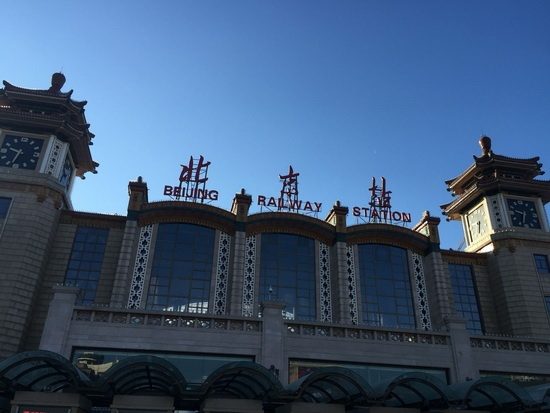}

    \includegraphics[width=0.12\textwidth]{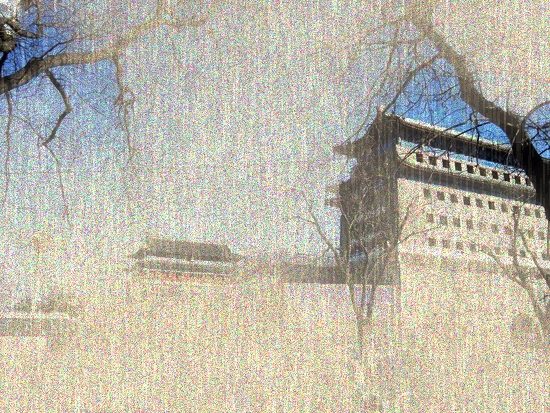}
    \includegraphics[width=0.12\textwidth]{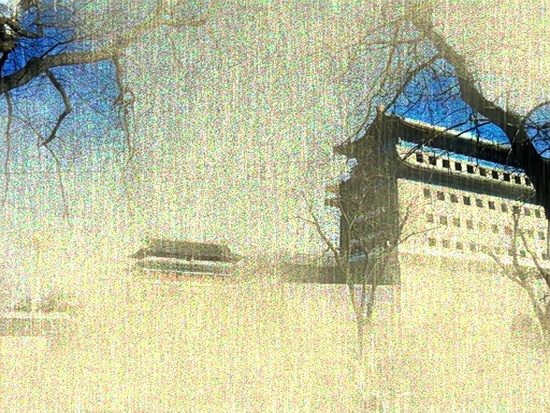}
    \includegraphics[width=0.12\textwidth]{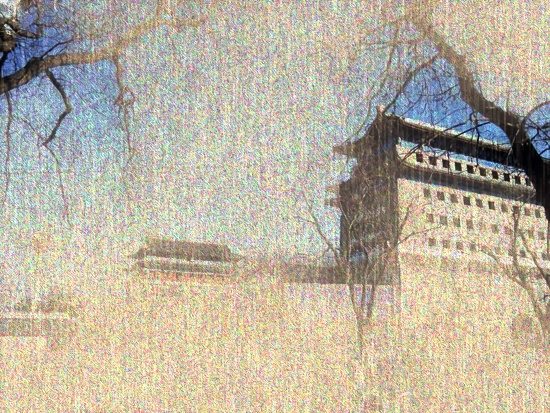}
    \includegraphics[width=0.12\textwidth]{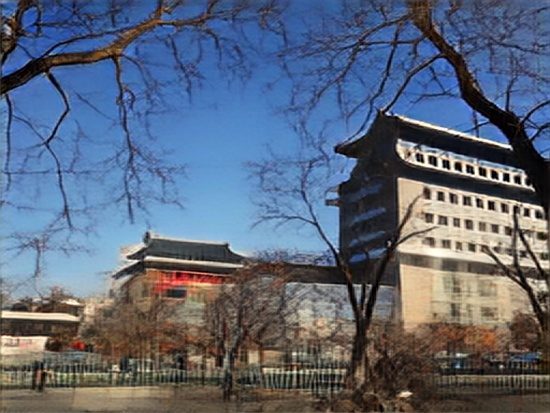}
    \includegraphics[width=0.12\textwidth]{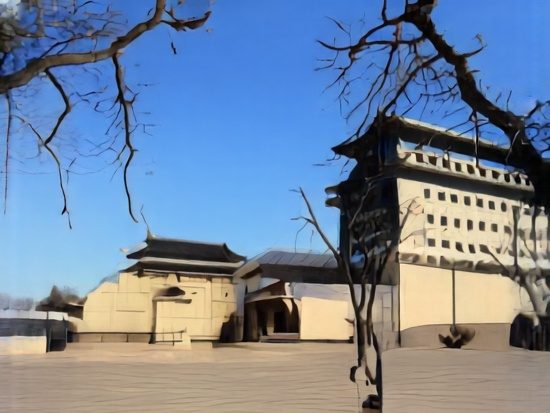}
    \includegraphics[width=0.12\textwidth]{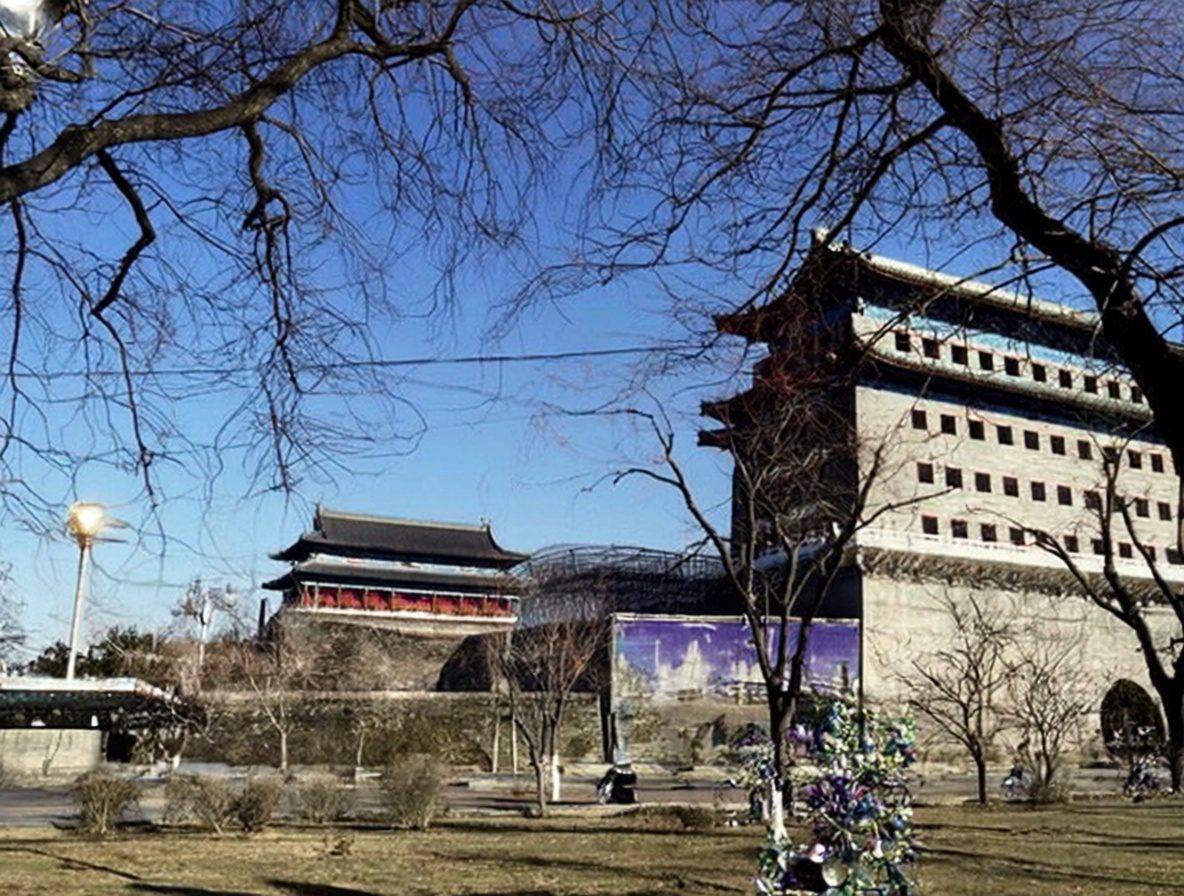}
    \includegraphics[width=0.12\textwidth]{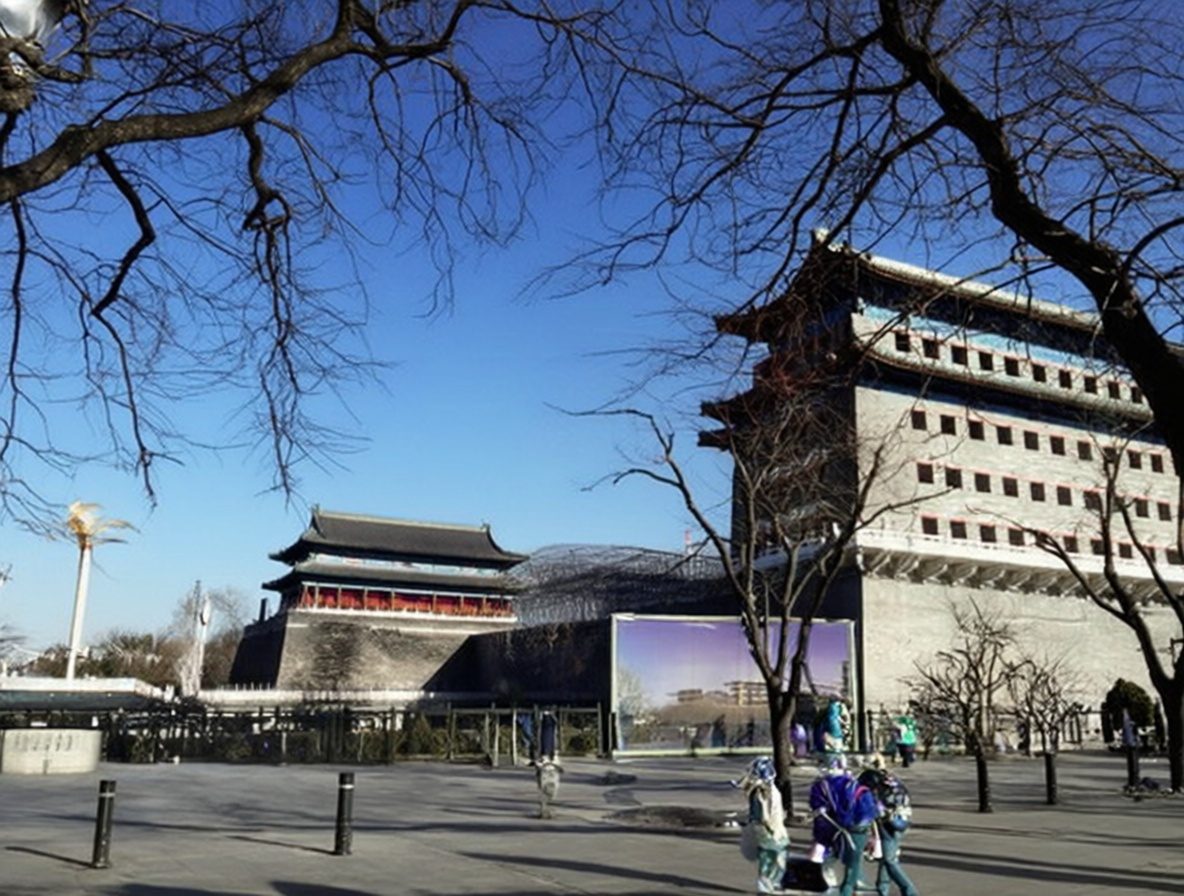}
    \includegraphics[width=0.12\textwidth]{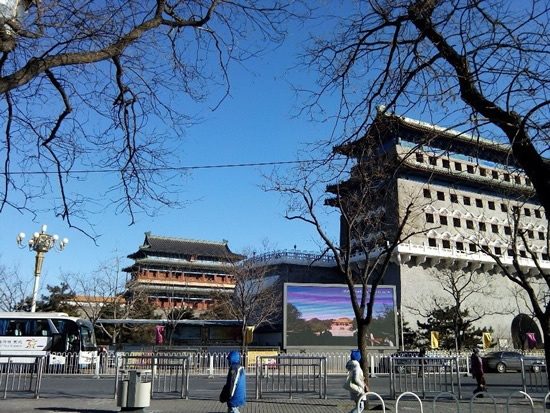}

    \includegraphics[width=0.12\textwidth]{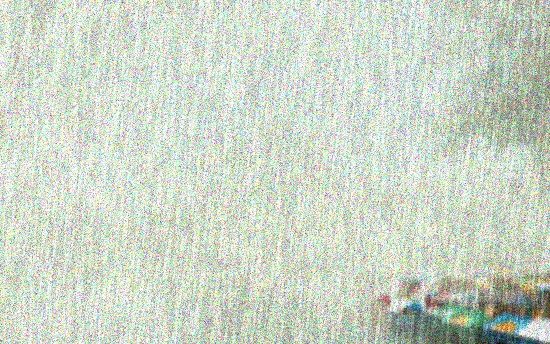}
    \includegraphics[width=0.12\textwidth]{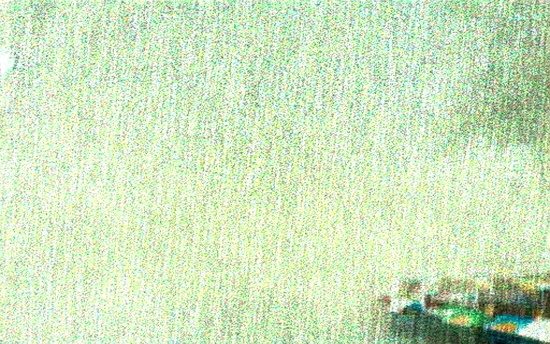}
    \includegraphics[width=0.12\textwidth]{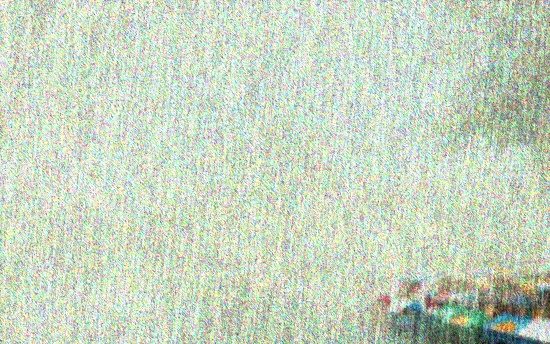}
    \includegraphics[width=0.12\textwidth]{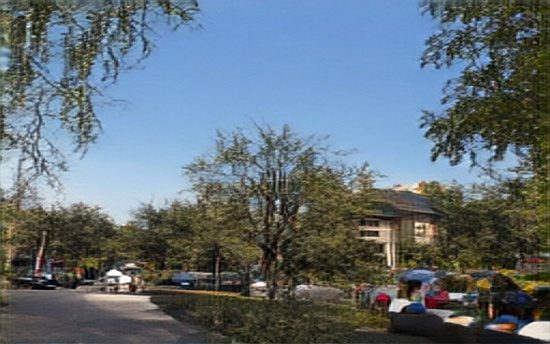}
    \includegraphics[width=0.12\textwidth]{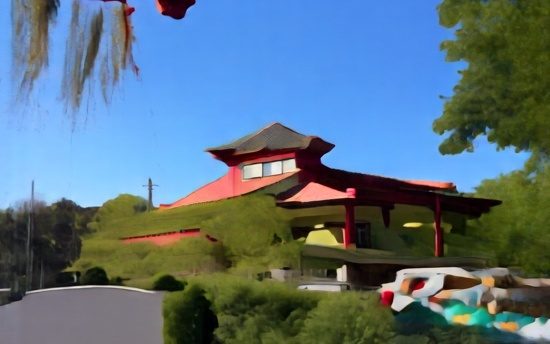}
    \includegraphics[width=0.12\textwidth]{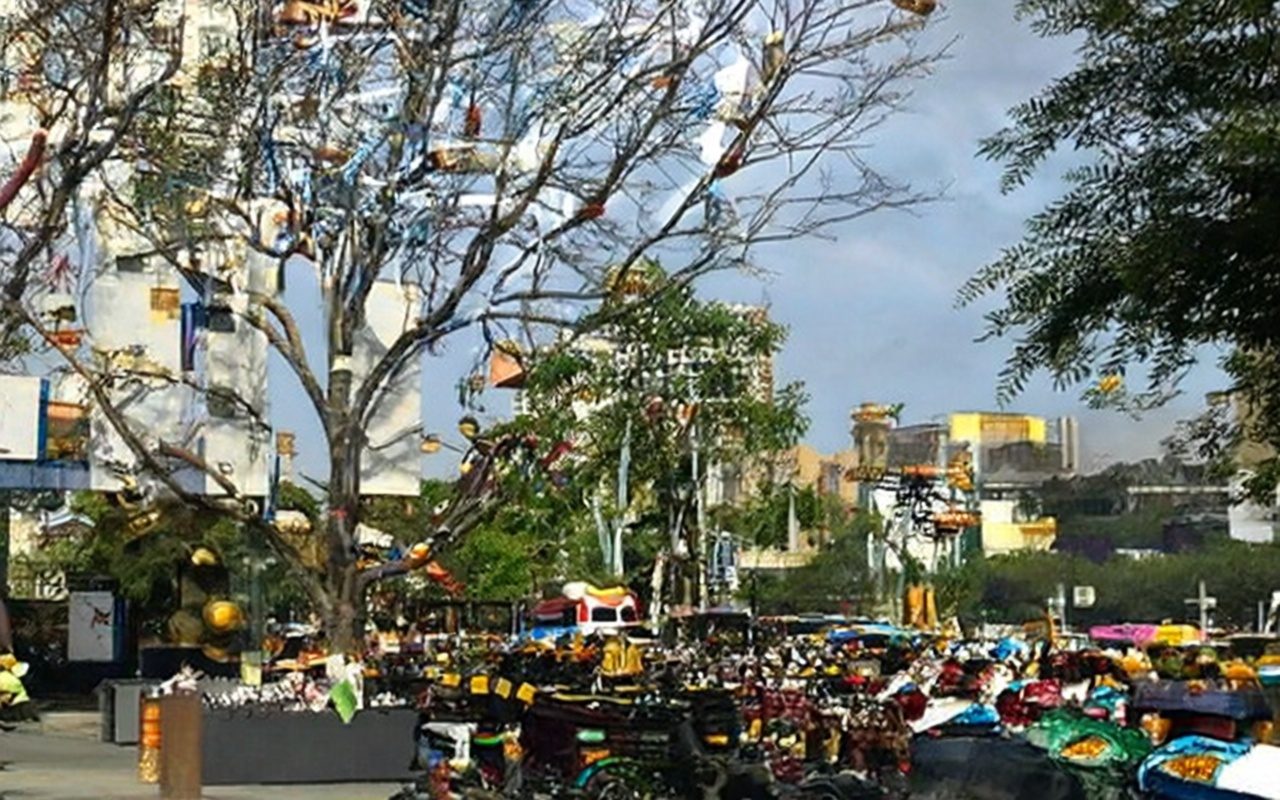}
    \includegraphics[width=0.12\textwidth]{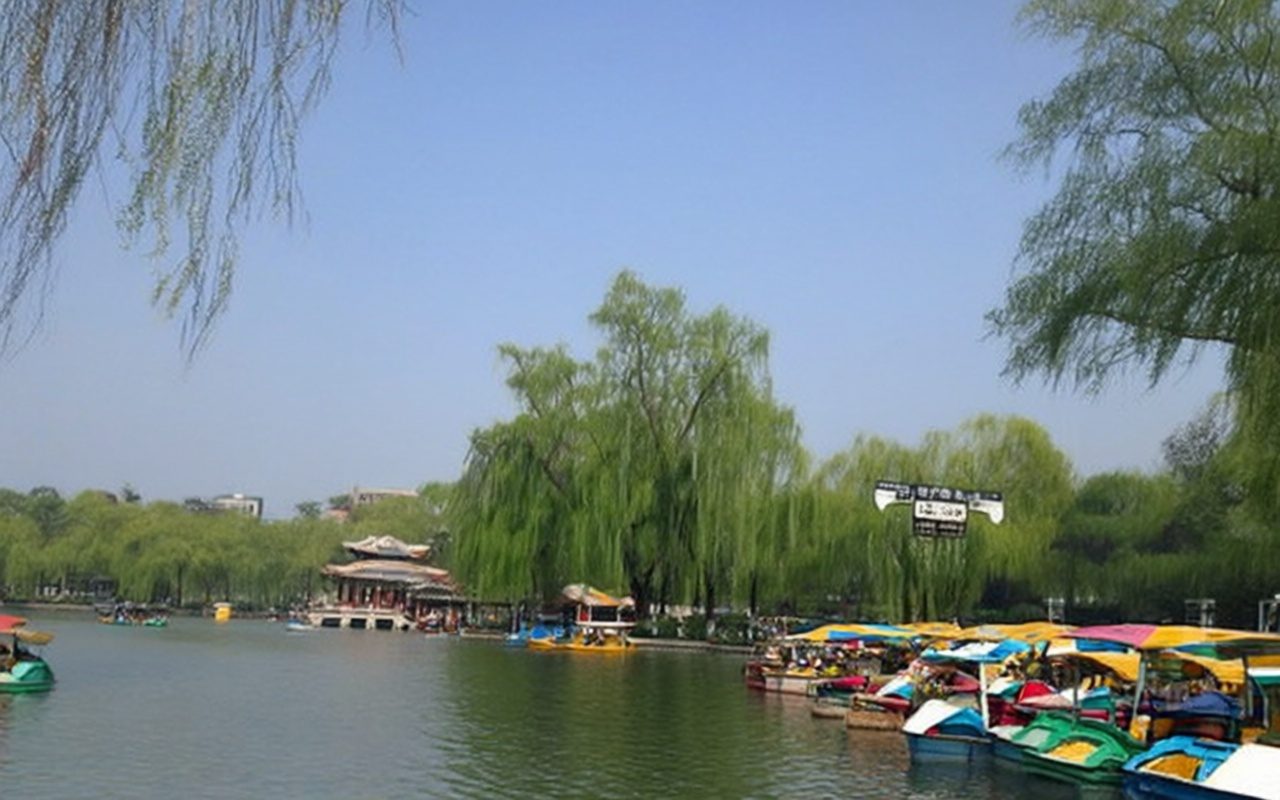}
    \includegraphics[width=0.12\textwidth]{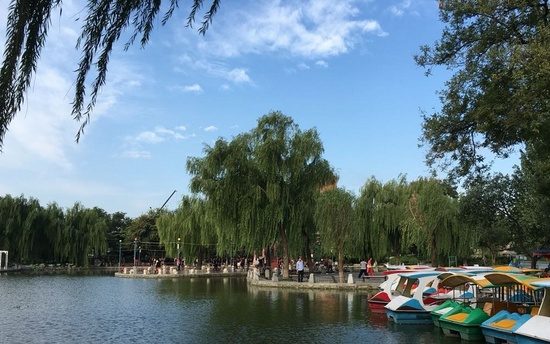}    

    \includegraphics[width=0.12\textwidth]{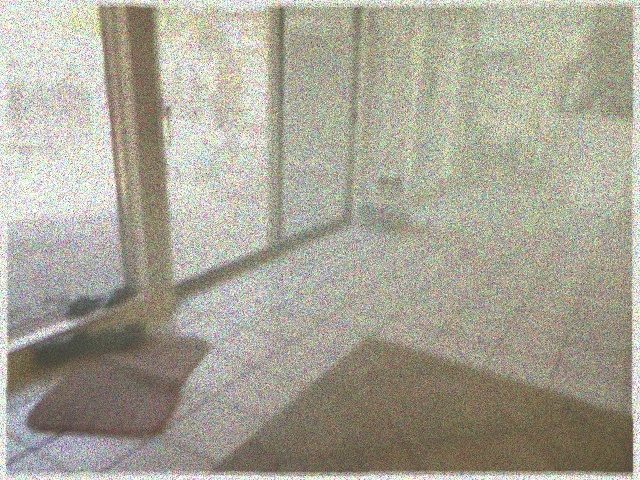}
    \includegraphics[width=0.12\textwidth]{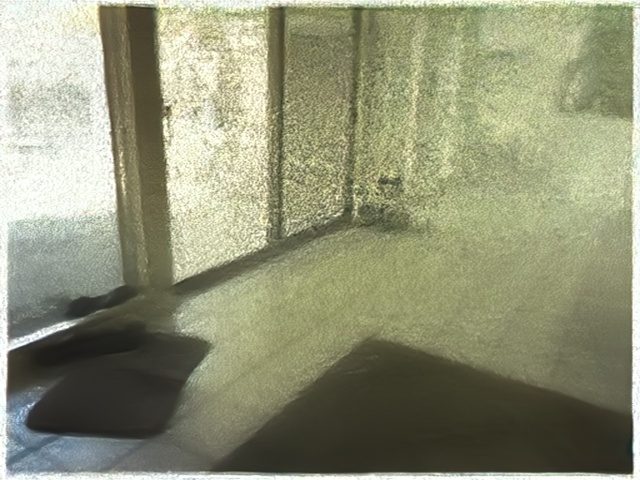}
    \includegraphics[width=0.12\textwidth]{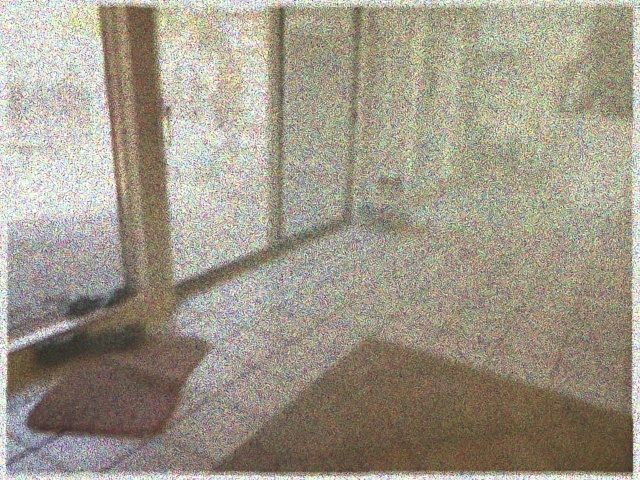}
    \includegraphics[width=0.12\textwidth]{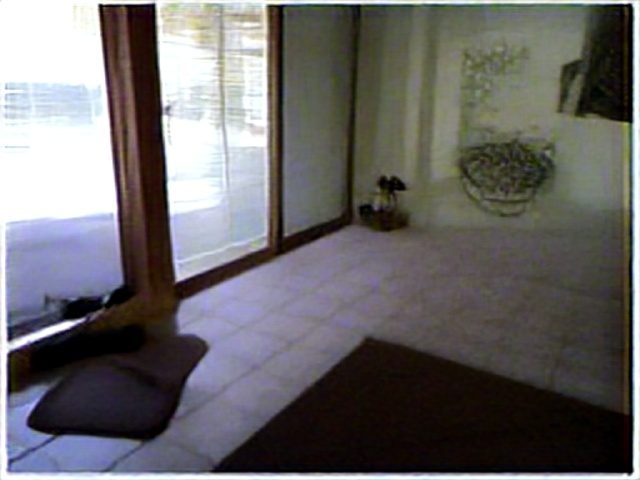}
    \includegraphics[width=0.12\textwidth]{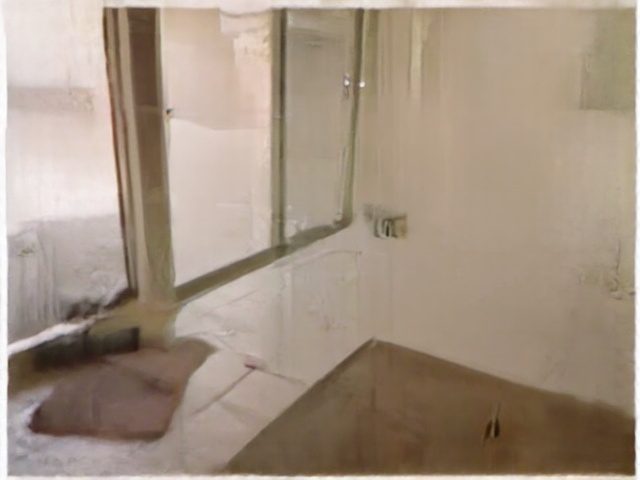}
    \includegraphics[width=0.12\textwidth]{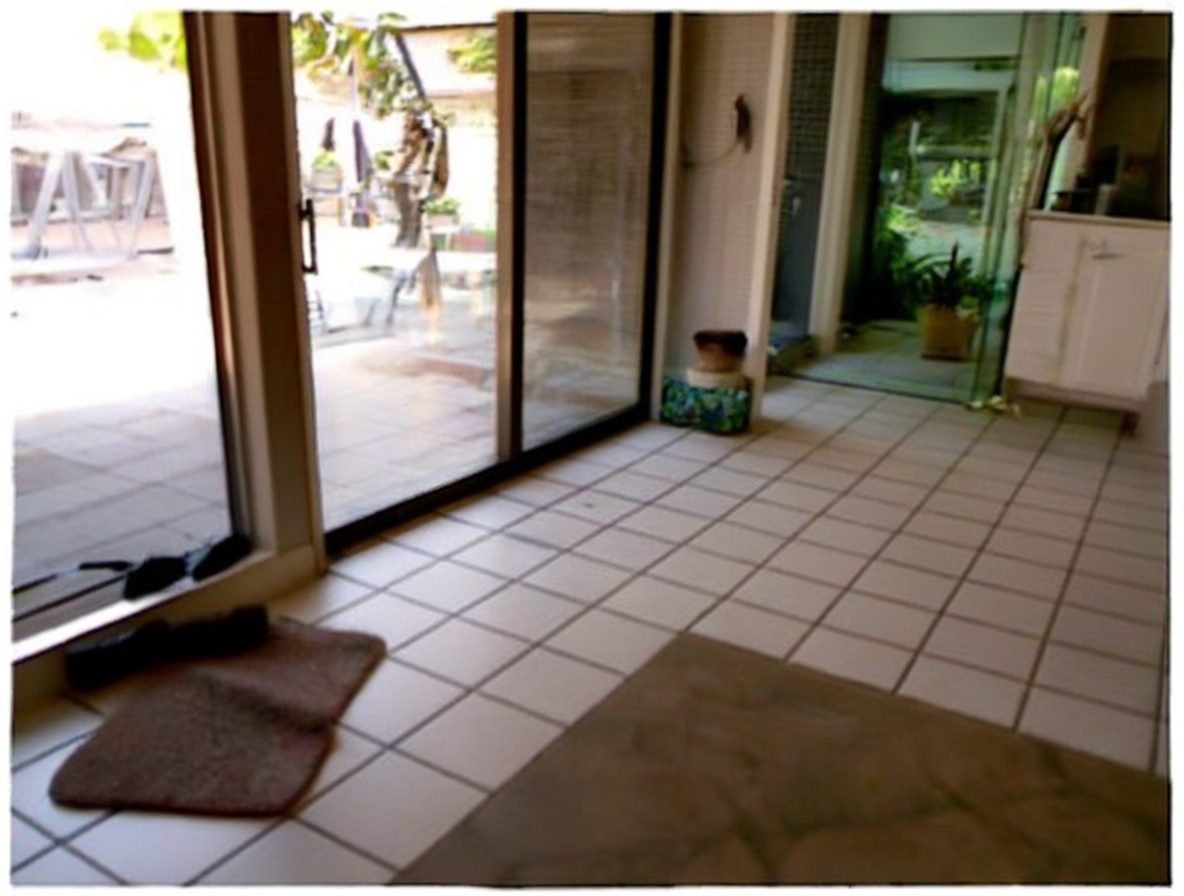}
    \includegraphics[width=0.12\textwidth]{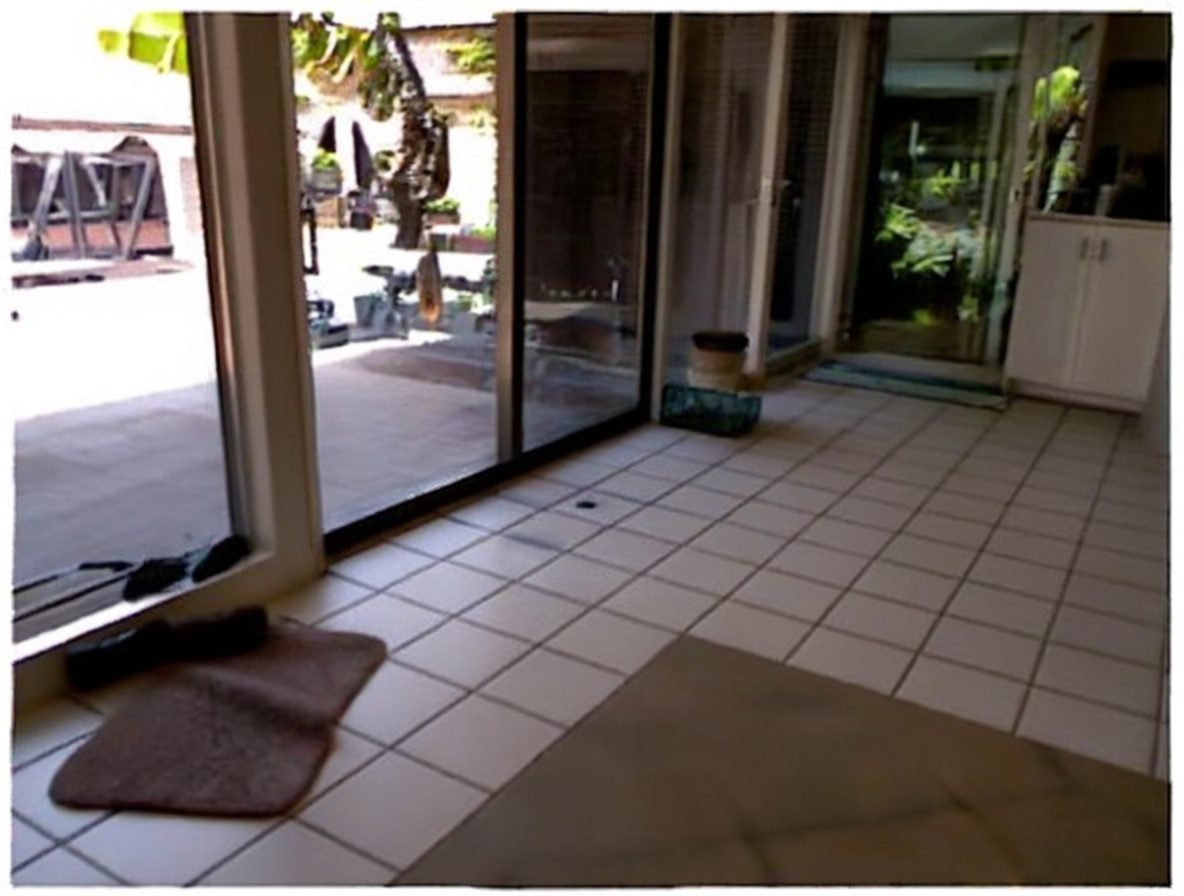}
    \includegraphics[width=0.12\textwidth]{figures/RESIDE/RESIDE_test_gt_jpg/1410.jpg}
    
    \includegraphics[width=0.12\textwidth]{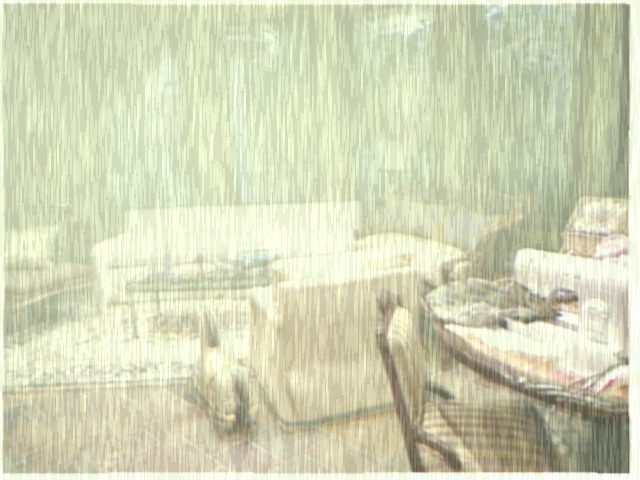}
    \includegraphics[width=0.12\textwidth]{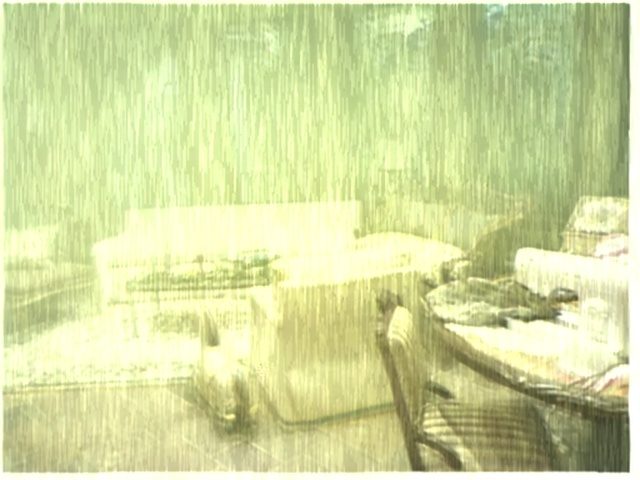}
    \includegraphics[width=0.12\textwidth]{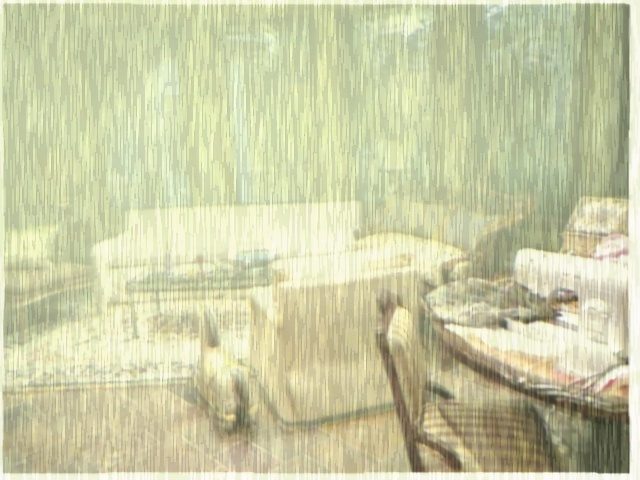}
    \includegraphics[width=0.12\textwidth]{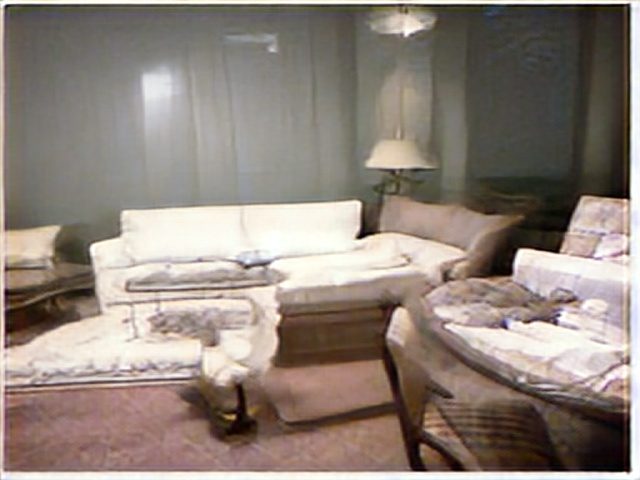}
    \includegraphics[width=0.12\textwidth]{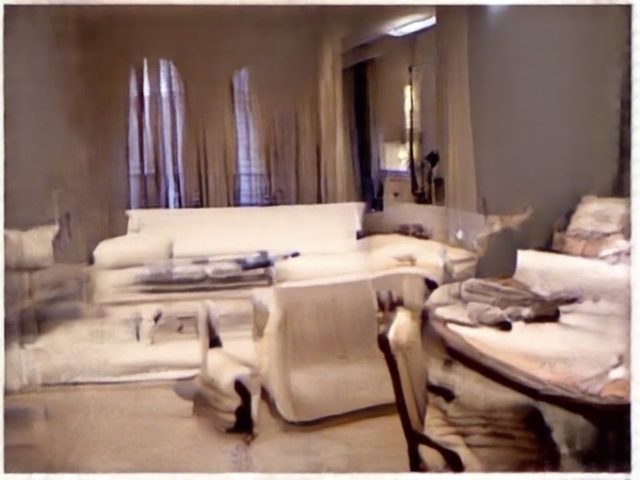}
    \includegraphics[width=0.12\textwidth]{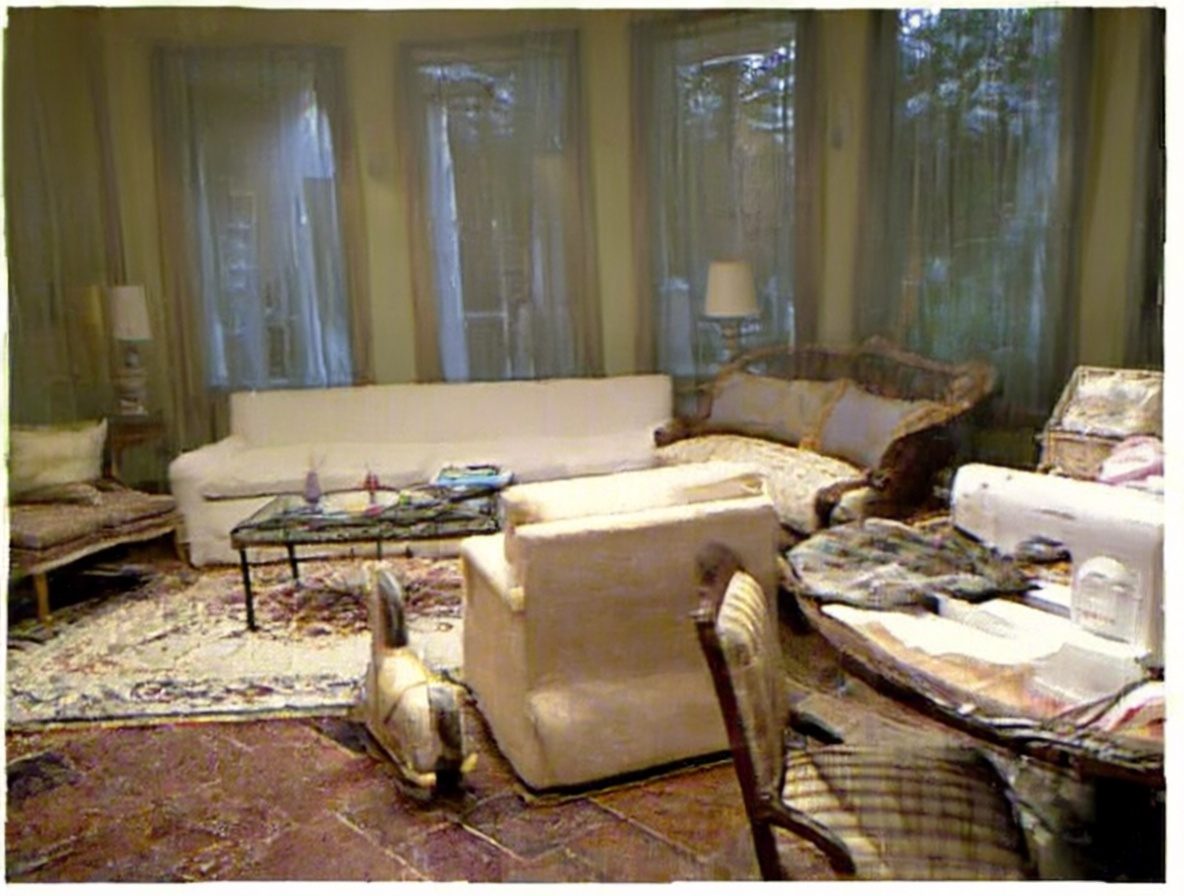}
    \includegraphics[width=0.12\textwidth]{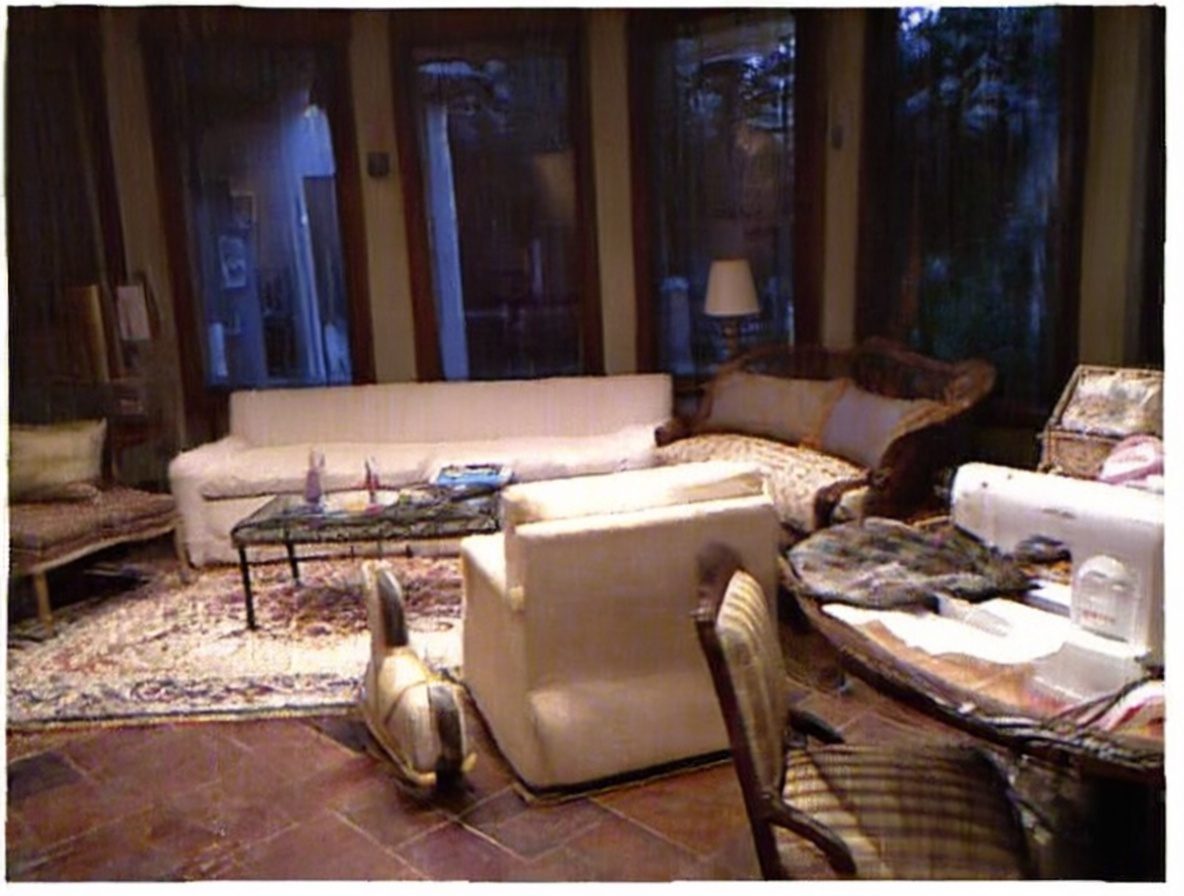}
    \includegraphics[width=0.12\textwidth]{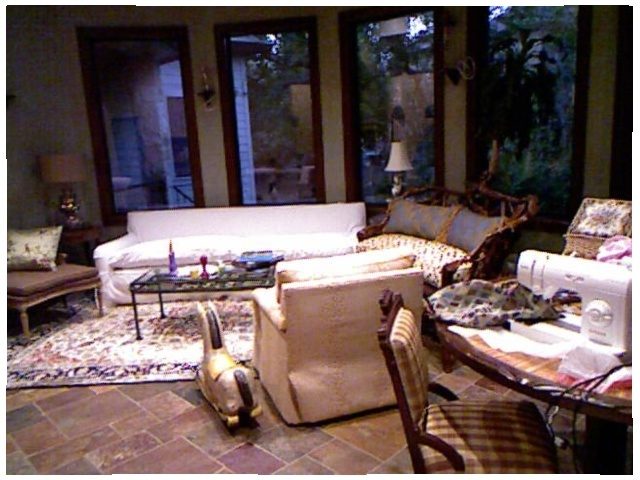}

    \includegraphics[width=0.12\textwidth]{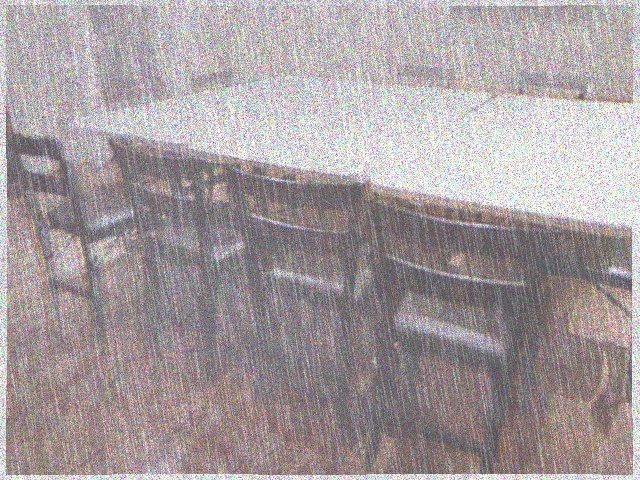}
    \includegraphics[width=0.12\textwidth]{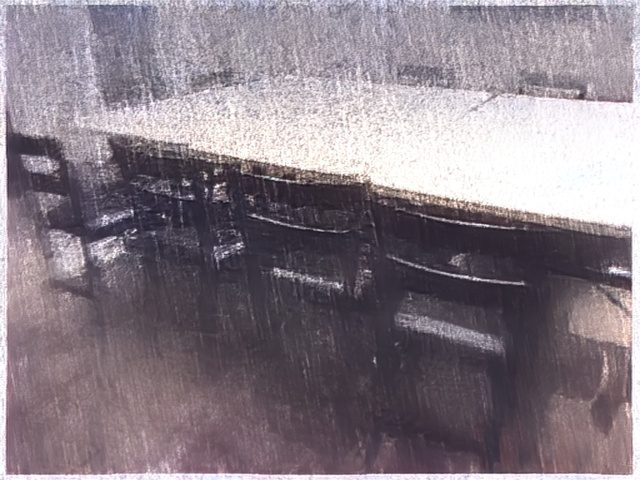}
    \includegraphics[width=0.12\textwidth]{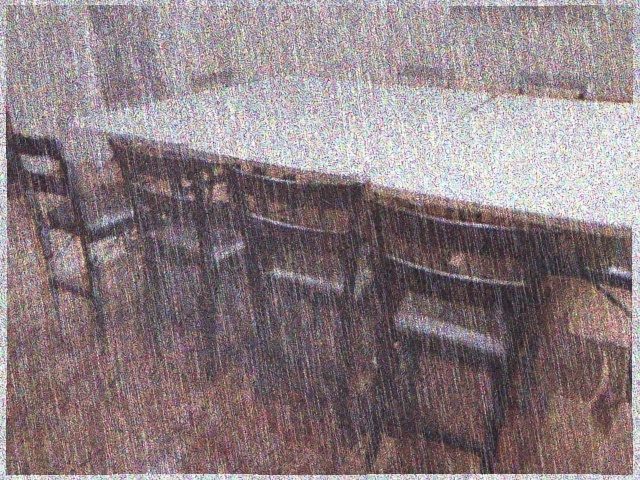}
    \includegraphics[width=0.12\textwidth]{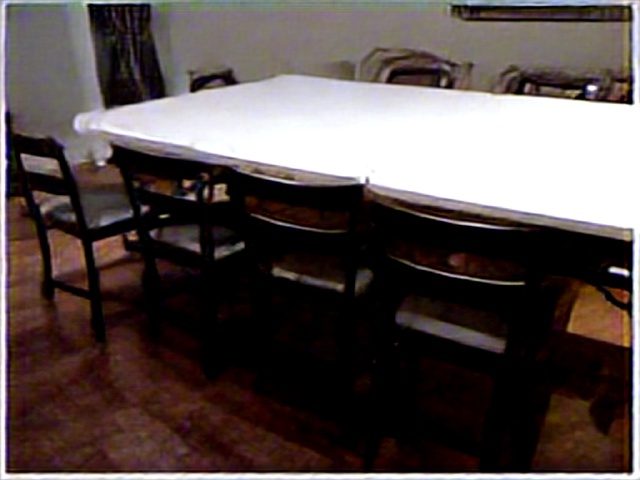}
    \includegraphics[width=0.12\textwidth]{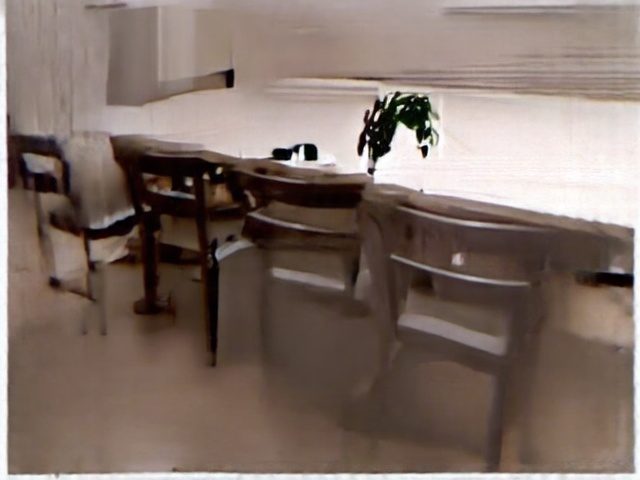}
    \includegraphics[width=0.12\textwidth]{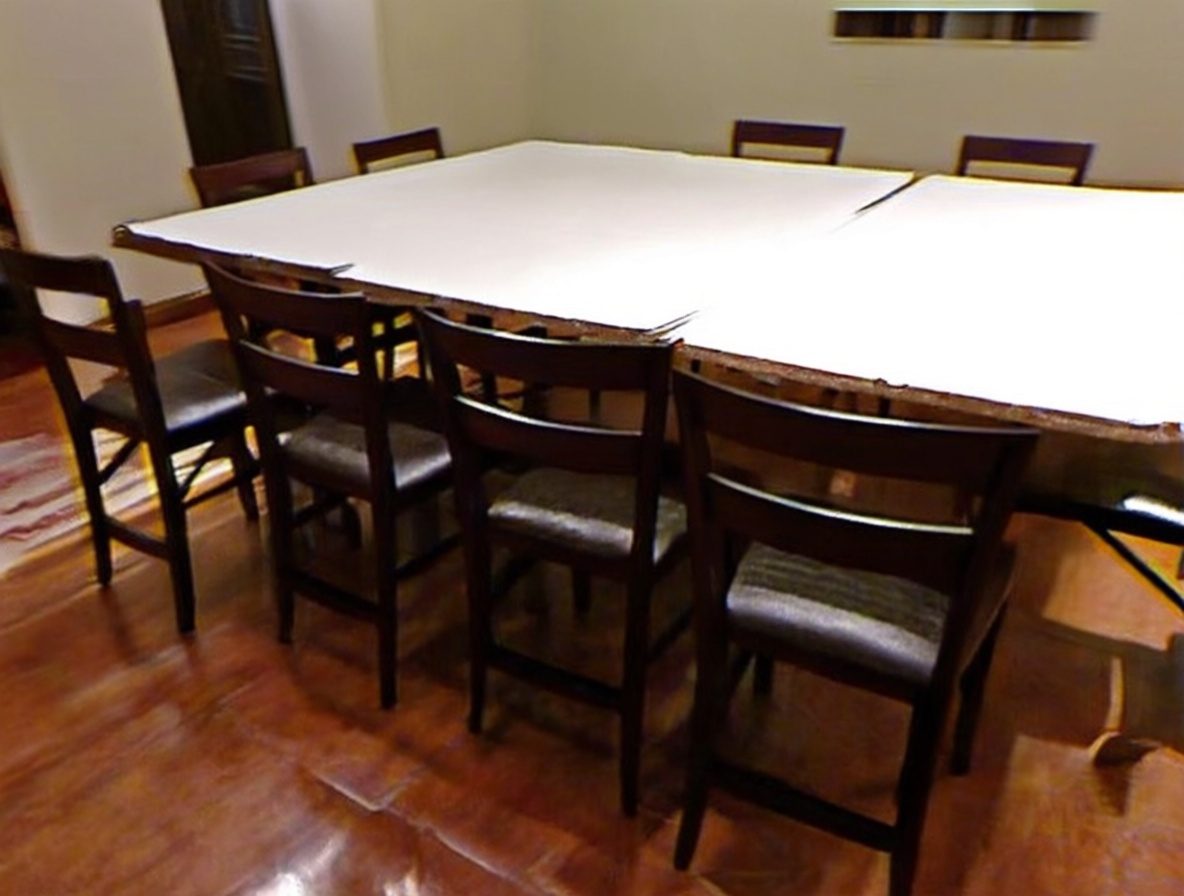}
    \includegraphics[width=0.12\textwidth]{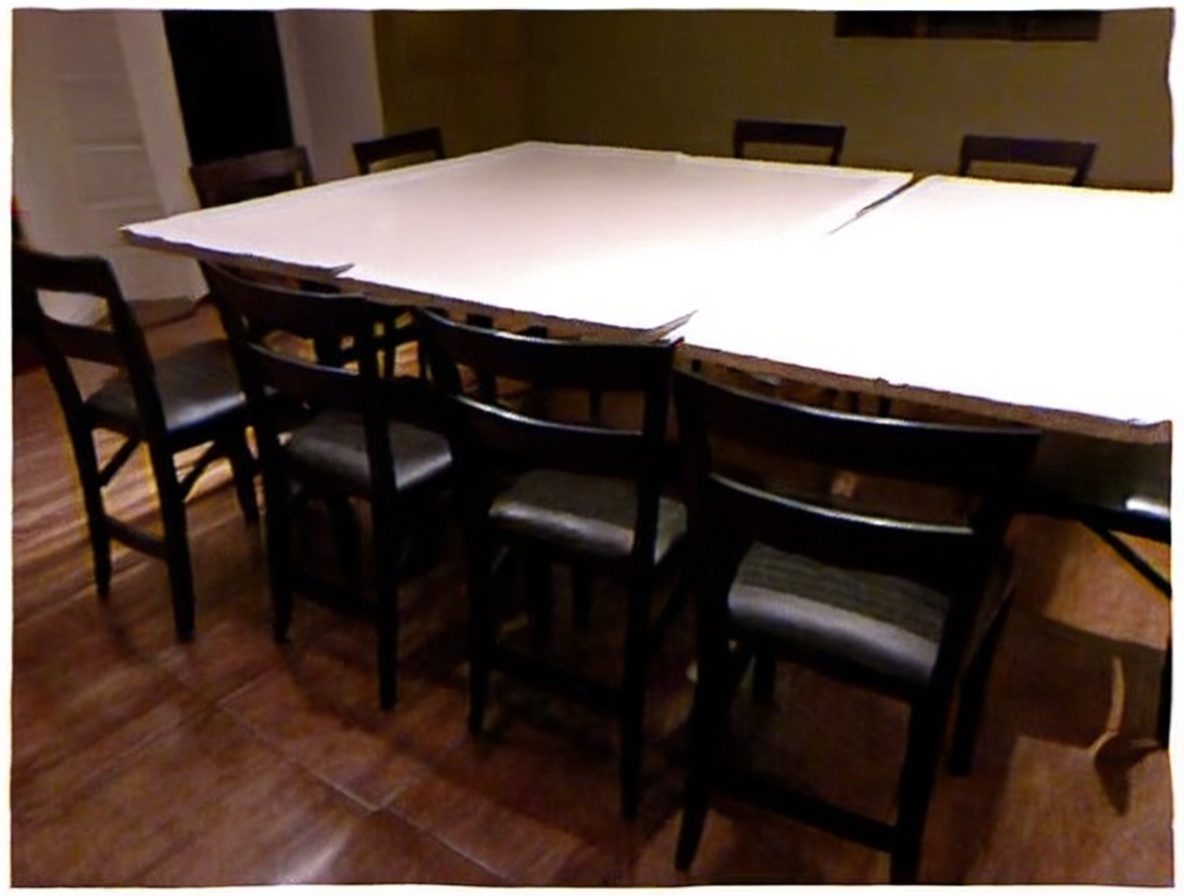}
    \includegraphics[width=0.12\textwidth]{figures/RESIDE/RESIDE_test_gt_jpg/1402.jpg}

    \caption{Qualitative comparison on the OTS (first 3 rows) and RESIDE (last 3 rows) test sets.}
    \label{fig:ots_visual}
\end{figure*}
\subsection{Qualitative Experiment}
\begin{figure*}[t]
    \centering

    \begin{subfigure}{0.13\textwidth}
        \centering
        \textbf{Input}
    \end{subfigure}
    \begin{subfigure}{0.13\textwidth}
        \centering
        \textbf{3D}
    \end{subfigure}
    \begin{subfigure}{0.13\textwidth}
        \centering
        \textbf{FoundIR}
    \end{subfigure}    
    \begin{subfigure}{0.13\textwidth}
        \centering
        \textbf{Img2Img-Turbo}
    \end{subfigure}
    \begin{subfigure}{0.13\textwidth}
        \centering
        \textbf{Stable Diffusion3}
    \end{subfigure}
    \begin{subfigure}{0.13\textwidth}
        \centering
        \textbf{Qwen3-Image}
    \end{subfigure}
    \begin{subfigure}{0.13\textwidth}
        \centering
        \textbf{(Ours) R\&R}
    \end{subfigure}
    \vspace{2mm}

    \includegraphics[width=0.13\textwidth]{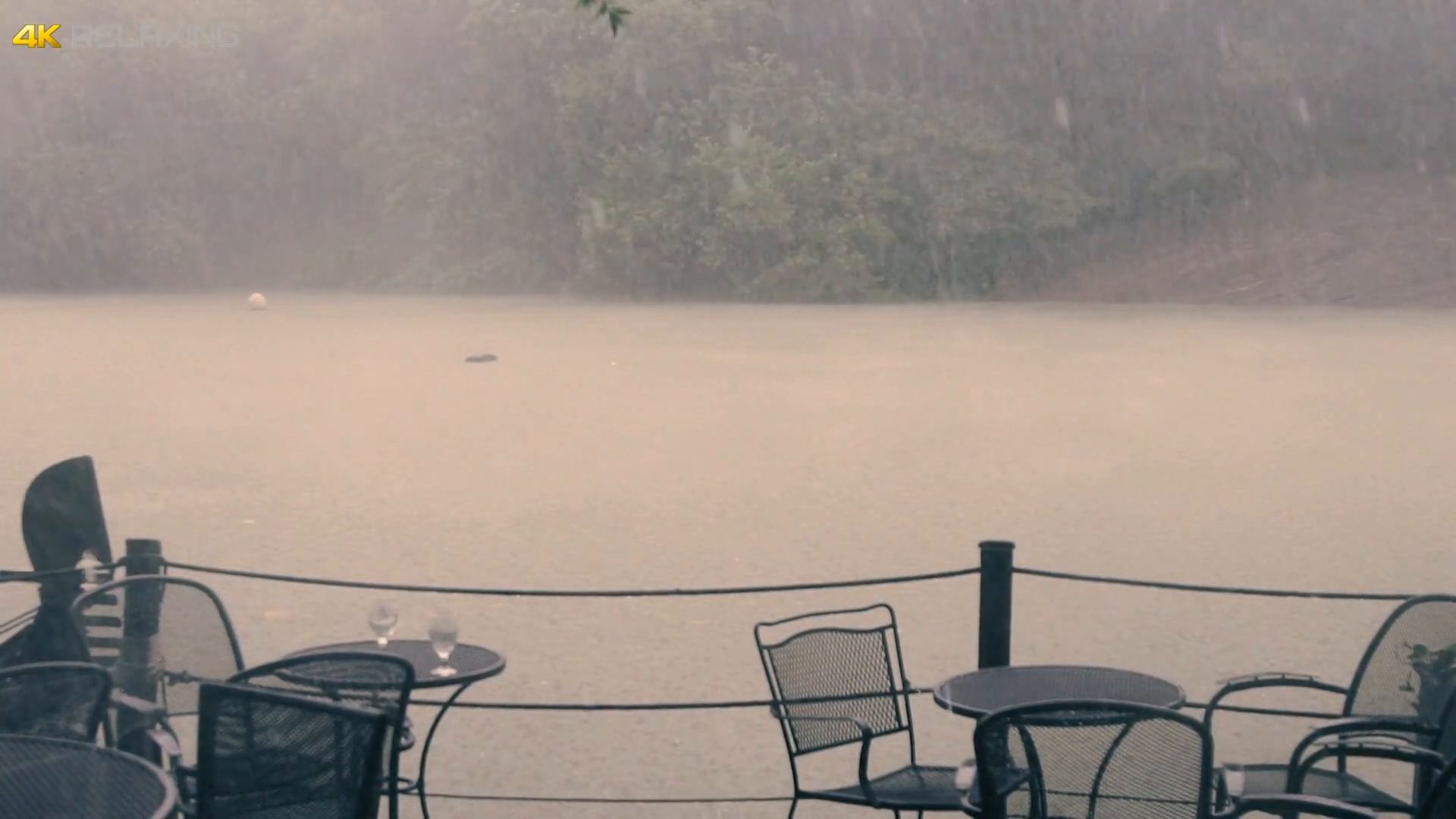}
    \includegraphics[width=0.13\textwidth]{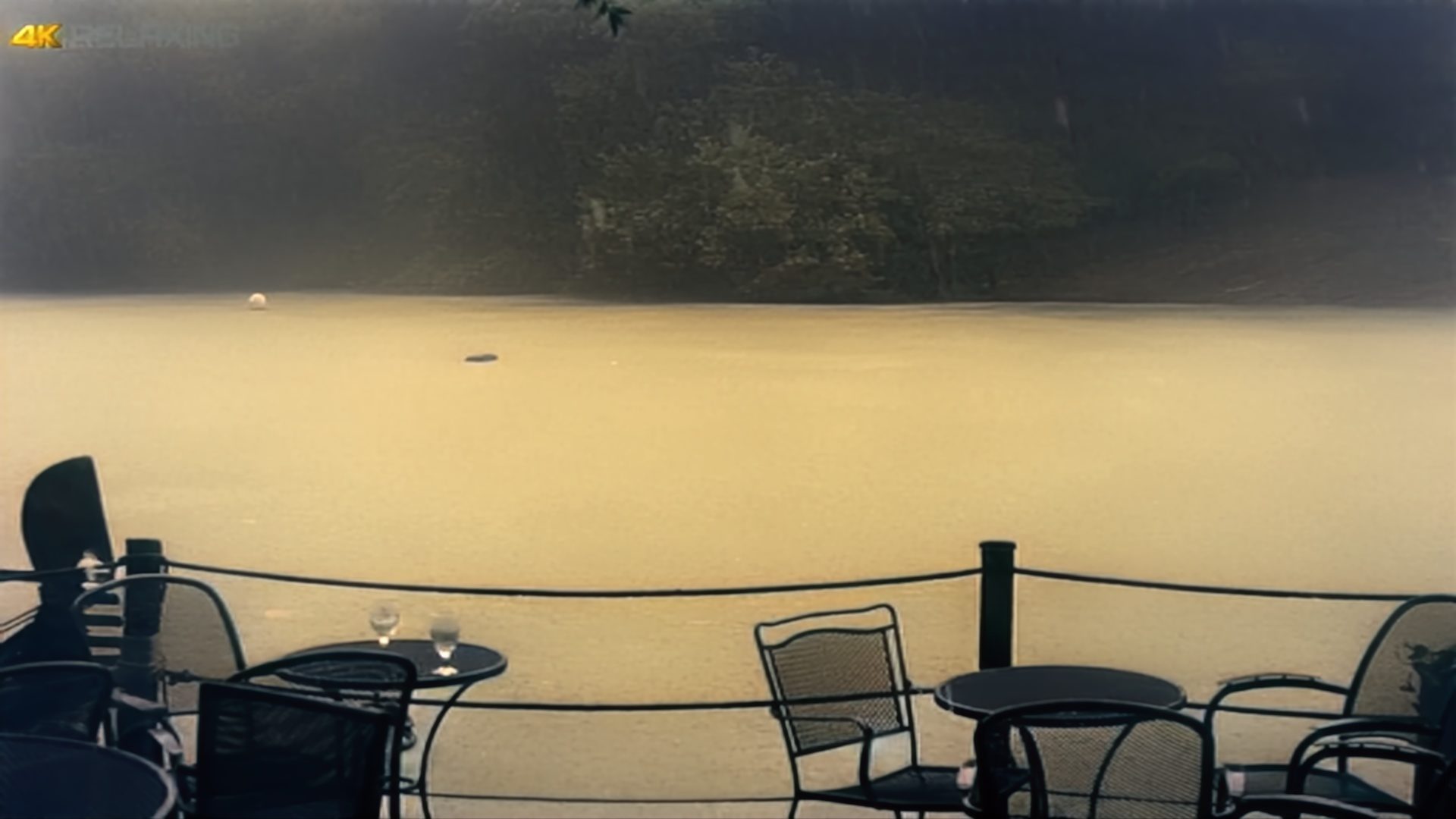}
    \includegraphics[width=0.13\textwidth]{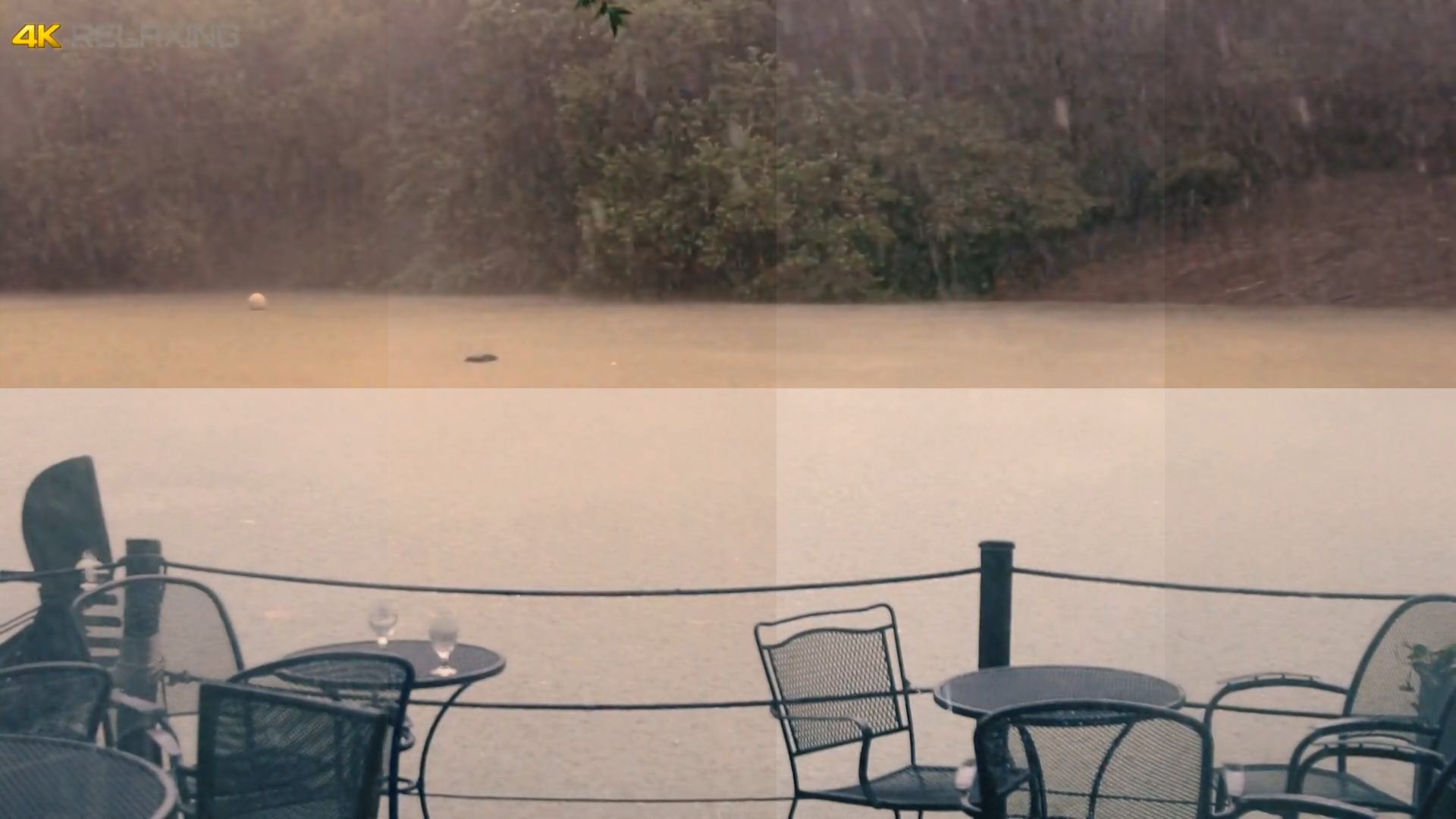}
    \includegraphics[width=0.13\textwidth]{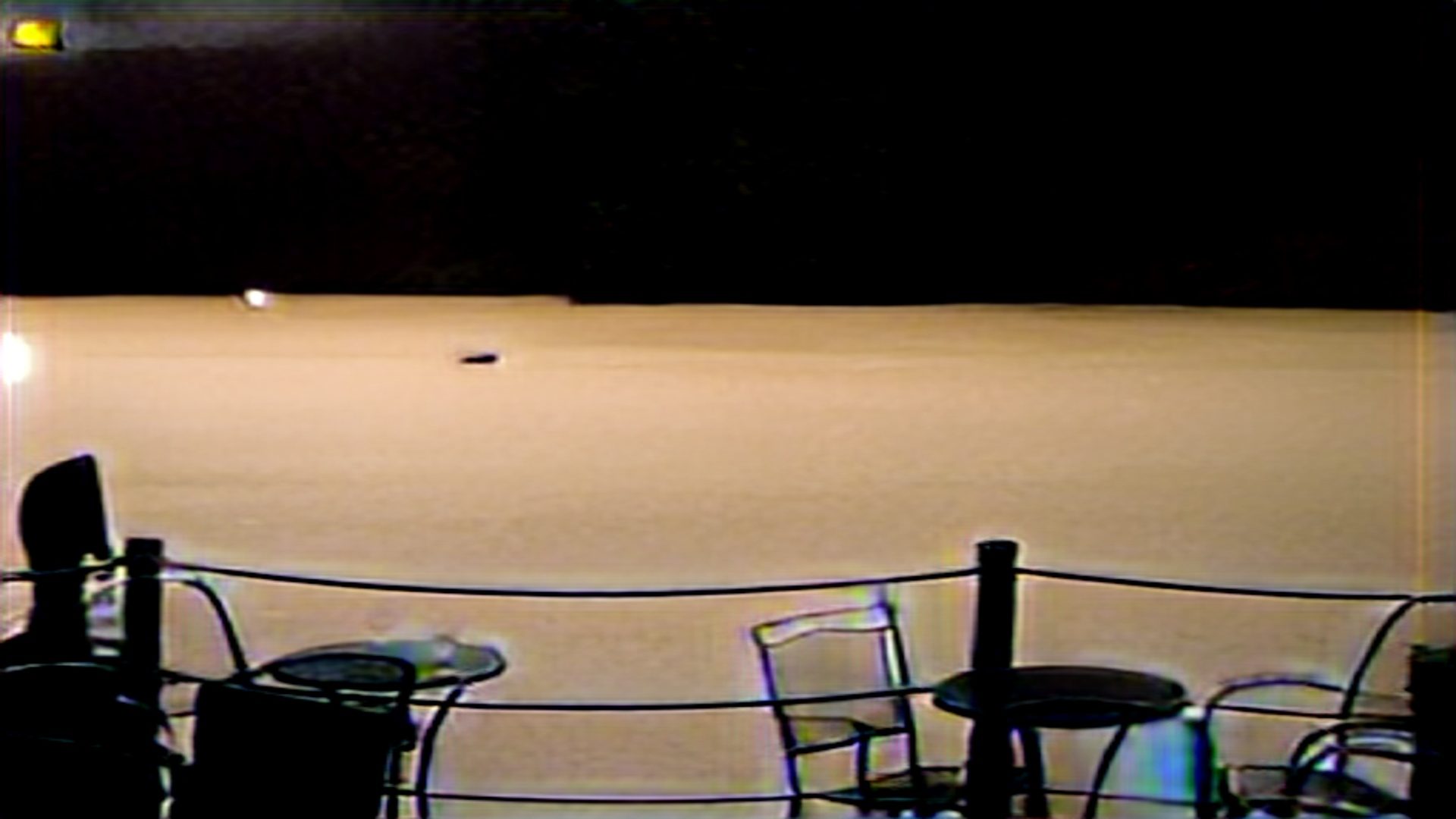}
    \includegraphics[width=0.13\textwidth]{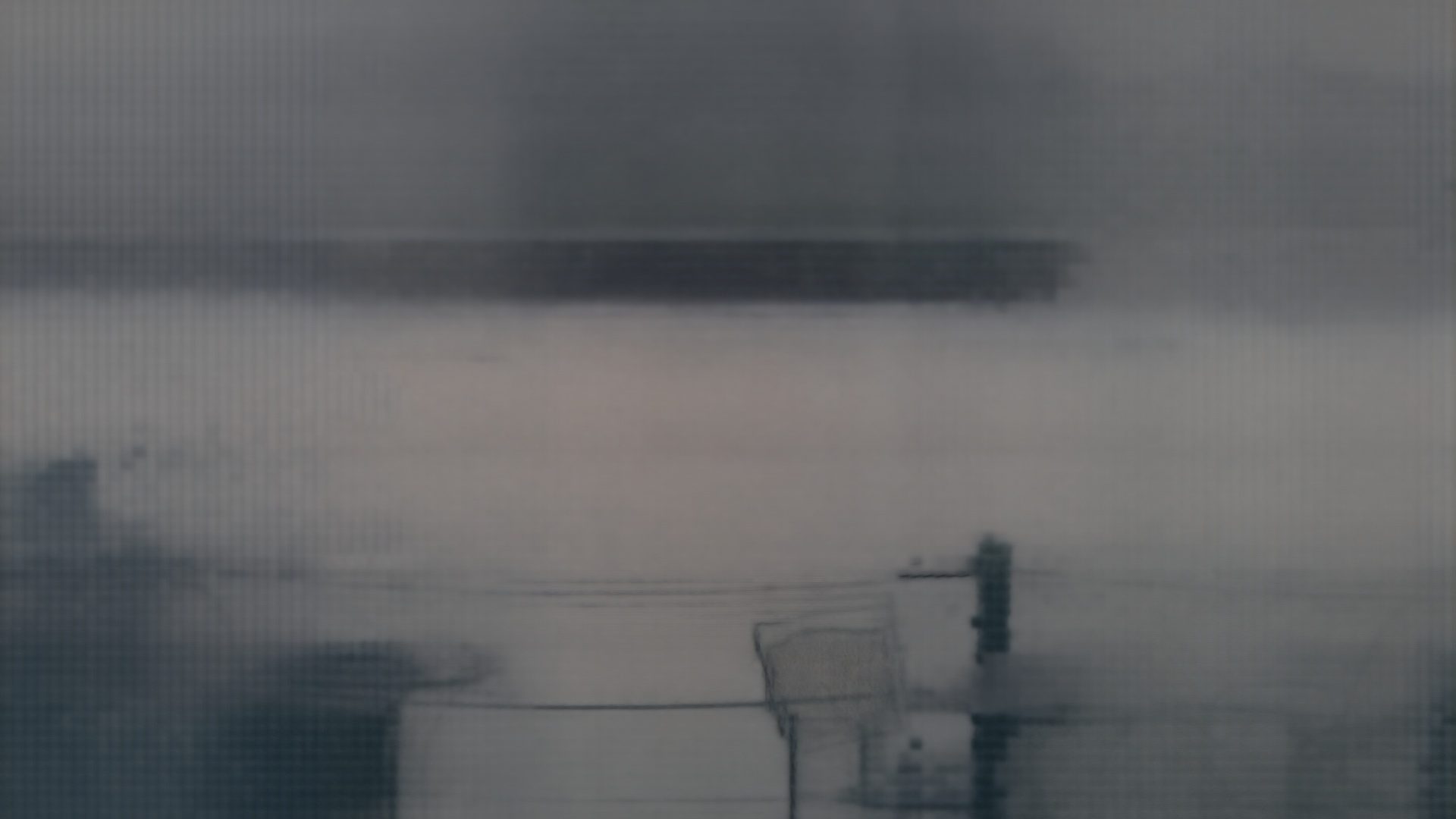}
    \includegraphics[width=0.13\textwidth]{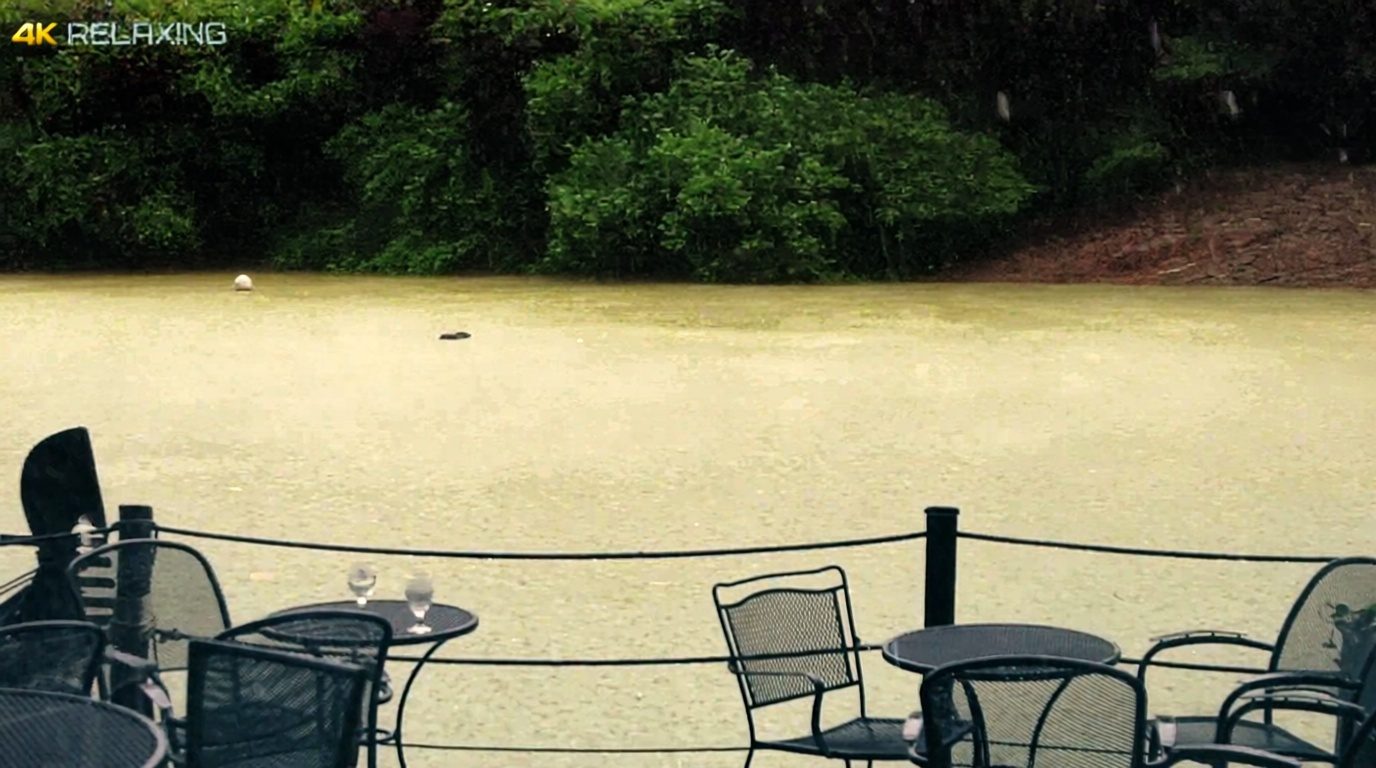}
    \includegraphics[width=0.13\textwidth]{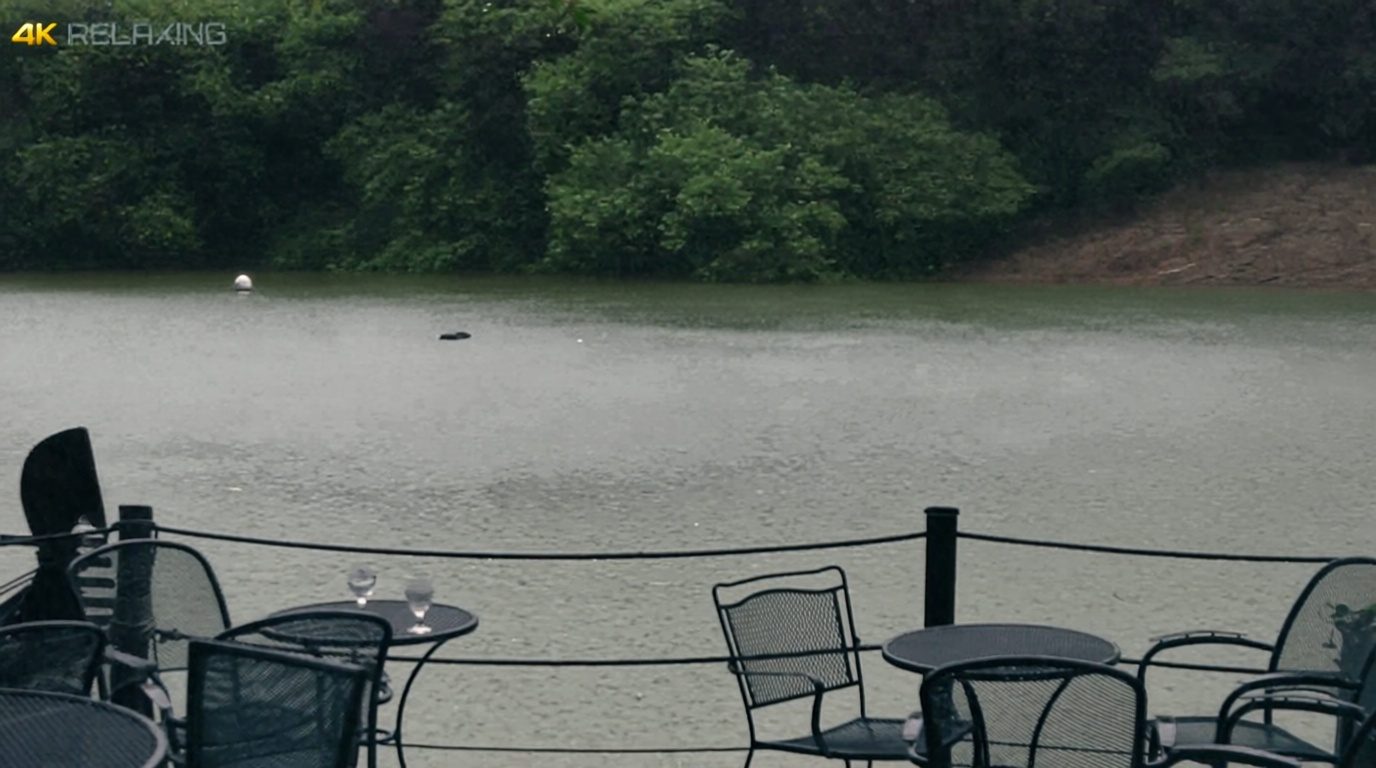}    

    \includegraphics[width=0.13\textwidth]{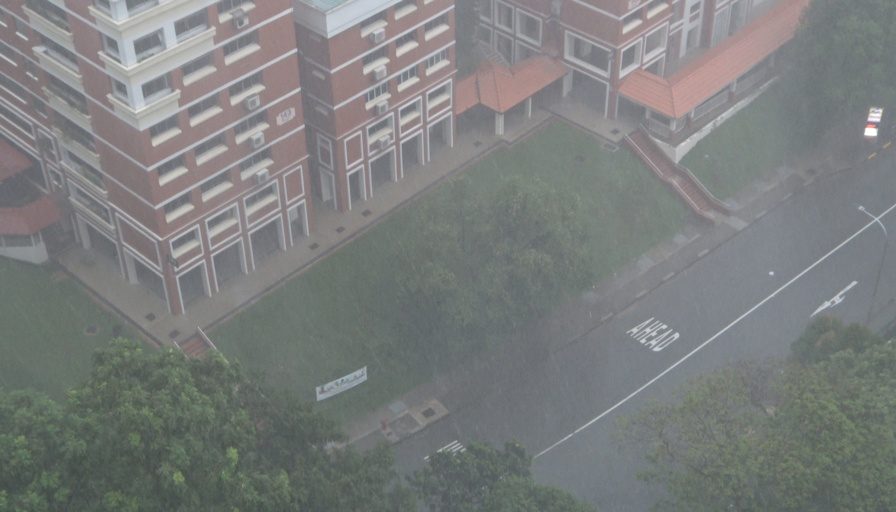}
    \includegraphics[width=0.13\textwidth]{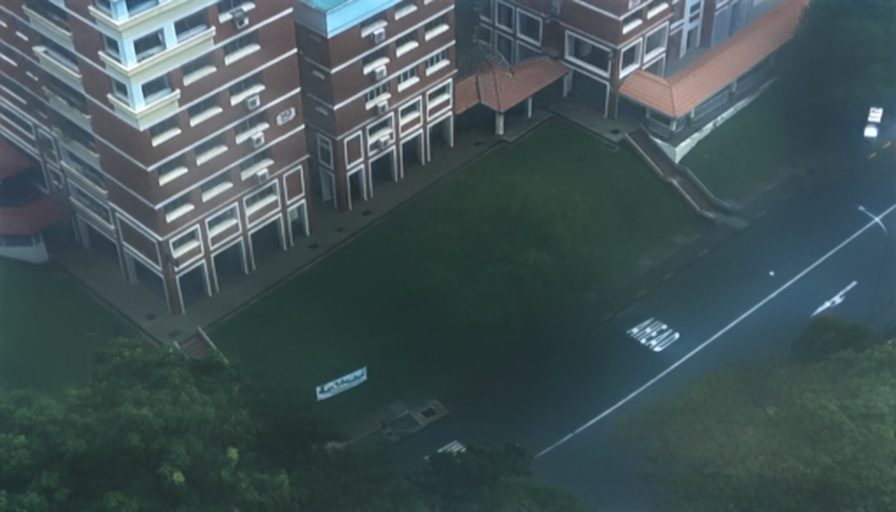}
    \includegraphics[width=0.13\textwidth]{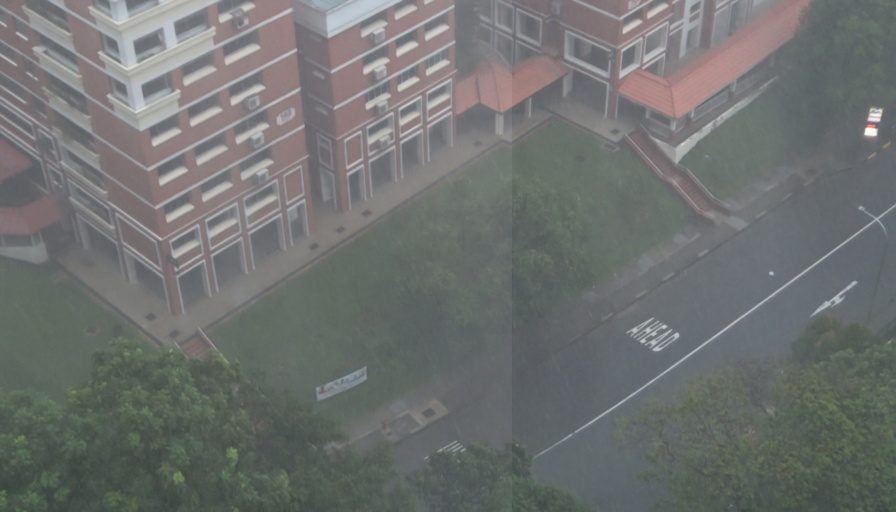}
    \includegraphics[width=0.13\textwidth]{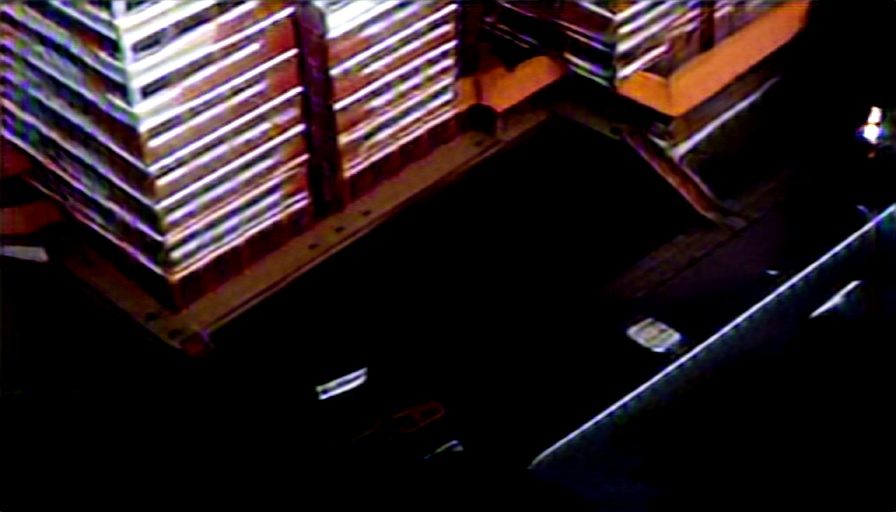}
    \includegraphics[width=0.13\textwidth]{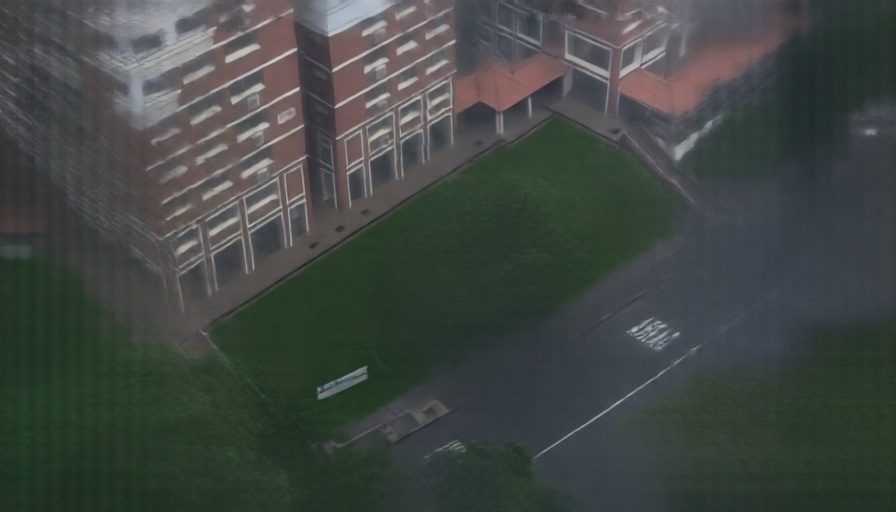}
    \includegraphics[width=0.13\textwidth]{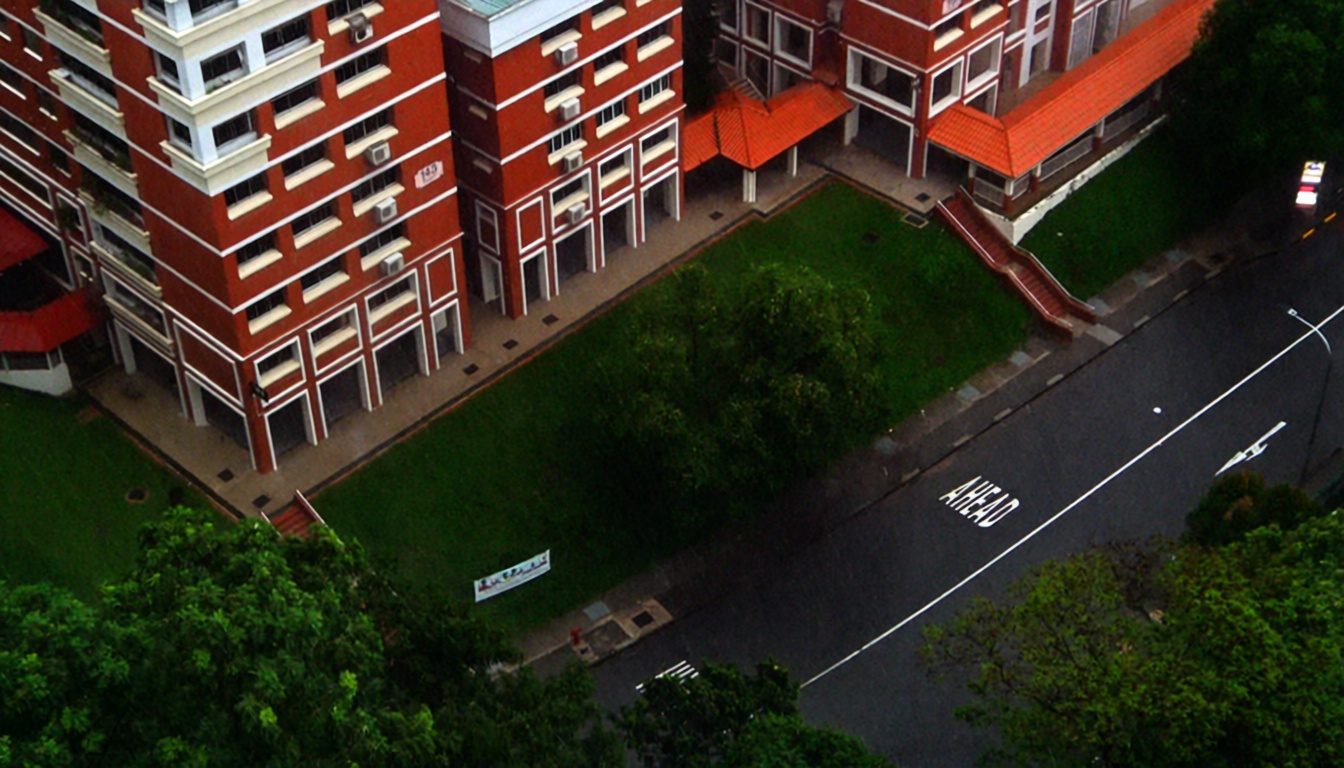}
    \includegraphics[width=0.13\textwidth]{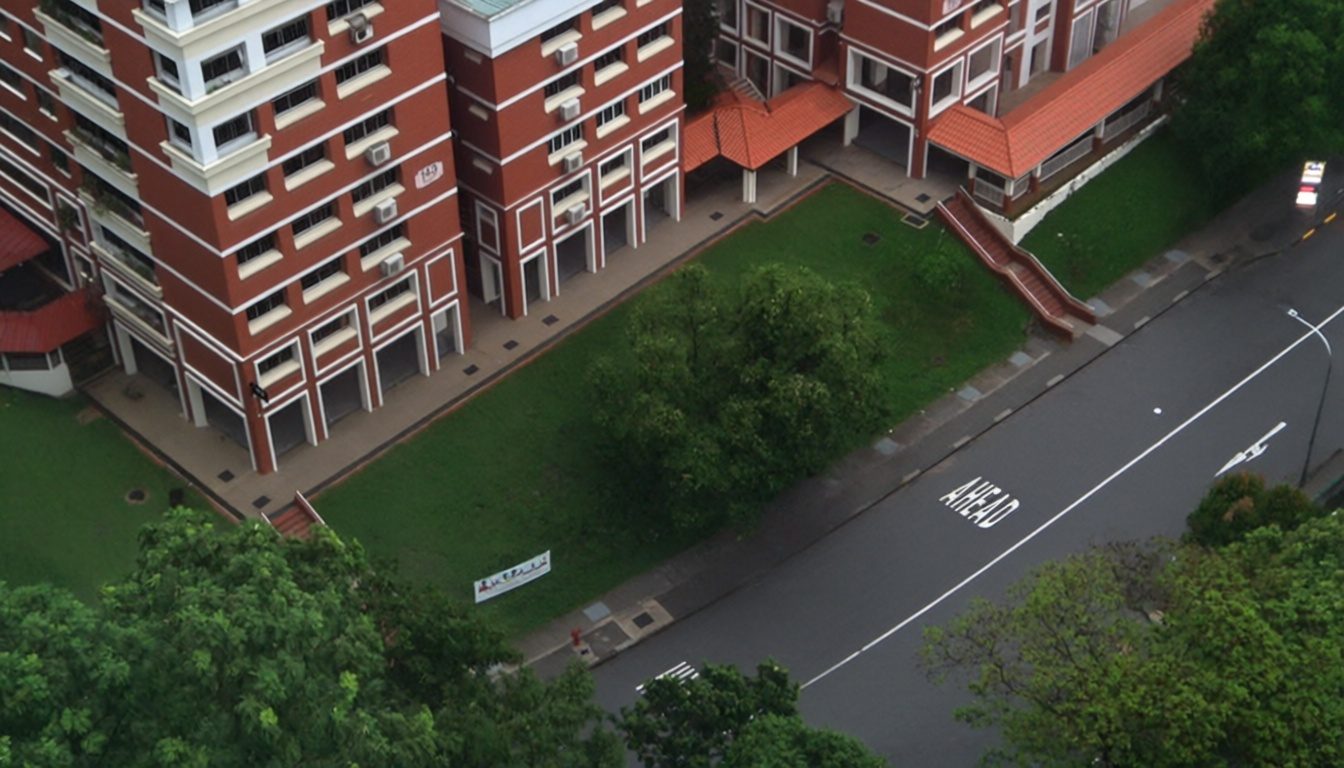}

    \includegraphics[width=0.13\textwidth]{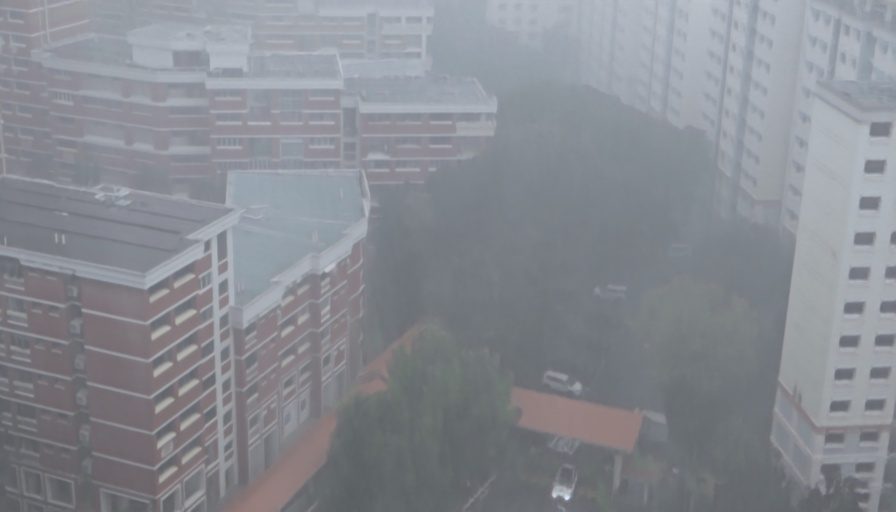}
    \includegraphics[width=0.13\textwidth]{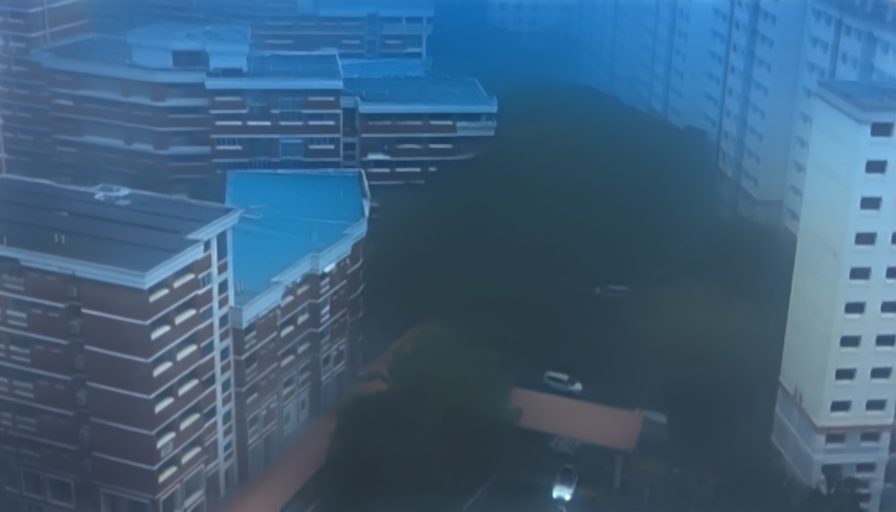}
    \includegraphics[width=0.13\textwidth]{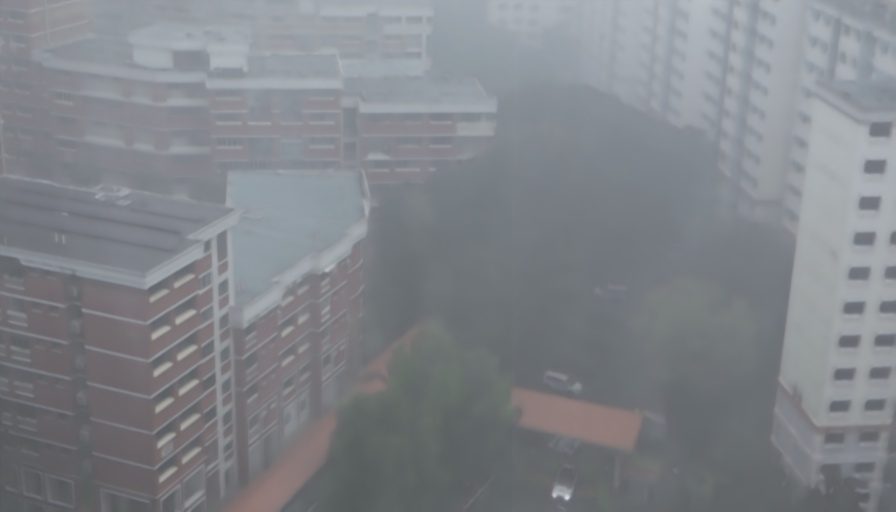}
    \includegraphics[width=0.13\textwidth]{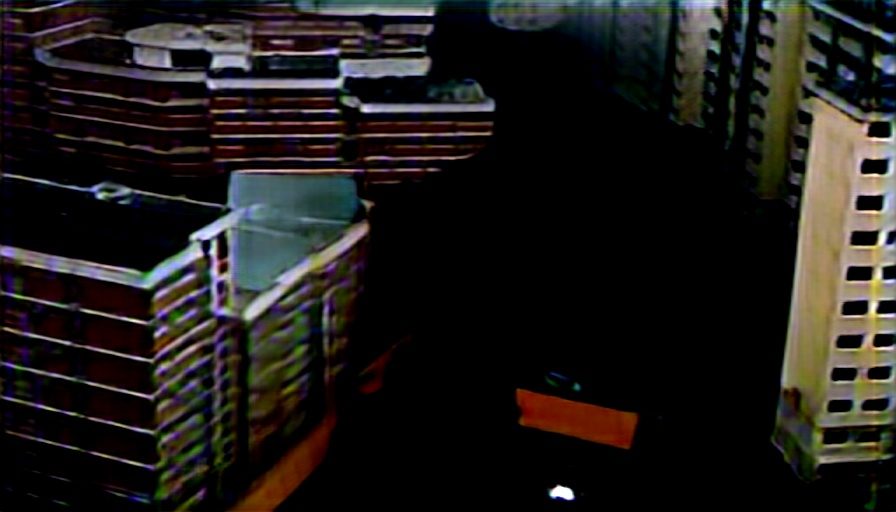}
    \includegraphics[width=0.13\textwidth]{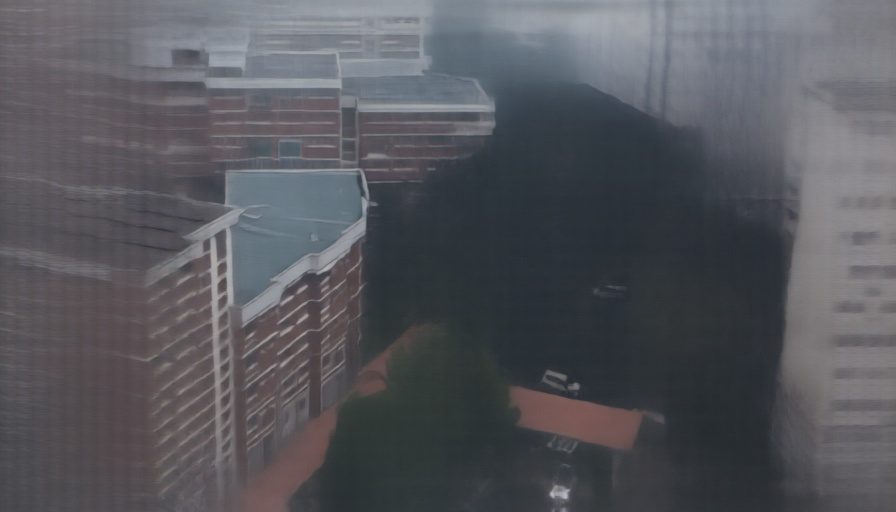}
    \includegraphics[width=0.13\textwidth]{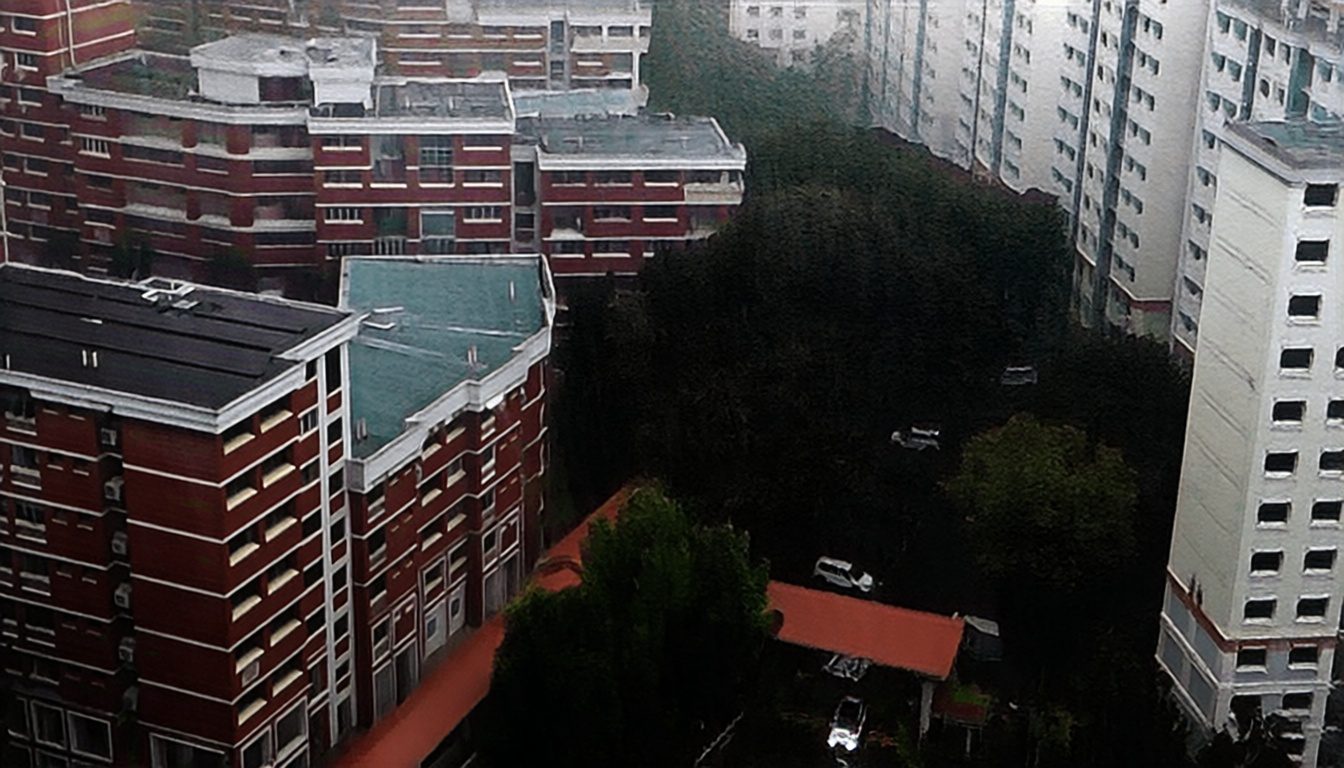}
    \includegraphics[width=0.13\textwidth]{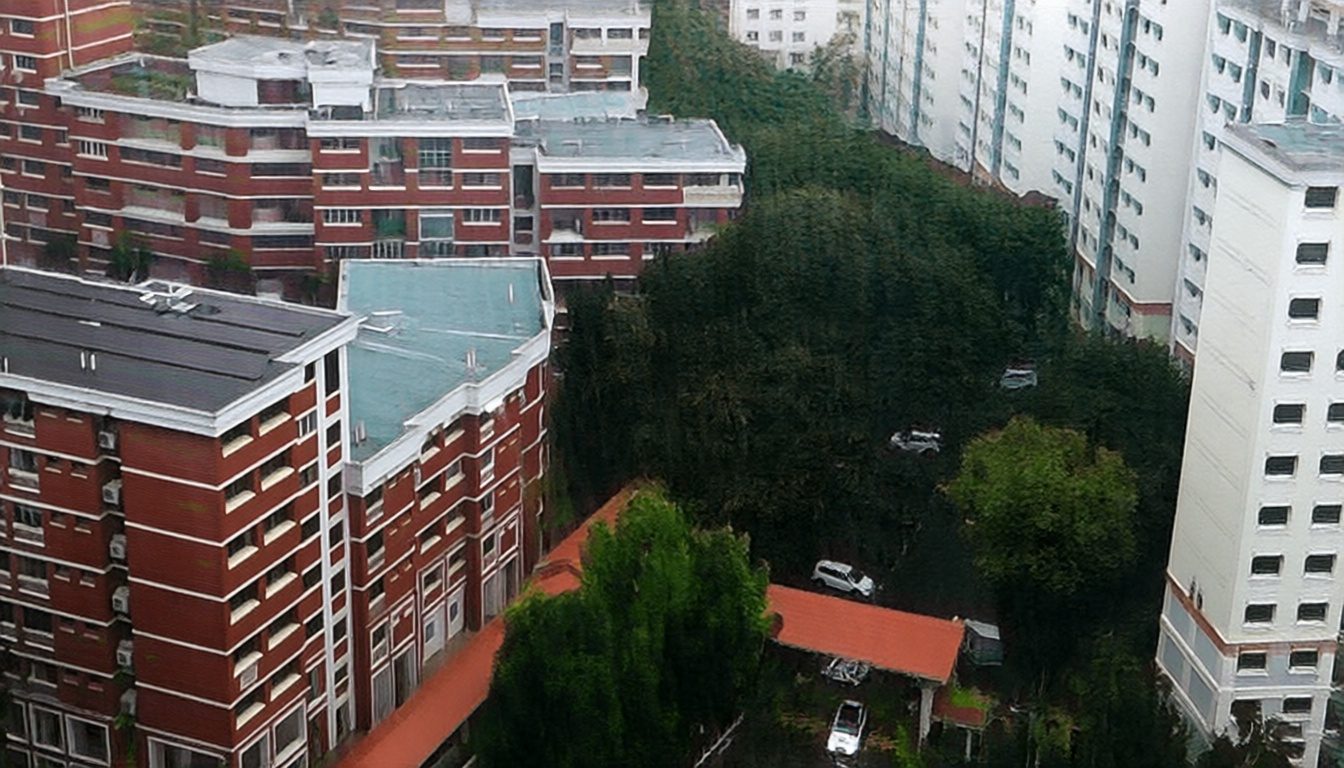}
    
    \includegraphics[width=0.13\textwidth]{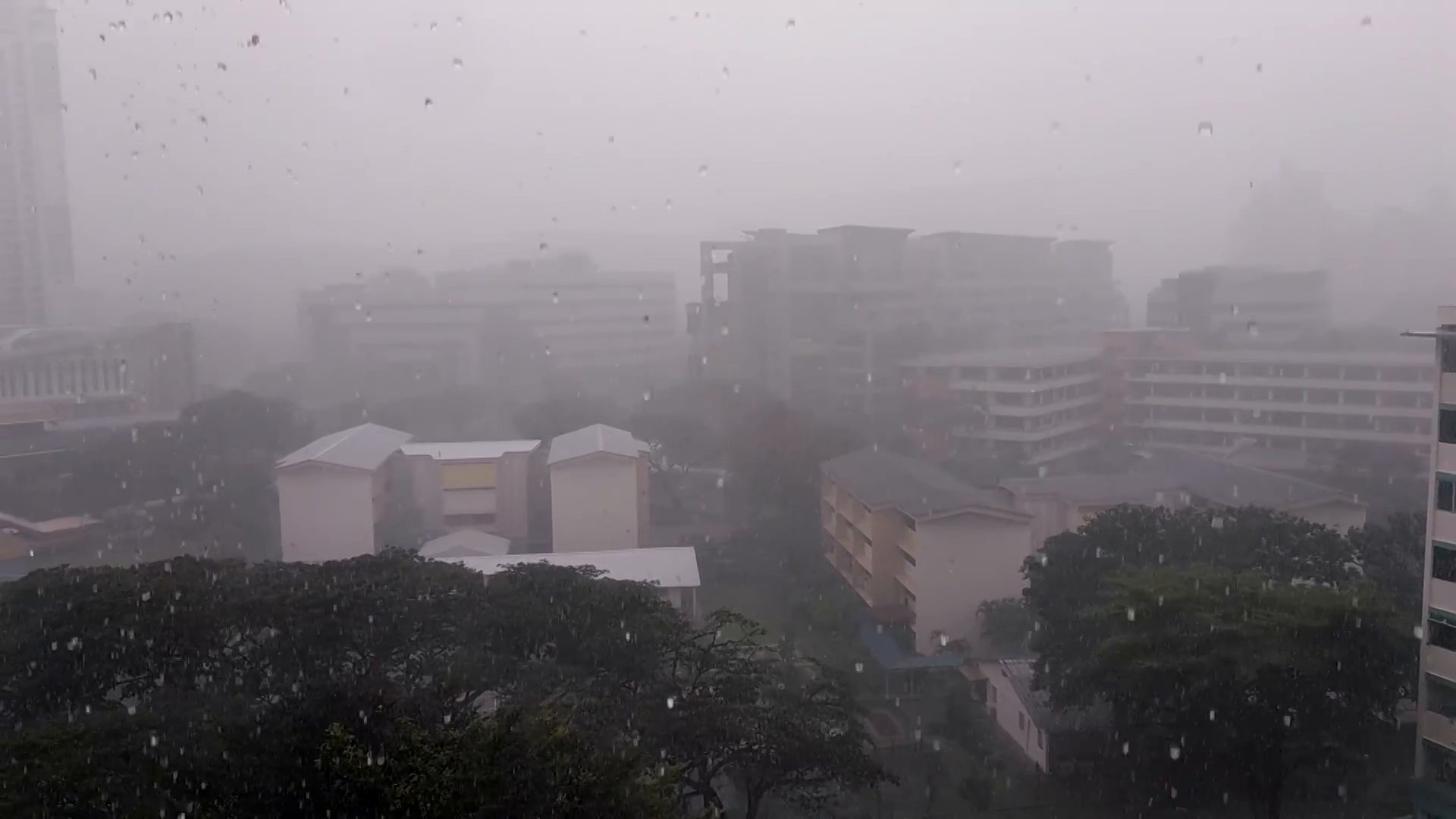}
    \includegraphics[width=0.13\textwidth]{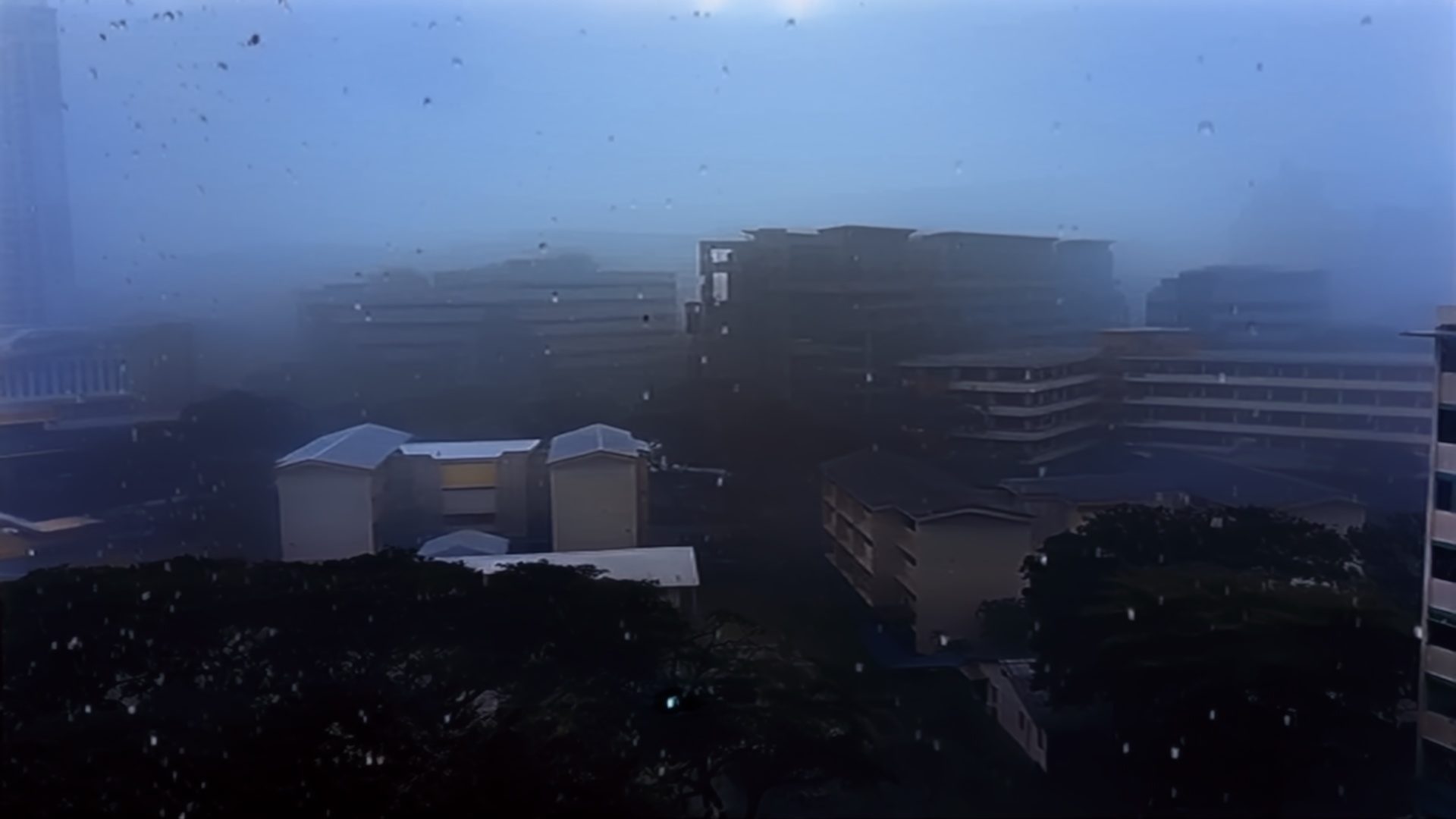}
    \includegraphics[width=0.13\textwidth]{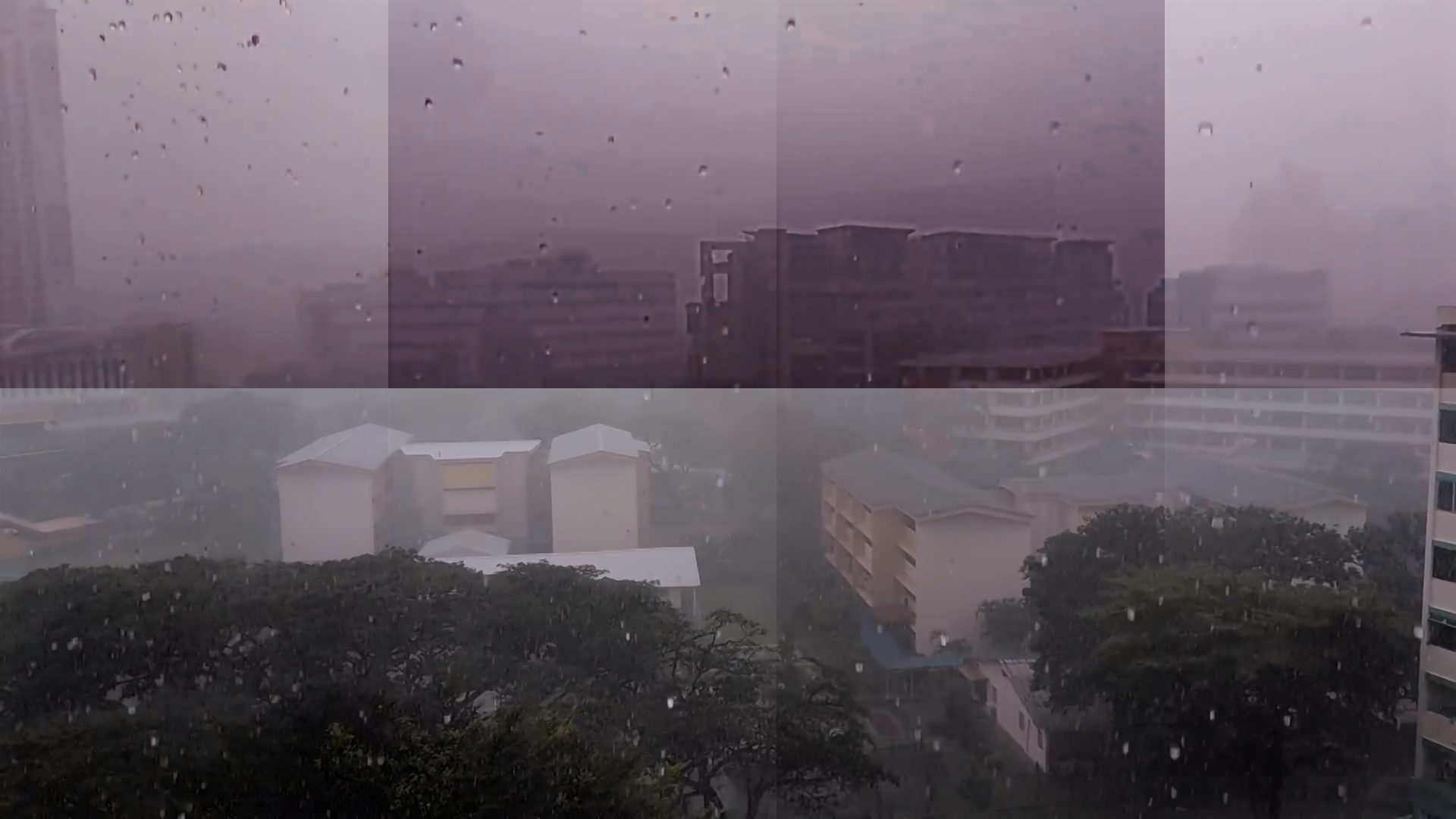}
    \includegraphics[width=0.13\textwidth]{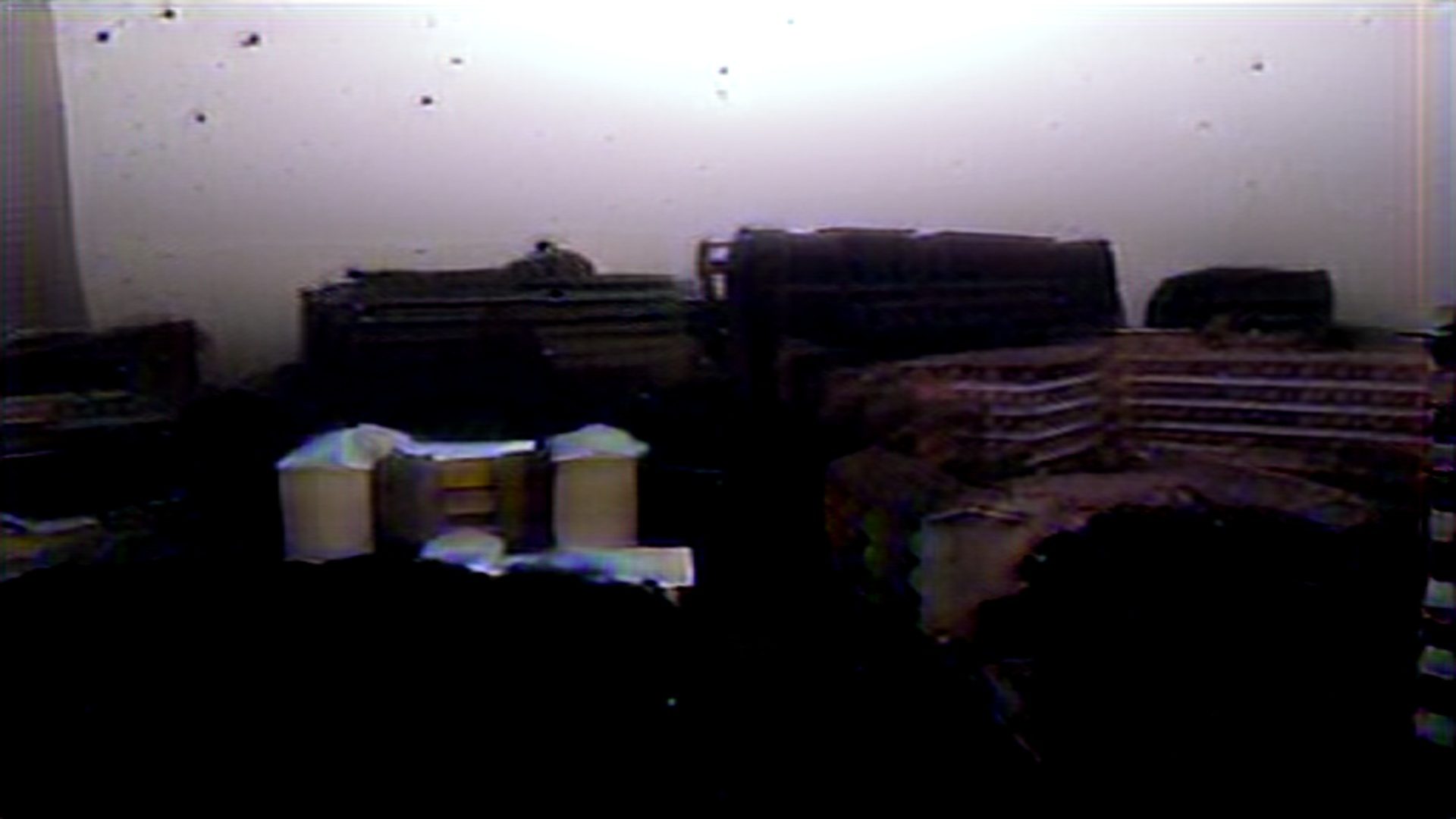}
    \includegraphics[width=0.13\textwidth]{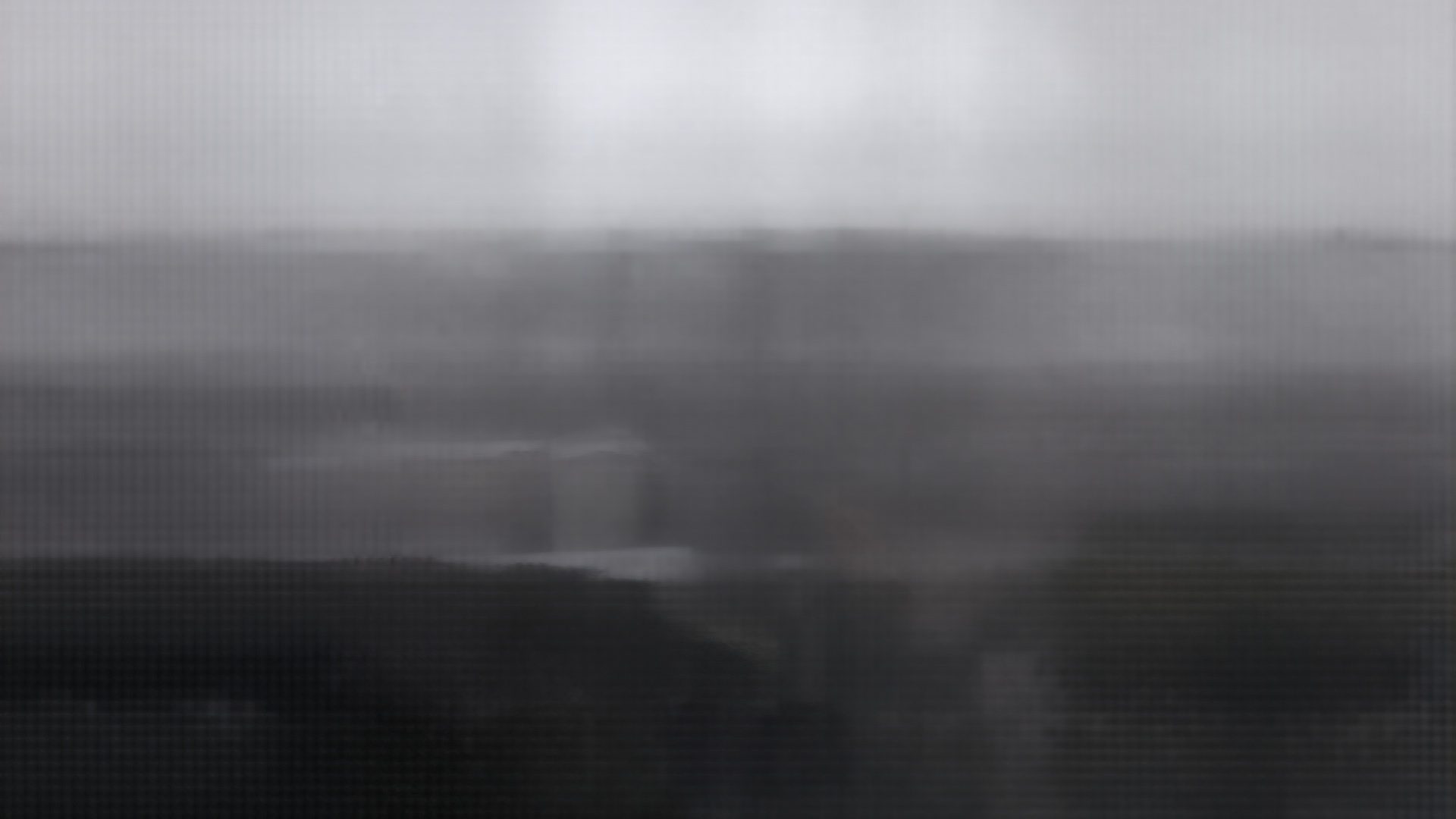}
    \includegraphics[width=0.13\textwidth]{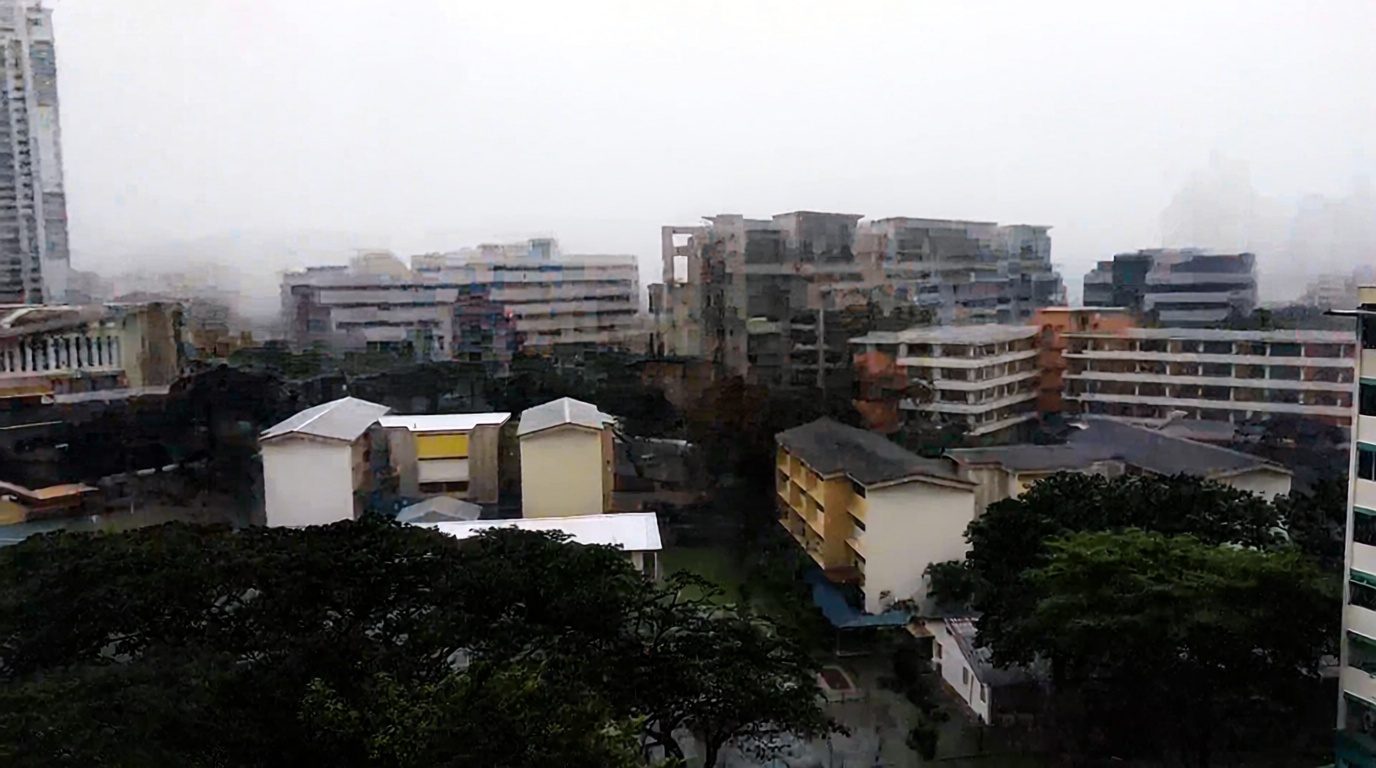}
    \includegraphics[width=0.13\textwidth]{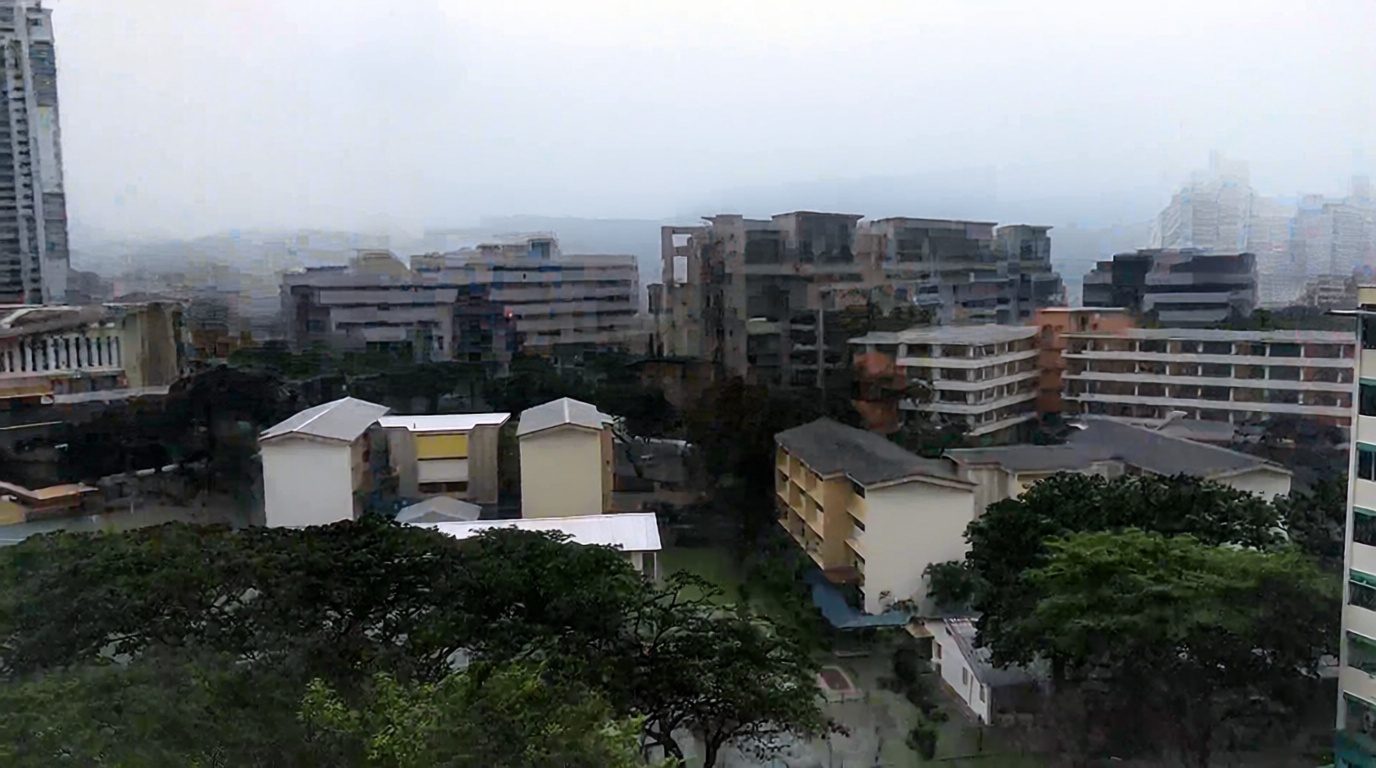}

    \includegraphics[width=0.13\textwidth]{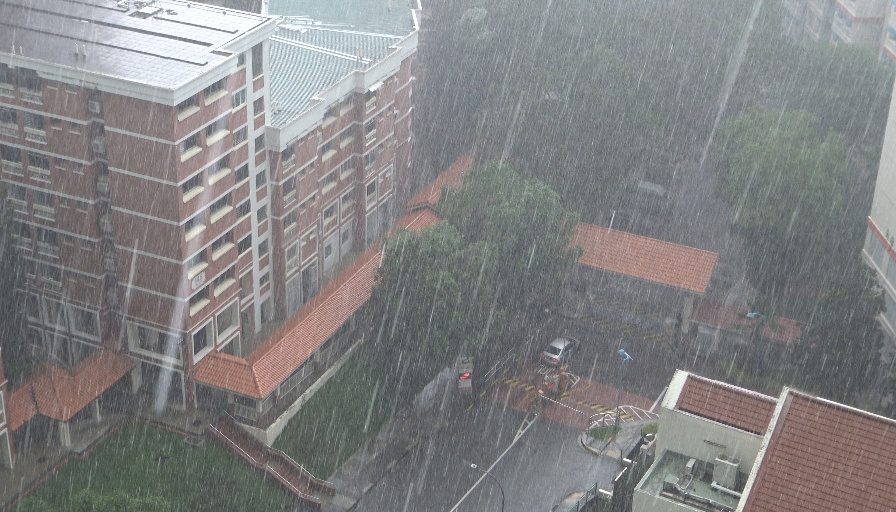}
    \includegraphics[width=0.13\textwidth]{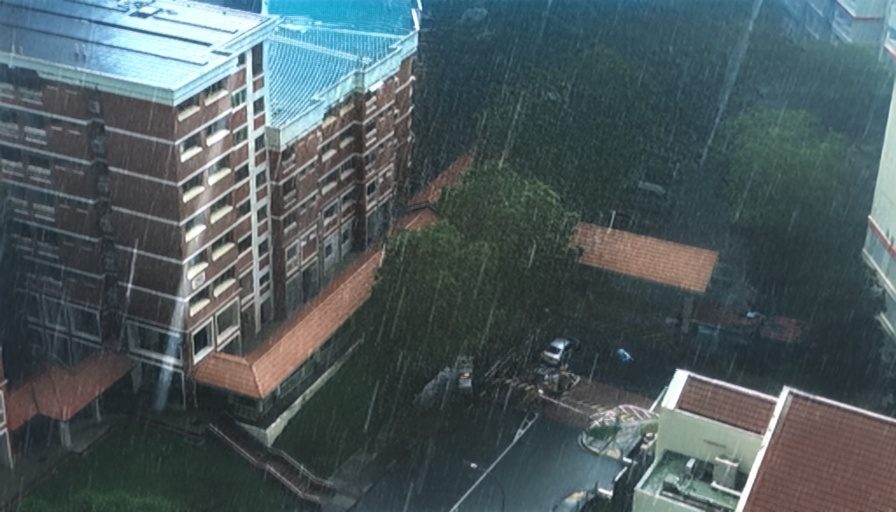}
    \includegraphics[width=0.13\textwidth]{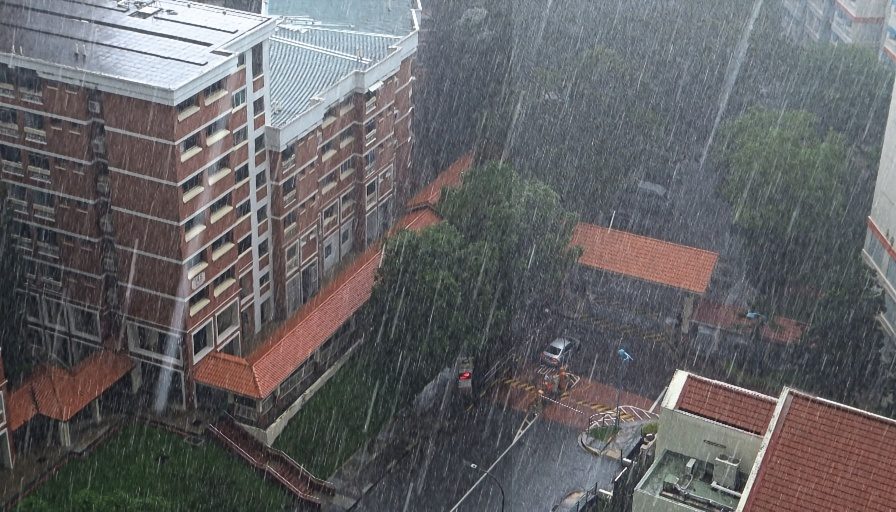}
    \includegraphics[width=0.13\textwidth]{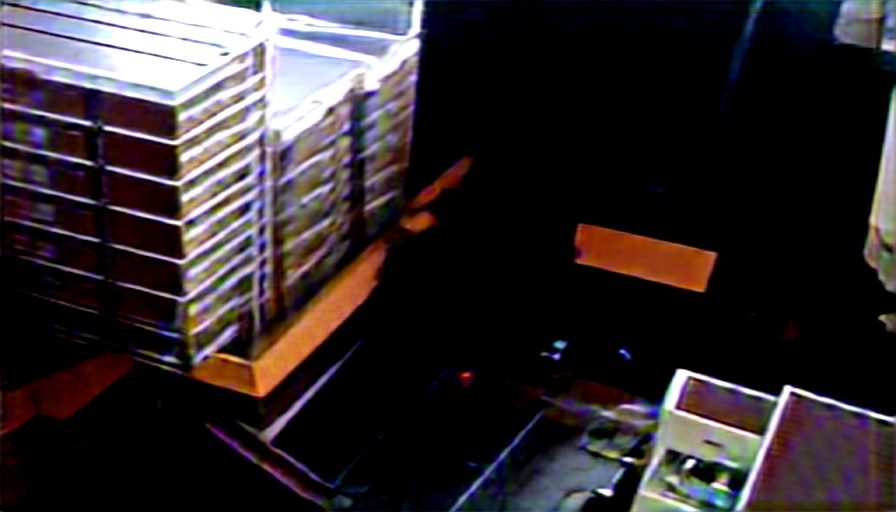}
    \includegraphics[width=0.13\textwidth]{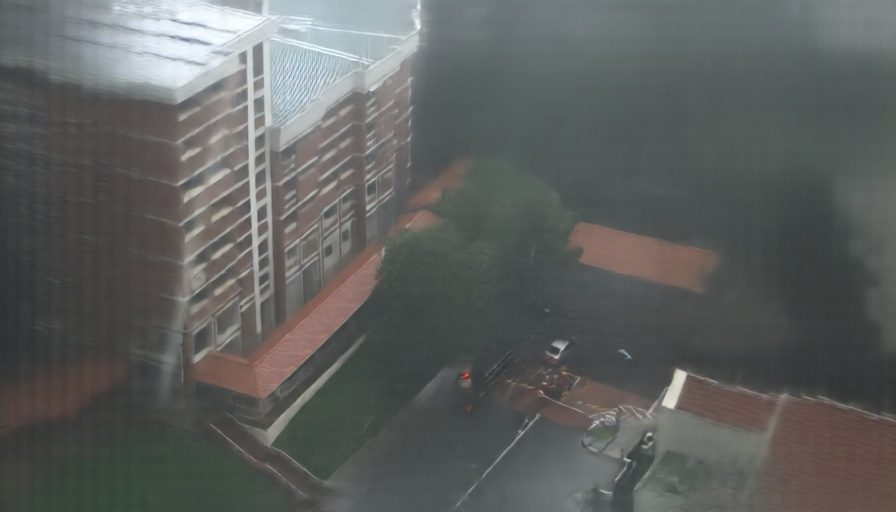}
    \includegraphics[width=0.13\textwidth]{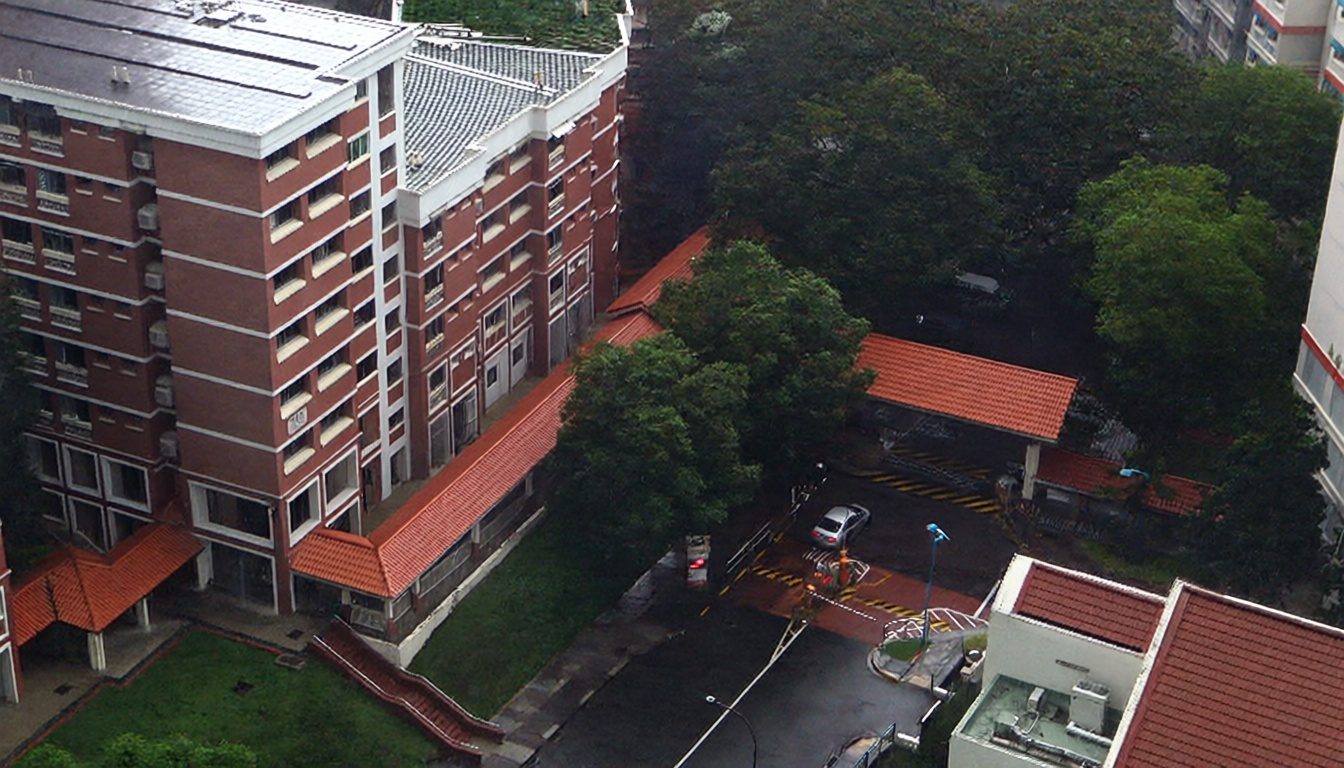}
    \includegraphics[width=0.13\textwidth]{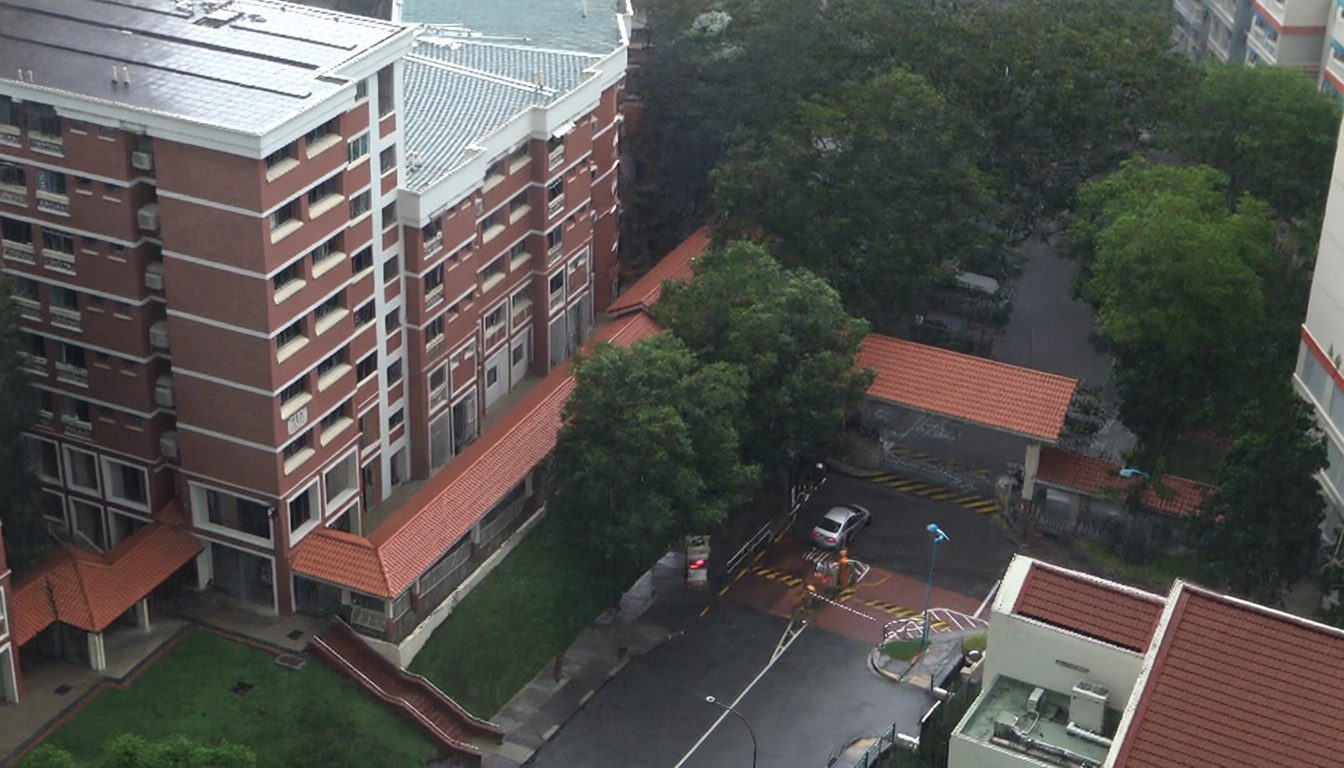}
    
    \caption{Qualitative comparison on real-world test set. }
    \label{fig:real_visual}
\end{figure*}

\cref{fig:ots_visual} and \cref{fig:real_visual} shows qualitative comparison on, outdoor dataset, indoor dataset and real-world dataset. 
The conventional restoration pipeline, which consists of three strong single-task restoration models for Derain/Deblur/Denoise~\cite{parmar2024one,chu2022nafssr} is denoted as 3D.\footnote{Restoration orders are aligned with the ground truth for synthetic datasets, which should be optimal for agentic IR. For real-world datasets, all degradations are assumed occurred and restored in the reverse order based on the proposed degradation model.} 3D and FoundIR~\cite{foundir2025} show similar performance. They remove degradation in some scenario, but fails when the degradation is severe, where the algorithms cannot effectively restore the objects' structures and textures. Stable-Diffusion-3~\cite{esser2024scaling} improves the overall quality of generation, but often causes structure drift and excessive smoothing, where thin branches or building edges are hallucinated or misaligned. In some hard cases in \cref{fig:real_visual}, it cannot remove the fog and even hallucinates non-existing structures. Img2Img-Turbo~\cite{parmar2024one} further improves the overall quality of the generation but fails in extreme situations such as extreme fog in \cref{fig:ots_visual} and noise in \cref{fig:real_visual}. Qwen3-Image~\cite{wu2025qwen} better preserves sharpness, but still losses some color-consistency when extreme fog or noise occurs. In contrast, the proposed full R\&R framework consistently delivers cleaner restoration with stable geometric structures and colors closer to the ground truth across diverse scenes. 
\begin{figure*}
    \centering
    \begin{subfigure}{0.12\textwidth}
        \centering
        \textbf{Input}
    \end{subfigure}
    \begin{subfigure}{0.12\textwidth}
        \centering
        \textbf{w/o Reasoner SFT}
    \end{subfigure}
    \begin{subfigure}{0.12\textwidth}
        \centering
        \textbf{w/o Score}
    \end{subfigure}
    \begin{subfigure}{0.12\textwidth}
        \centering
        \textbf{w/o Parameters}
    \end{subfigure}
    \begin{subfigure}{0.12\textwidth}
        \centering
        \textbf{w/o Semantics}
    \end{subfigure}
    \begin{subfigure}{0.12\textwidth}
        \centering
        \textbf{w/o RL}
    \end{subfigure}
    \begin{subfigure}{0.12\textwidth}
        \centering
        \textbf{Full Model}
    \end{subfigure}
    \begin{subfigure}{0.12\textwidth}
        \centering
        \textbf{GT}
    \end{subfigure}

    \includegraphics[width=0.12\textwidth]{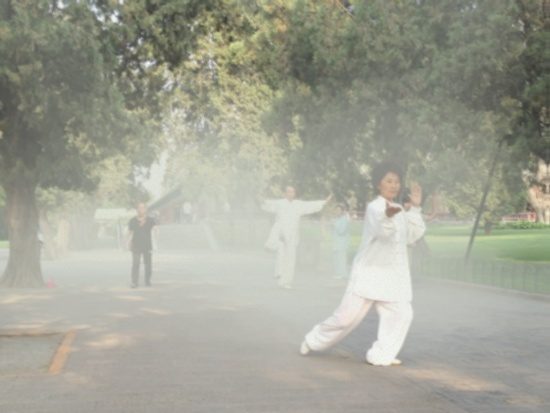}
    \includegraphics[width=0.12\textwidth]{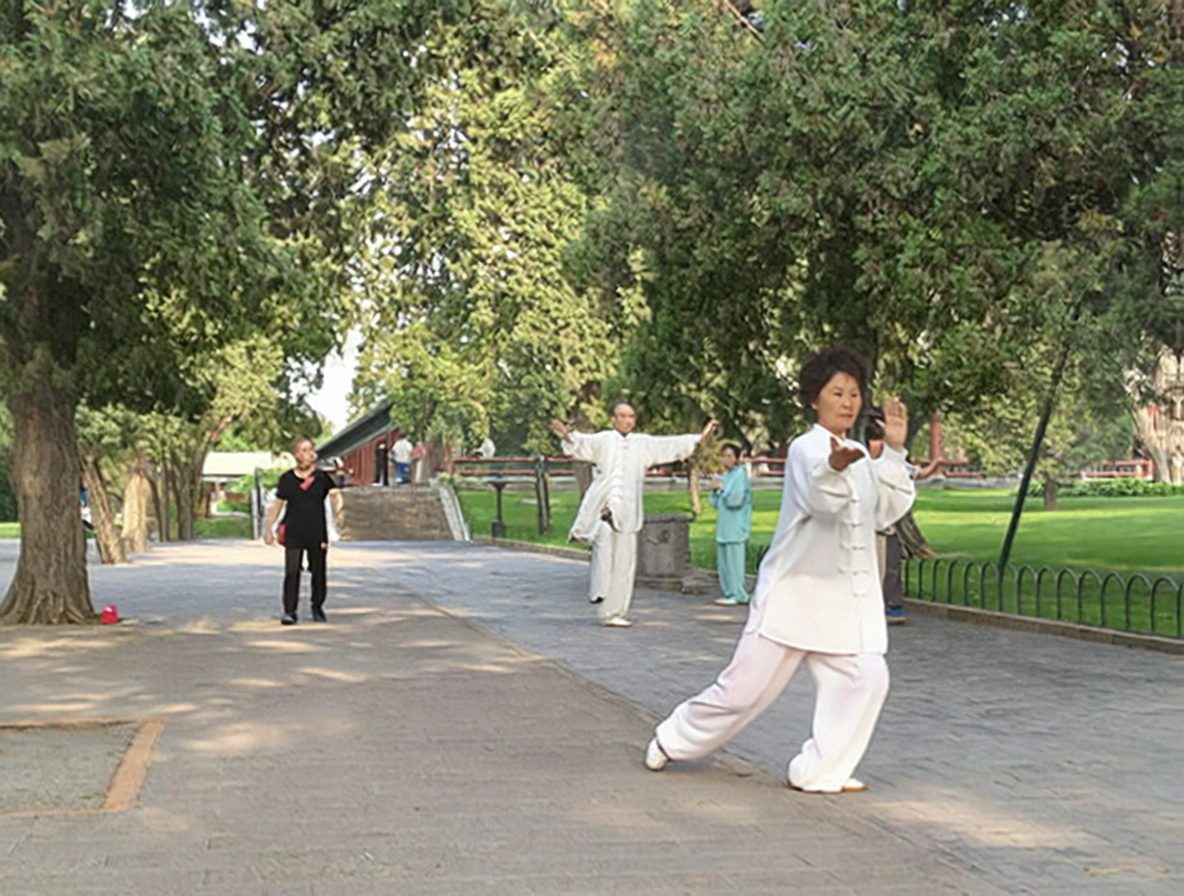}
    \includegraphics[width=0.12\textwidth]{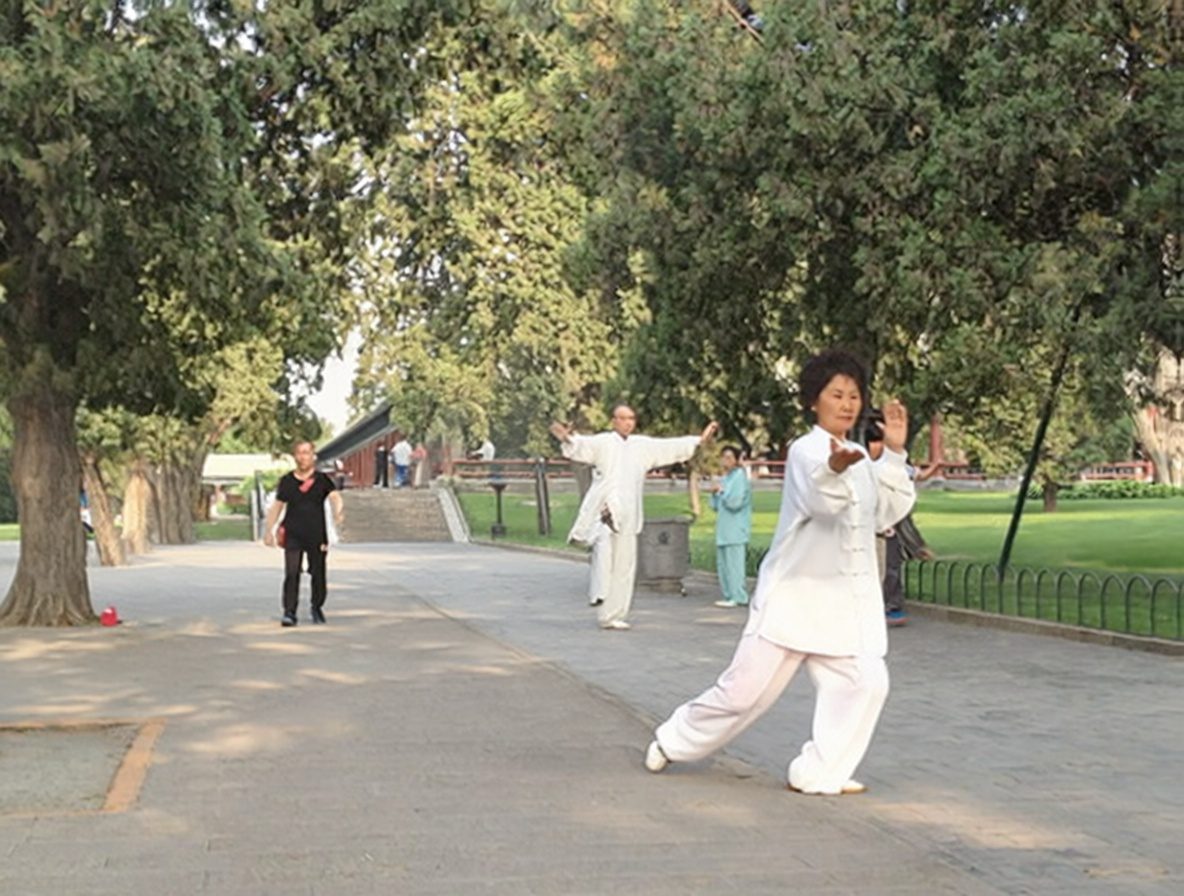}
    \includegraphics[width=0.12\textwidth]{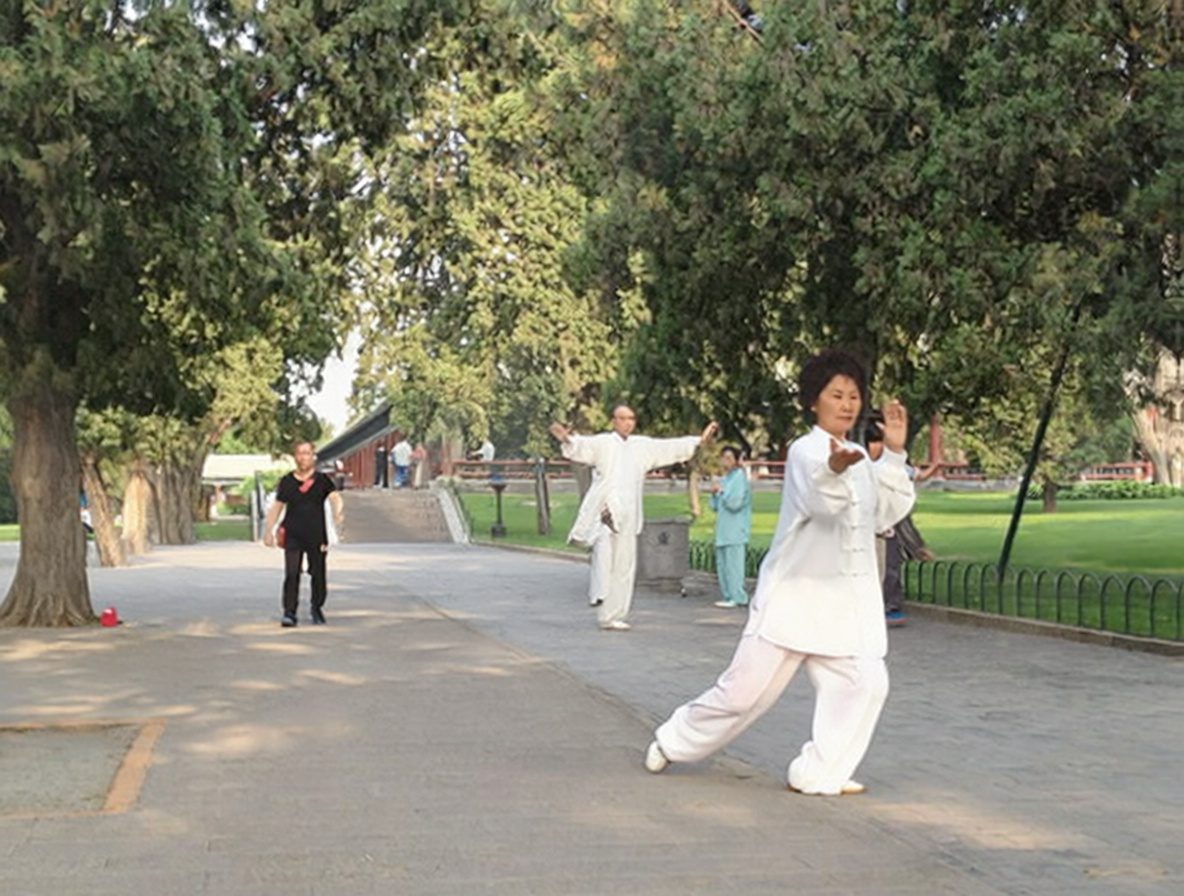}
    \includegraphics[width=0.12\textwidth]{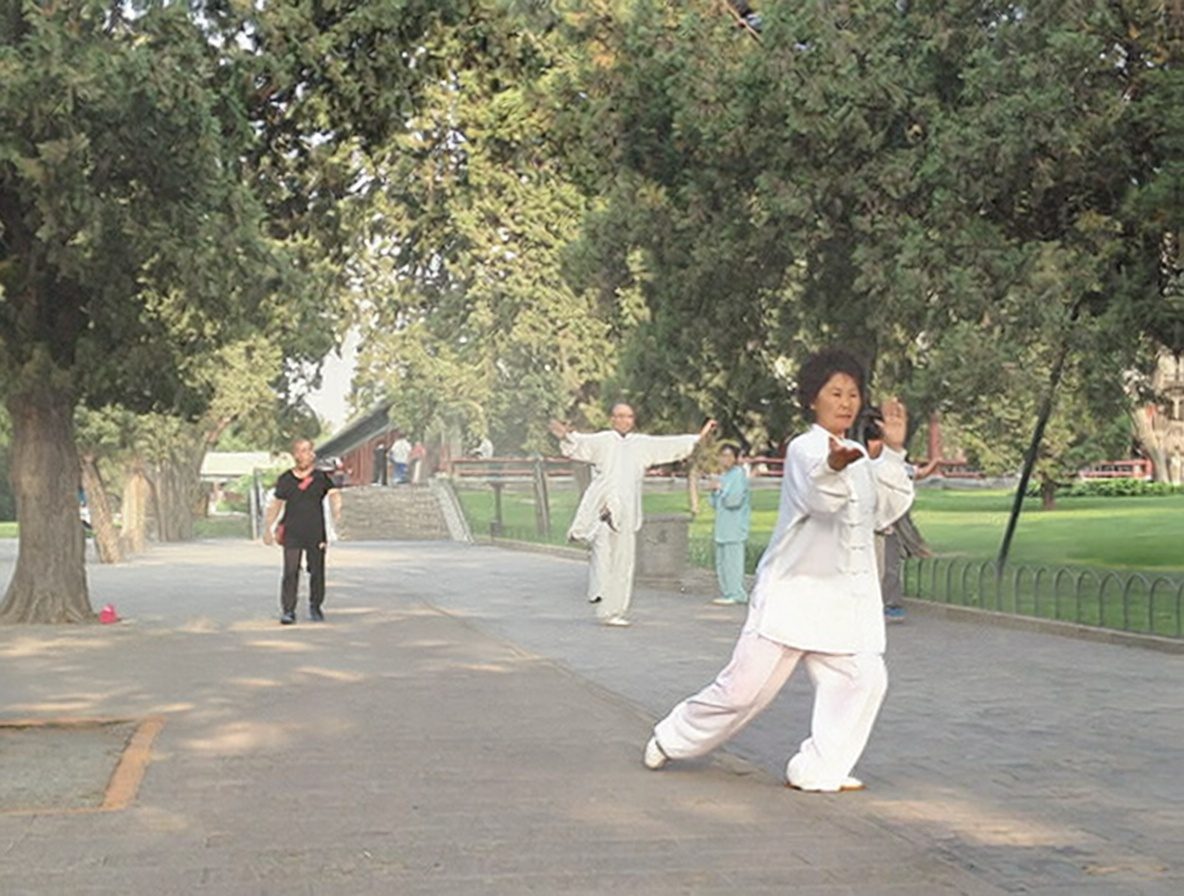}
    \includegraphics[width=0.12\textwidth]{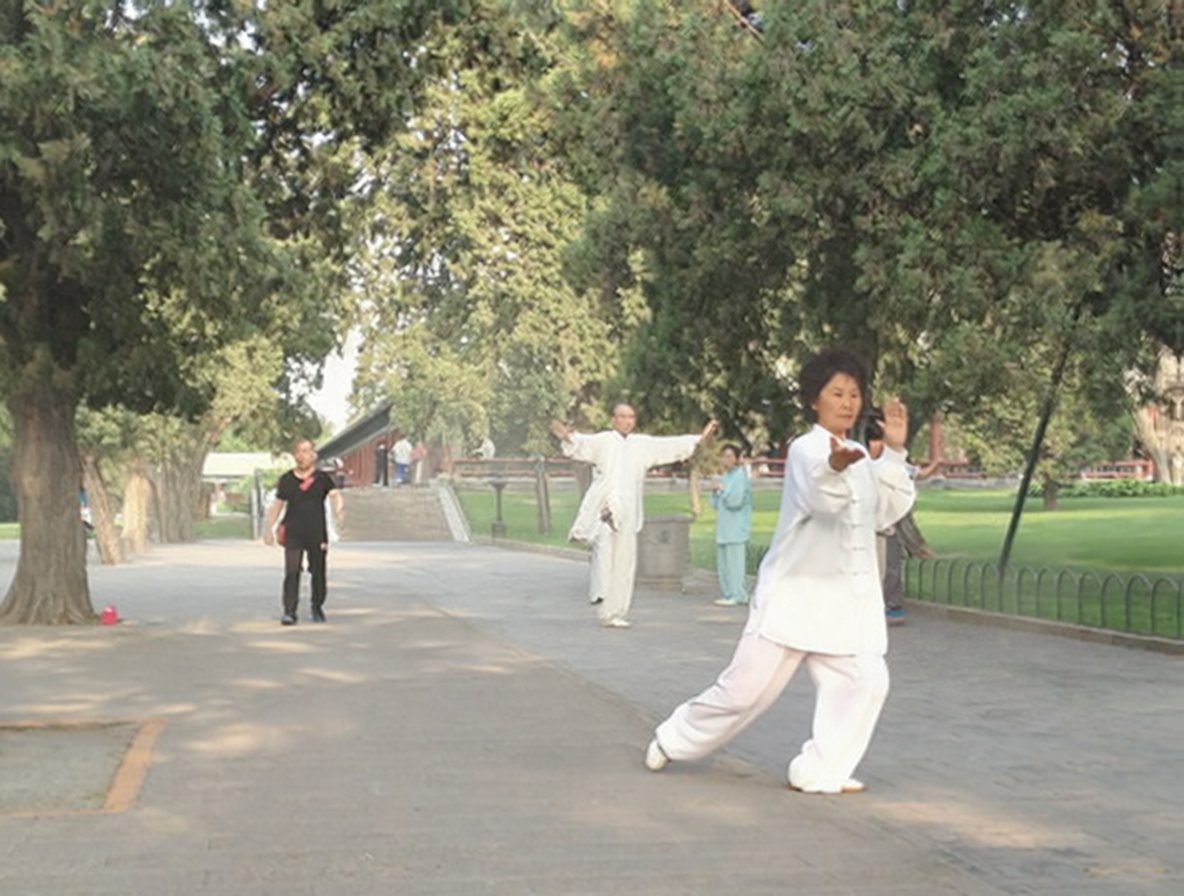}
    \includegraphics[width=0.12\textwidth]{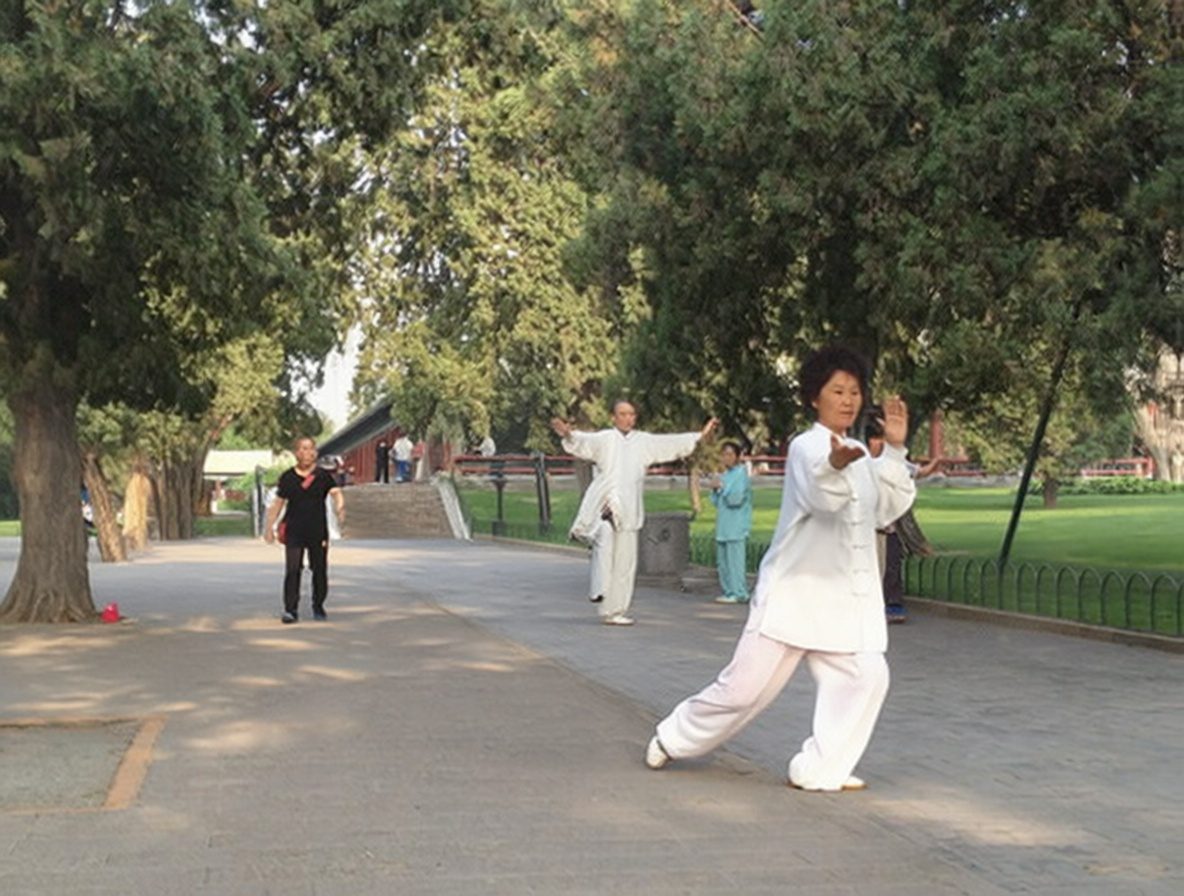}
    \includegraphics[width=0.12\textwidth]{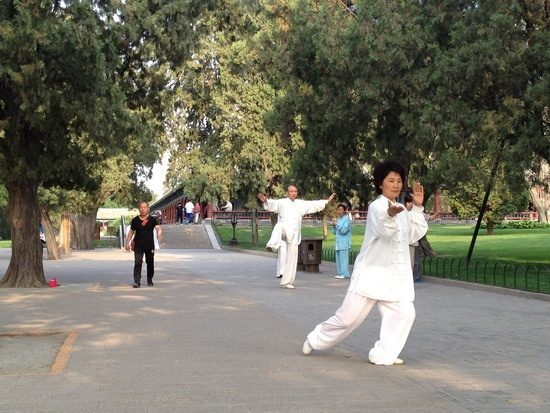}


    \includegraphics[width=0.12\textwidth]{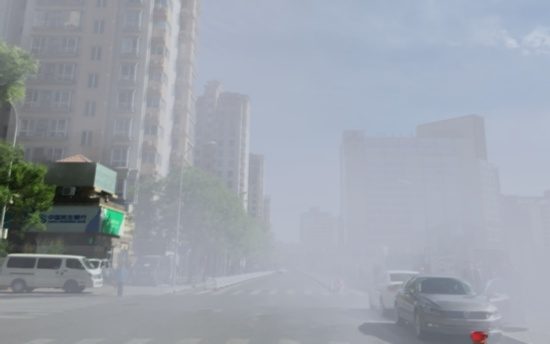}
    \includegraphics[width=0.12\textwidth]{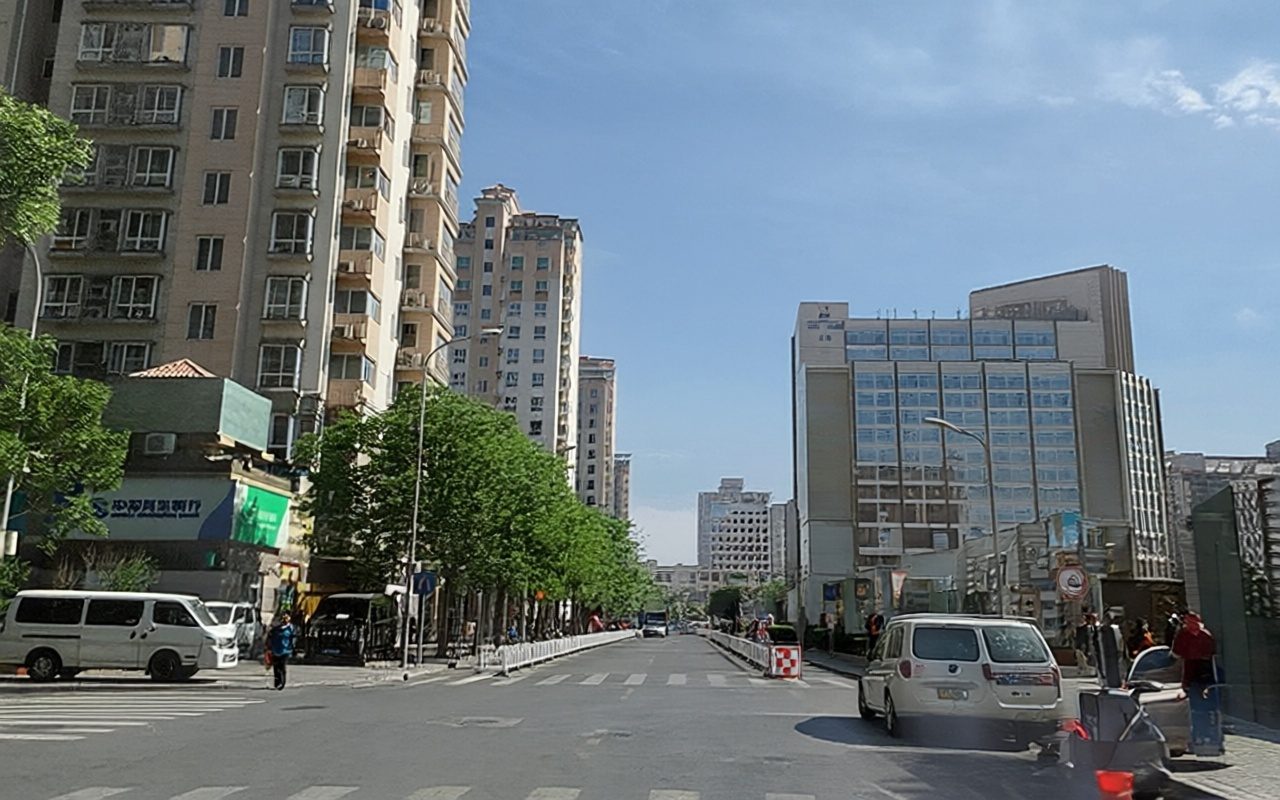}
    \includegraphics[width=0.12\textwidth]{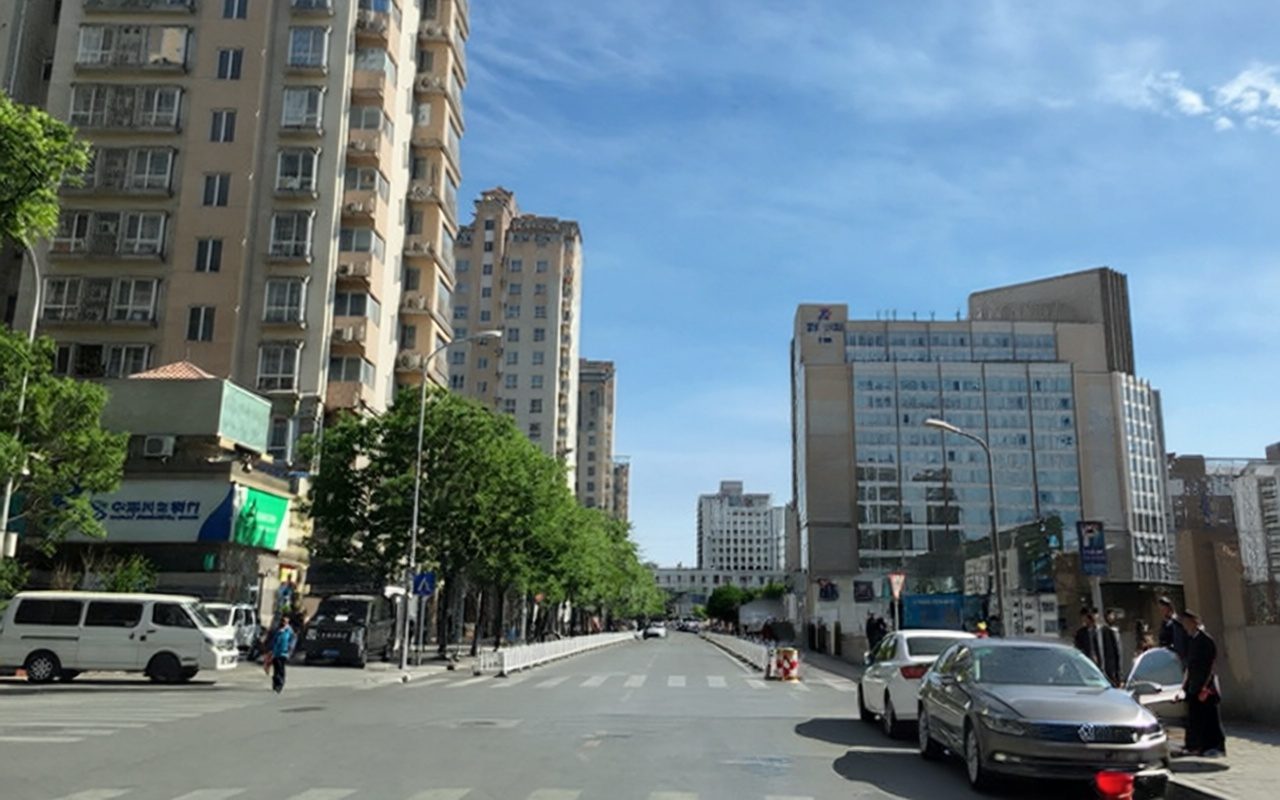}
    \includegraphics[width=0.12\textwidth]{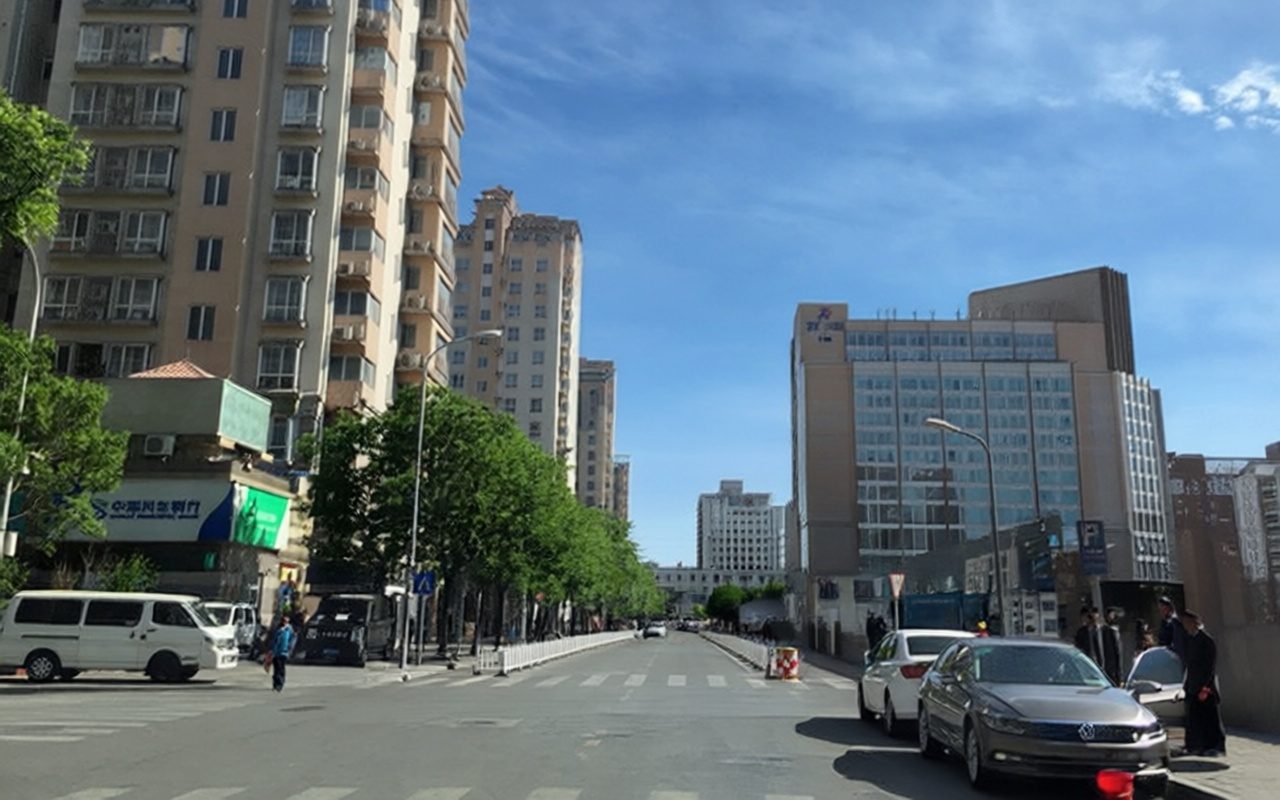}
    \includegraphics[width=0.12\textwidth]{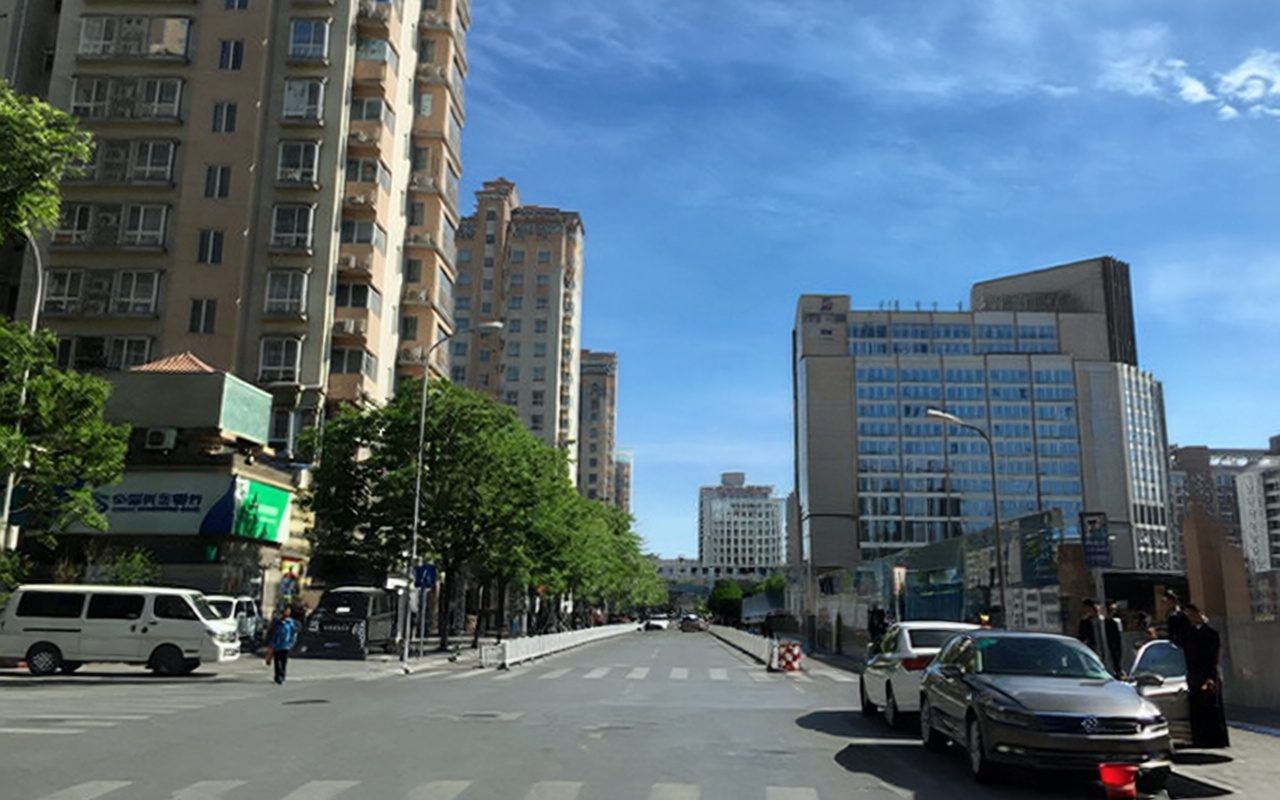}
    \includegraphics[width=0.12\textwidth]{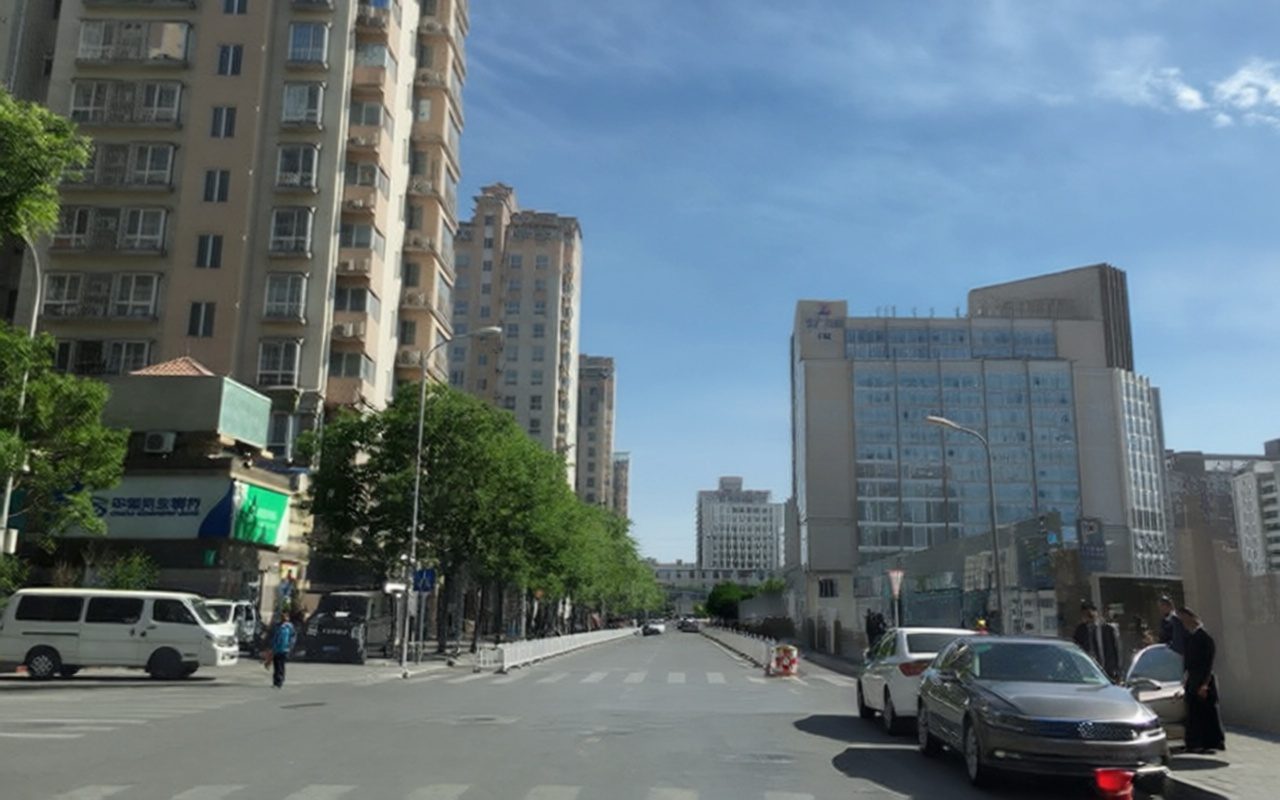}
    \includegraphics[width=0.12\textwidth]{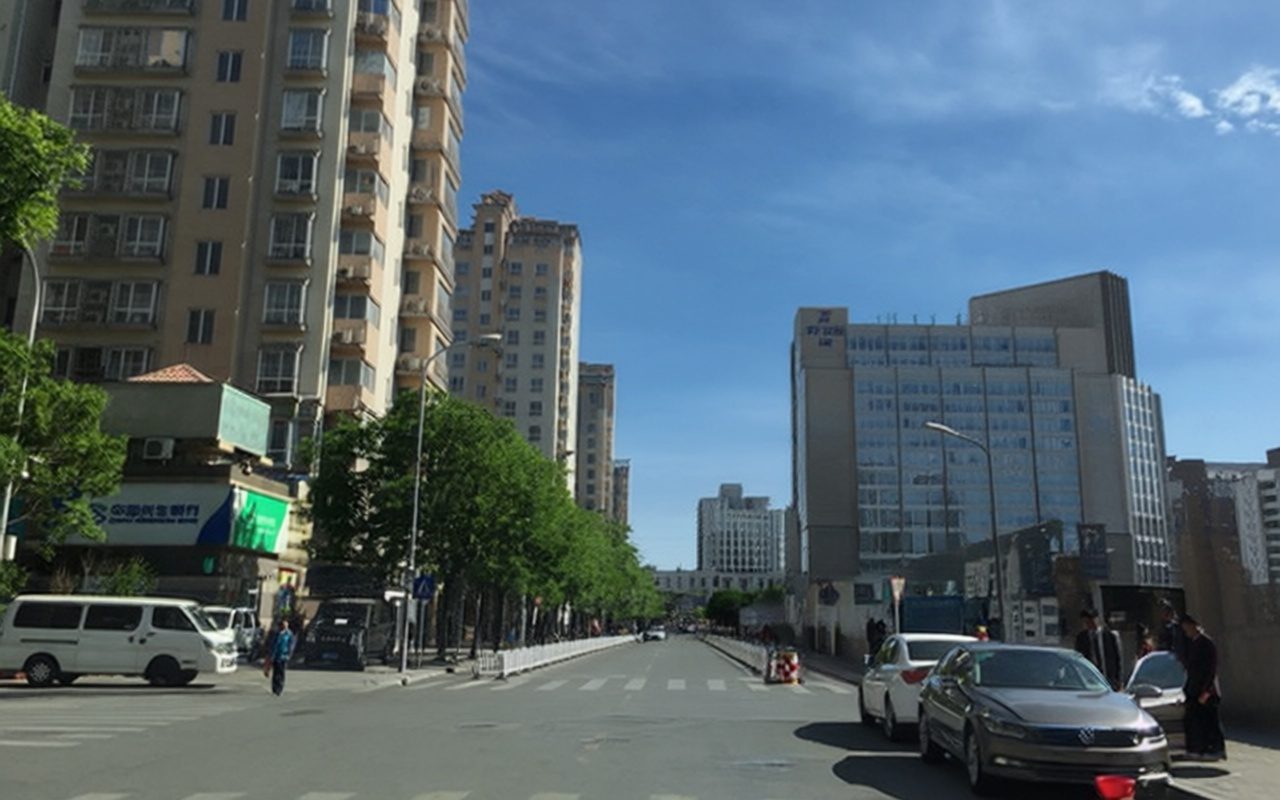}
    \includegraphics[width=0.12\textwidth]{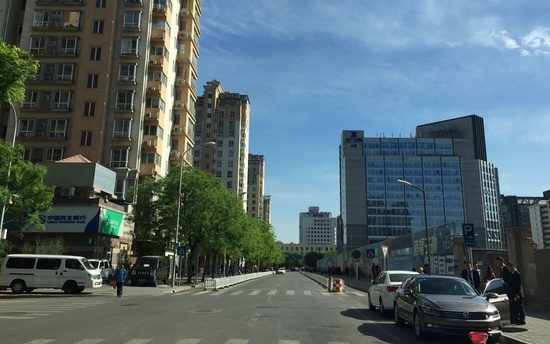}


    \includegraphics[width=0.12\textwidth]{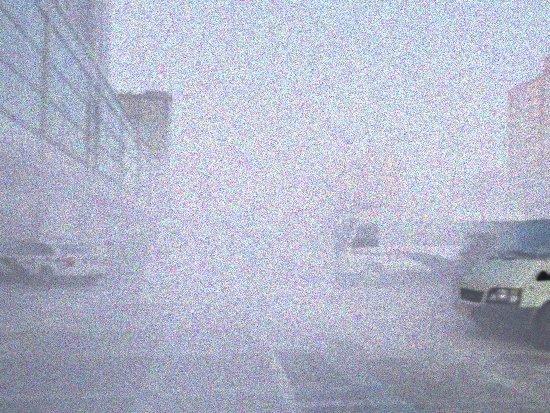}
    \includegraphics[width=0.12\textwidth]{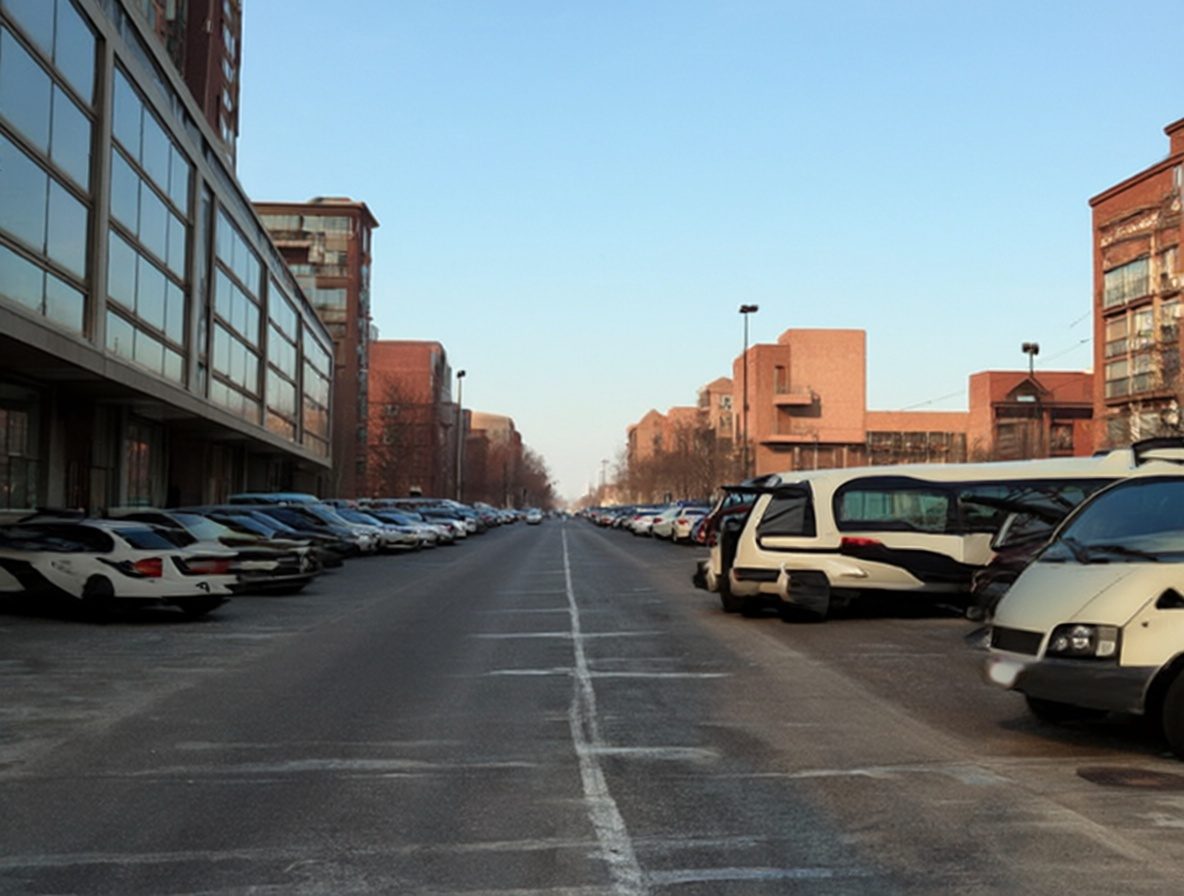}
    \includegraphics[width=0.12\textwidth]{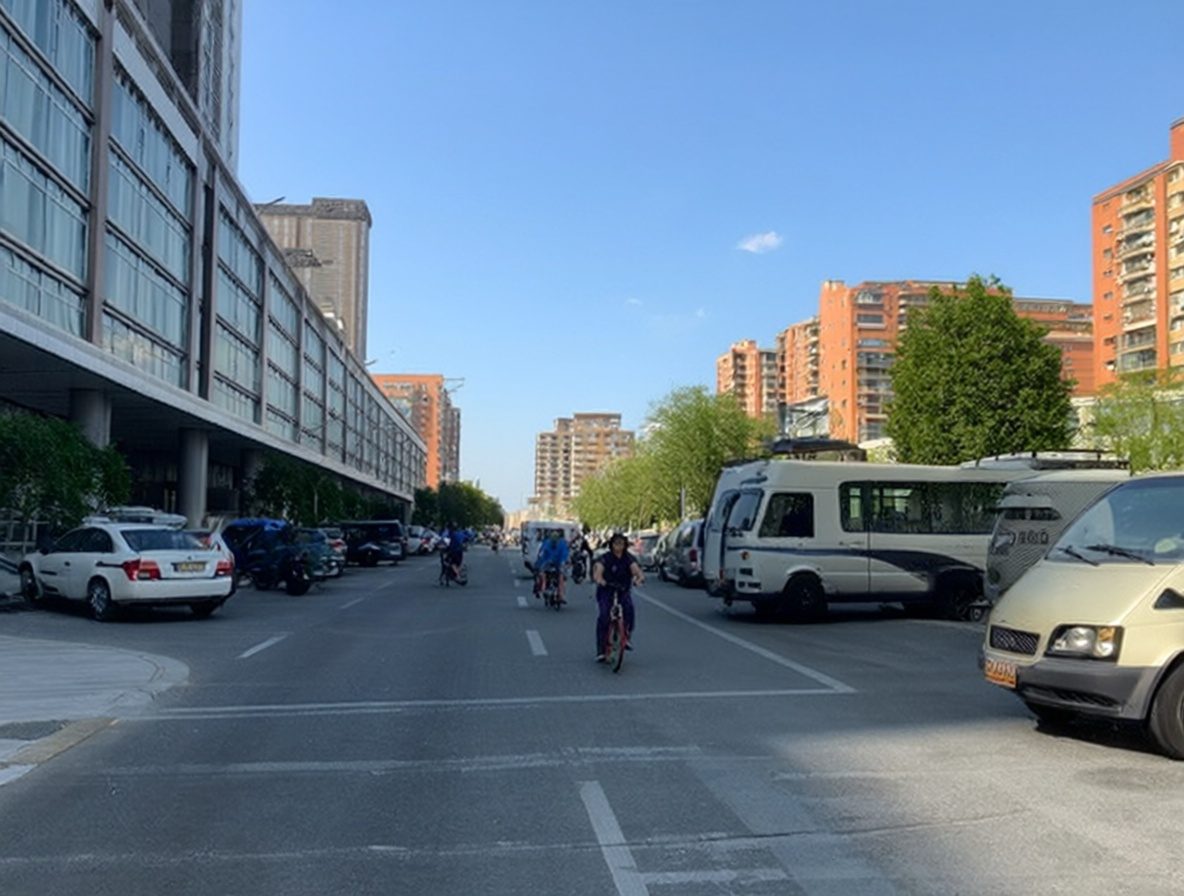}
    \includegraphics[width=0.12\textwidth]{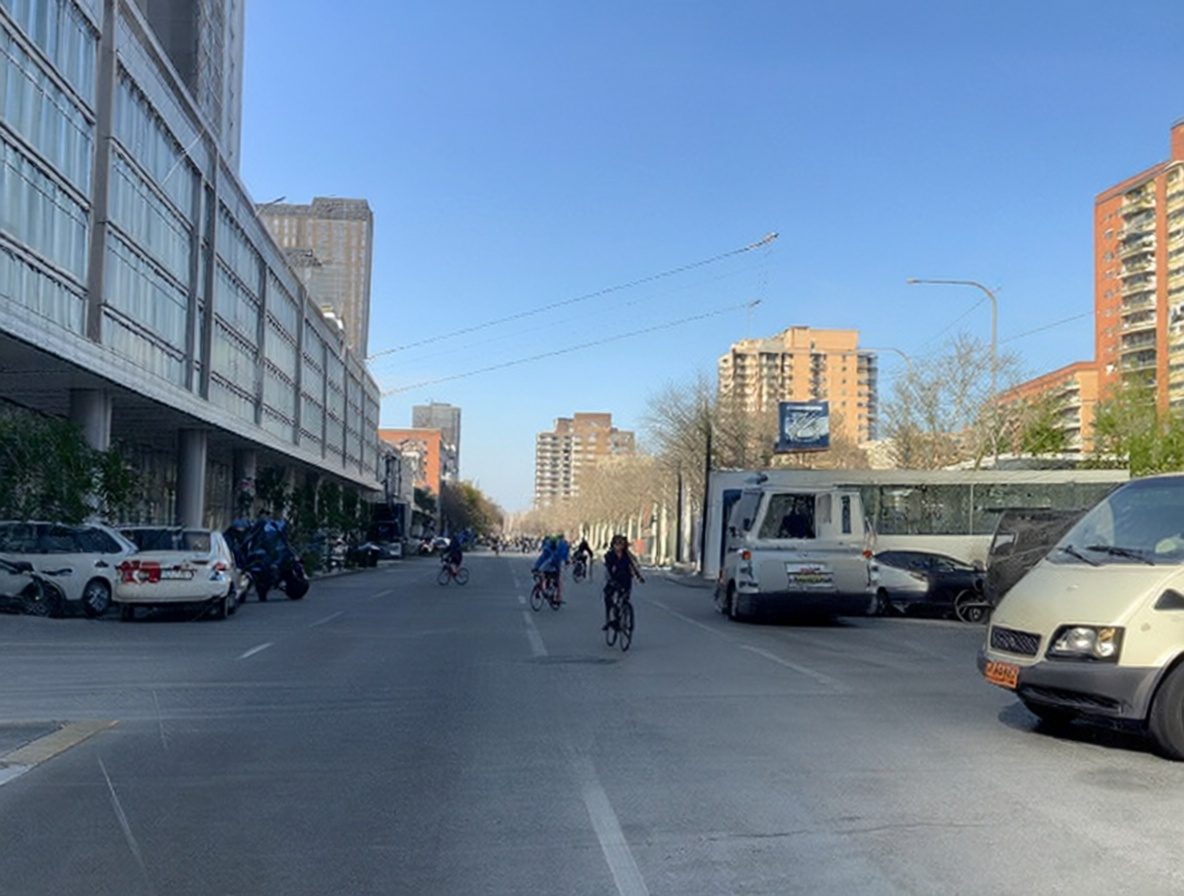}
    \includegraphics[width=0.12\textwidth]{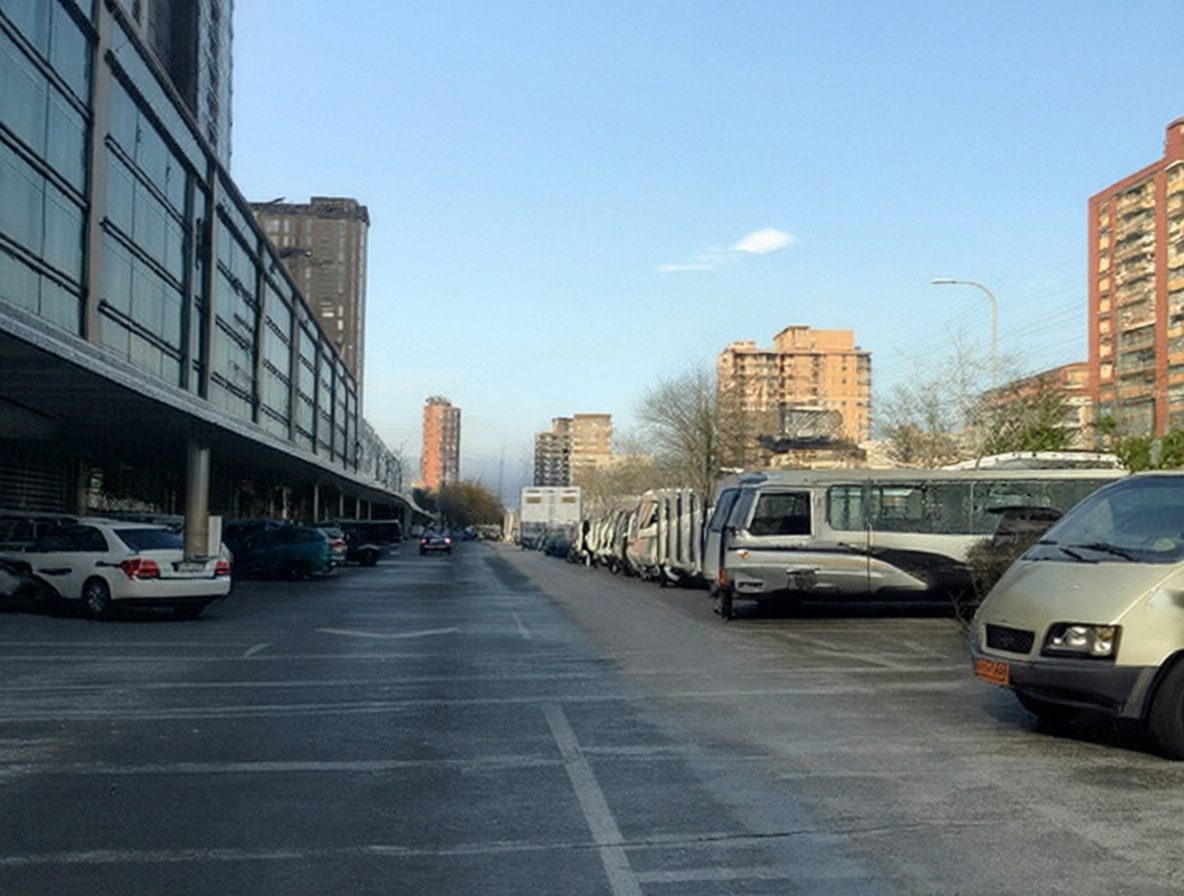}
    \includegraphics[width=0.12\textwidth]{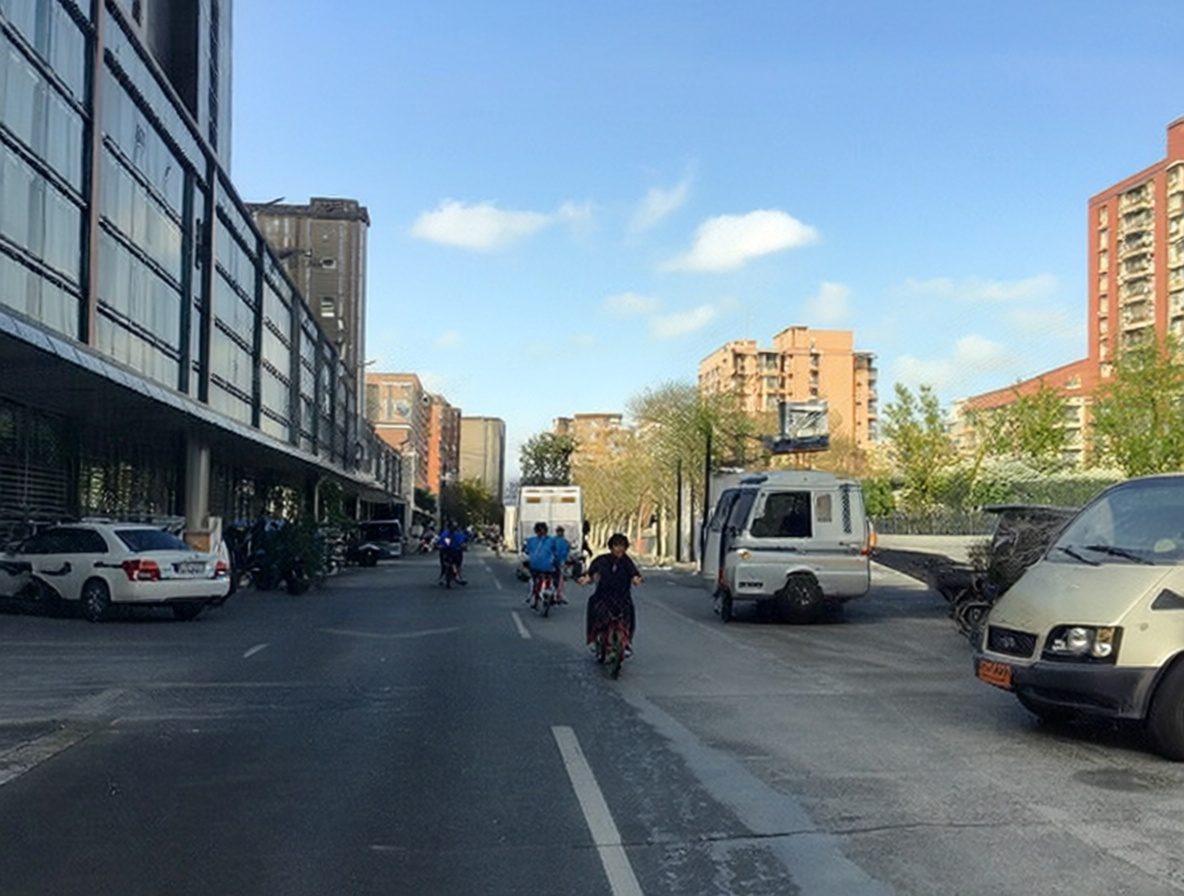}
    \includegraphics[width=0.12\textwidth]{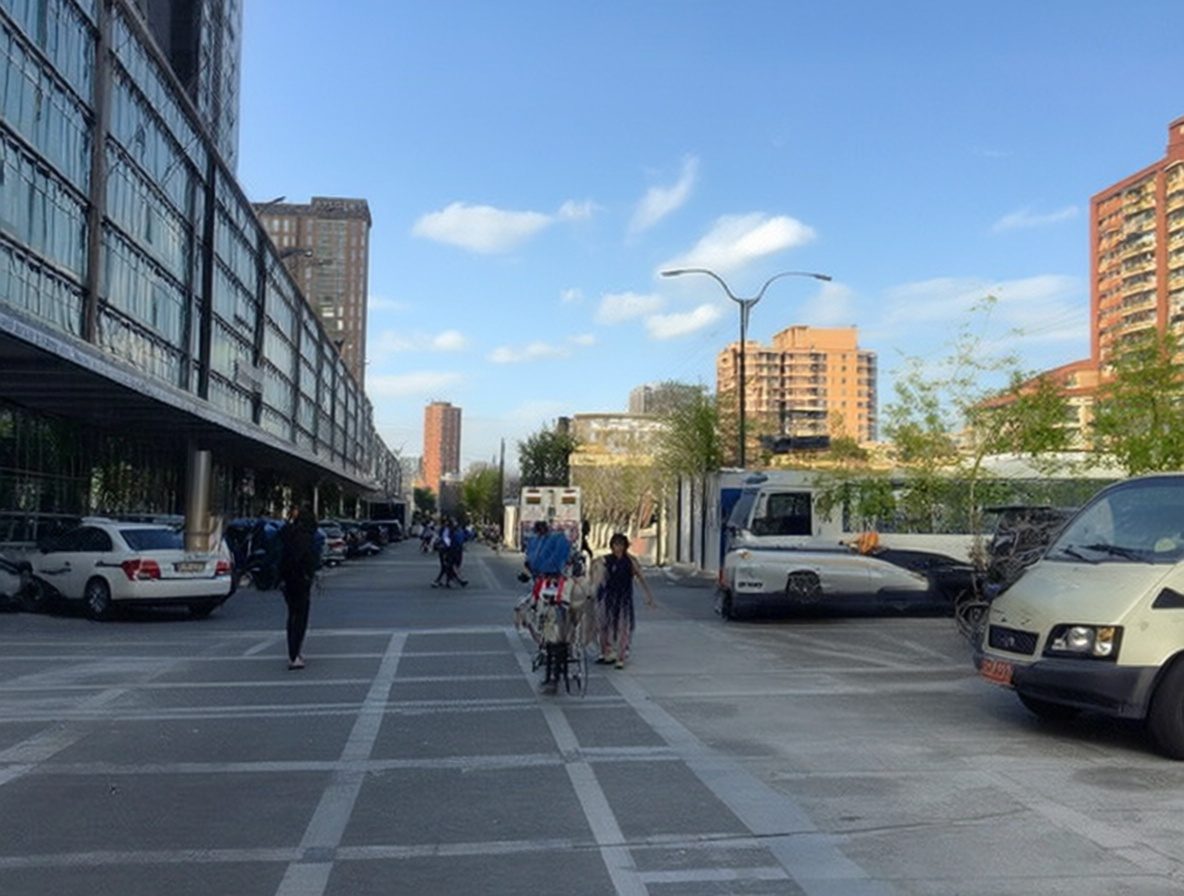}
    \includegraphics[width=0.12\textwidth]{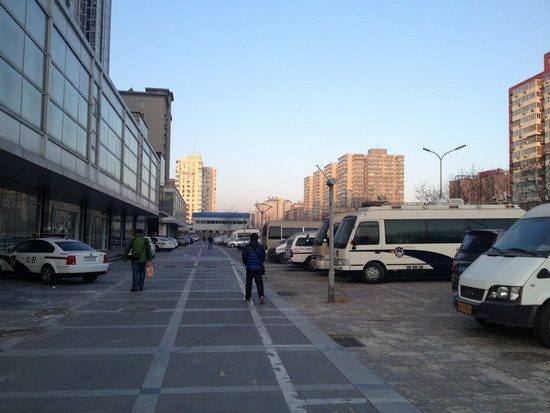}    
    \caption{Ablation Study.}
    \label{fig:ablation}
\end{figure*}
\vspace{-2mm}
\subsection{Quantitative Experiment}
\cref{tab:main_results} shows the quantitative comparison among different methods on the OTS and RESIDE test sets using PSNR and SSIM as evaluation metrics, where higher values indicate better restoration quality. 
The input images are applied with a heavy degradation and results in very low PSNR and SSIM scores. 
The conventional restoration pipeline (3D) provides only marginal improvement, indicating limited robustness under complex degradations. FoundIR~\cite{foundir2025}, Img2Img-Turbo~\cite{parmar2024one}, Stable Diffusion3~\cite{esser2024scaling} and Qwen-Image~\cite{wu2025qwen}~(prompted by a general restoration request), substantially improve both PSNR and SSIM on both datasets. 
Among them, Stable Diffusion 3 performs competitively on OTS in terms of PSNR which means a better color preservation.
Meanwhile, Qwen-Image achieves relatively higher SSIM on RESIDE, which indicates better structural preservation. 
However, these methods still fall short in maintaining consistent fidelity across datasets. 
Our proposed R\&R consistently achieves the best performance on both benchmarks, reaching 19.564 dB / 0.6214 SSIM on OTS and 17.0036 dB / 0.6188 SSIM on RESIDE. These results show the superiority of our proposed R\&R which is generalized to different scenarios and complicated environments.
\subsection{Ablation Study}
\cref{tab:ablation} presents the ablation results in the OTS and RESIDE test sets that evaluate the contribution of each component to our proposed framework. We validate the CoT in the reasoner by ablating a specific chain of reasoning.
Without reasoner refinement (“w/o Reasoner SFT”), the model produces vague semantic reasoning instead of degradation-specific CoT, which provides misleading guidance, resulting in degraded restoration quality.
Removing the scoring module (“w/o Score”) leads to a noticeable drop in both PSNR and SSIM in both datasets, indicating that the scoring mechanism plays an important role in guiding high-quality restoration. 
Excluding parameters (“w/o Parameters”) cases a significant drop on SSIM which indicates the parameters are essential for structural preservation. 
When the semantic guidance is removed (“w/o Semantics”), the model shows reduced structural fidelity, where the SSIM scores are low in both datasets. 
Eliminating reinforcement learning optimization (“w/o RL”) results in a consistent performance decline compared to the full model, demonstrating that RL-based optimization effectively improves both reconstruction accuracy and perceptual quality. 
Overall, the full model achieves the best performance, reaching 19.564 dB / 0.6214 SSIM on OTS and 17.0036 dB / 0.6188 SSIM on RESIDE. This indicates that each component contributes positively and their combination is critical for achieving optimal restoration performance, where visual ablation results in \cref{fig:ablation} show this trend as well.
\begin{table}[ht]
\small
\centering
\caption{Ablation study on OTS and RESIDE test sets.}
\label{tab:ablation}
\begin{tabular}{l|cc|cc}
\hline
\multirow{2}{*}{Setting} 
& \multicolumn{2}{c|}{OTS} 
& \multicolumn{2}{c}{RESIDE} \\
& PSNR$\uparrow$ & SSIM$\uparrow$ & PSNR$\uparrow$ & SSIM$\uparrow$ \\
\hline
Input             & 13.4610 & 0.3258 & 13.0450 & 0.3182 \\
w/o Reasoner SFT & 14.2184 & 0.4605 & 13.1645 & 0.4607 \\
w/o Score         & 18.0576 & 0.6003 & 15.8875 & 0.6000 \\
w/o Parameters    & 18.4592 & 0.6105 & 16.4289 & 0.5825 \\
w/o Semantics     & 18.7185 & 0.5952 & 16.3532 & 0.5900 \\
w/o RL            & 18.9478 & 0.6213 & 16.7890 & 0.6113 \\
\hline
Full Model        & \textbf{19.564} & \textbf{0.6214} & \textbf{17.0036} & \textbf{0.6188} \\
\hline

\end{tabular}
\end{table}
\vspace{-2mm}

\label{sec:exp}

\section{Conclusion}
In this paper, we propose \textbf{Reason and Restore (R\&R)}, a unified framework that incorporates structured Chain-of-Thought reasoning into universal image restoration. By explicitly diagnosing degradation composition, severity, degradation parameters, and scene semantics before pixel-level reconstruction, R\&R enables degradation-aware and scene-adaptive restoration under complex and mixed degradations. The diagnostic outputs serve as structured priors for restoration and are further aligned with reconstruction quality through severity-based reinforcement learning. 
Extensive experiments on synthetic benchmarks and real-world images show that R\&R consistently outperforms strong foundation-model-based baselines, highlighting the promise of tightly integrating explicit reasoning with image restoration.
\label{sec:conclusion}


\bibliography{example_paper}

@article{zhang2017beyond,
  title={Beyond a Gaussian denoiser: Residual learning of deep CNN for image denoising},
  author={Zhang, Kai and Zuo, Wangmeng and Chen, Yunjin and Meng, Deyu and Zhang, Lei},
  journal={IEEE Transactions on Image Processing},
  volume={26},
  number={7},
  pages={3142--3155},
  year={2017}
}

@inproceedings{liang2021swinir,
  title={SwinIR: Image restoration using swin transformer},
  author={Liang, Jingyun and Cao, Jiezhang and Sun, Guolei and Zhang, Kai and Van Gool, Luc and Timofte, Radu},
  booktitle={Proceedings of the IEEE/CVF International Conference on Computer Vision},
  pages={1833--1844},
  year={2021}
}

@inproceedings{zamir2022restormer,
  title={Restormer: Efficient transformer for high-resolution image restoration},
  author={Zamir, Syed Waqas and Arora, Aditya and Khan, Salman and Hayat, Munawar and Khan, Fahad Shahbaz and Yang, Ming-Hsuan},
  booktitle={Proceedings of the IEEE/CVF Conference on Computer Vision and Pattern Recognition},
  pages={5728--5739},
  year={2022}
}

@inproceedings{wang2024exploiting,
 author = {Wang, Jianyi and Yue, Zongsheng and Zhou, Shangchen and Chan, Kelvin C.K. and Loy, Chen Change},
  title = {Exploiting Diffusion Prior for Real-World Image Super-Resolution},
  booktitle = {International Journal of Computer Vision},
  year = {2024}
}

@inproceedings{yue2023resshift,
  title={ResShift: Efficient Diffusion Model for Image Super-resolution by Residual Shifting},
  author={Yue, Zongsheng and Wang, Jianyi and Chen, Chao and Yi, Zewei},
  booktitle={Advances in Neural Information Processing Systems},
  year={2023}
}

@inproceedings{foundir2025,
  title={FoundIR: Unleashing Million-scale Training Data to Advance Foundation Models for Image Restoration},
      author={Li, Hao and Chen, Xiang and Dong, Jiangxin and Tang, Jinhui and Pan, Jinshan},
      booktitle={ICCV},
      year={2025}
}

@article{he2024restoreagent,
  title={Restoreagent: Autonomous image restoration agent via multimodal large language models},
  author={Chen, Haoyu and Li, Wenbo and Gu, Jinjin and Ren, Jingjing and Chen, Sixiang and Ye, Tian and Pei, Renjing and Zhou, Kaiwen and Song, Fenglong and Zhu, Lei},
  journal={Advances in Neural Information Processing Systems},
  volume={37},
  pages={110643--110666},
  year={2024}
}

@article{qagent2025,
  title={Q-Agent: Quality-Driven Chain-of-Thought Image Restoration Agent through Robust Multimodal Large Language Model},
  author={Zhou, Yingjie and Cao, Jiezhang and Zhang, Zicheng and Wen, Farong and Jiang, Yanwei and Jia, Jun and Liu, Xiaohong and Min, Xiongkuo and Zhai, Guangtao},
  journal={arXiv preprint arXiv:2504.07148},
  year={2025}
}

@misc{shao2024deepseekmath,
      title={DeepSeekMath: Pushing the Limits of Mathematical Reasoning in Open Language Models}, 
      author={Zhihong Shao and Peiyi Wang and Qihao Zhu and Runxin Xu and Junxiao Song and Mingchuan Zhang and Y. K. Li and Y. Wu and Daya Guo},
      year={2024},
      eprint={2402.03300},
      archivePrefix={arXiv},
      primaryClass={cs.CL}
}

@inproceedings{black2024training,
  title={Training Diffusion Models with Reinforcement Learning},
  author={Kevin Black and Michael Janner and Yilun Du and Ilya Kostrikov and Sergey Levine},
  booktitle={International Conference on Learning Representations},
  year={2024}
}

@inproceedings{chen2024cot,
  title={Measuring and Improving Chain-of-Thought Reasoning in Vision-Language Models},
  author={Chen, Yangyi and Sikka, Karan and Cogswell, Michael and Ji, Heng and Divakaran, Ajay},
  booktitle={NAACL},
  year={2024}
}

@inproceedings{zhang2025cot,
  title={Improve Vision-Language Model Chain-of-Thought Reasoning},
  author={Zhang, Ruohong and Zhang, Bowen and Zhang, Cheng and Yang, Yiming},
  booktitle={ACL},
  year={2025}
}

@inproceedings{agenticir,
      title={An Intelligent Agentic System for Complex Image Restoration Problems},
      author={Kaiwen Zhu and Jinjin Gu and Zhiyuan You and Yu Qiao and Chao Dong},
      booktitle={The Thirteenth International Conference on Learning Representations},
      year={2025},
}

@article{jiang2025survey,
  title={A survey on all-in-one image restoration: Taxonomy, evaluation and future trends},
  author={Jiang, Junjun and Zuo, Zengyuan and Wu, Gang and Jiang, Kui and Liu, Xianming},
  journal={IEEE Transactions on Pattern Analysis and Machine Intelligence},
  year={2025},
  publisher={IEEE}
}

@article{zhang2023perception,
  title={Perception and sensing for autonomous vehicles under adverse weather conditions: A survey},
  author={Zhang, Yuxiao and Carballo, Alexander and Yang, Hanting and Takeda, Kazuya},
  journal={ISPRS Journal of Photogrammetry and Remote Sensing},
  volume={196},
  pages={146--177},
  year={2023},
  publisher={Elsevier}
}

@article{freeman2021aerial,
  title={Aerial robotic technologies for civil engineering: established and emerging practice},
  author={Freeman, Madelaine R and Kashani, Mehdi M and Vardanega, Paul J},
  journal={Journal of Unmanned Vehicle Systems},
  volume={9},
  number={2},
  pages={75--91},
  year={2021},
  publisher={Canadian Science Publishing 1840 Woodward Drive, Suite 1, Ottawa, ON K2C 0P7}
}

@article{galshetwar2025clear,
  title={Clear Roads, Clear Vision: Advancements in Multi-Weather Restoration for Smart Transportation},
  author={Galshetwar, Vijay M and Hambarde, Praful and Patil, Prashant W and Dudhane, Akshay and Chaudhary, Sachin and Vipparathi, Santosh Kumar and Murala, Subrahmanyam},
  journal={arXiv preprint arXiv:2510.09228},
  year={2025}
}

@article{zhang2021plug,
  title={Plug-and-play image restoration with deep denoiser prior},
  author={Zhang, Kai and Li, Yawei and Zuo, Wangmeng and Zhang, Lei and Van Gool, Luc and Timofte, Radu},
  journal={IEEE Transactions on Pattern Analysis and Machine Intelligence},
  volume={44},
  number={10},
  pages={6360--6376},
  year={2021},
  publisher={IEEE}
}

@inproceedings{schmidt2014shrinkage,
  title={Shrinkage fields for effective image restoration},
  author={Schmidt, Uwe and Roth, Stefan},
  booktitle={Proceedings of the IEEE conference on computer vision and pattern recognition},
  pages={2774--2781},
  year={2014}
}

@inproceedings{yan2020optical,
  title={Optical flow in dense foggy scenes using semi-supervised learning},
  author={Yan, Wending and Sharma, Aashish and Tan, Robby T},
  booktitle={Proceedings of the IEEE/CVF Conference on Computer Vision and Pattern Recognition},
  pages={13259--13268},
  year={2020}
}

@inproceedings{luo2023refusion,
  title={Refusion: Enabling large-size realistic image restoration with latent-space diffusion models},
  author={Luo, Ziwei and Gustafsson, Fredrik K and Zhao, Zheng and Sj{\"o}lund, Jens and Sch{\"o}n, Thomas B},
  booktitle={Proceedings of the IEEE/CVF conference on computer vision and pattern recognition},
  pages={1680--1691},
  year={2023}
}

@article{luo2025taming,
  title={Taming diffusion models for image restoration: a review},
  author={Luo, Ziwei and Gustafsson, Fredrik and Zhao, Zheng and Sj{\"o}lund, Jens and Sch{\"o}n, Thomas},
  journal={Philosophical Transactions A},
  volume={383},
  number={2299},
  pages={20240358},
  year={2025},
  publisher={The Royal Society}
}

@inproceedings{wang2020model,
  title={A model-driven deep neural network for single image rain removal},
  author={Wang, Hong and Xie, Qi and Zhao, Qian and Meng, Deyu},
  booktitle={Proceedings of the IEEE/CVF conference on computer vision and pattern recognition},
  pages={3103--3112},
  year={2020}
}

@inproceedings{yan2020nighttime,
  title={Nighttime defogging using high-low frequency decomposition and grayscale-color networks},
  author={Yan, Wending and Tan, Robby T and Dai, Dengxin},
  booktitle={European Conference on Computer Vision},
  pages={473--488},
  year={2020},
  organization={Springer}
}

@article{gui2023comprehensive,
  title={A comprehensive survey and taxonomy on single image dehazing based on deep learning},
  author={Gui, Jie and Cong, Xiaofeng and Cao, Yuan and Ren, Wenqi and Zhang, Jun and Zhang, Jing and Cao, Jiuxin and Tao, Dacheng},
  journal={ACM Computing Surveys},
  volume={55},
  number={13s},
  pages={1--37},
  year={2023},
  publisher={ACM New York, NY}
}

@inproceedings{vaishnav2023promptir,
  title={PromptIR: Prompting for All-in-One Image Restoration},
  author={Potlapalli, Vaishnav and Zamir, Syed Waqas and Khan, Salman and Khan, Fahad},
  booktitle={Thirty-seventh Conference on Neural Information Processing Systems},
  year={2023}
}

@inproceedings{conde2024instructir,
  title={InstructIR: High-quality image restoration following human instructions},
  author={Conde, Marcos V and Geigle, Gregor and Timofte, Radu},
  booktitle={European Conference on Computer Vision},
  pages={1--21},
  year={2024},
  organization={Springer}
}

@article{floridi2020gpt,
  title={GPT-3: Its nature, scope, limits, and consequences},
  author={Floridi, Luciano and Chiriatti, Massimo},
  journal={Minds and machines},
  volume={30},
  number={4},
  pages={681--694},
  year={2020},
  publisher={Springer}
}

@article{wei2022chain,
  title={Chain-of-thought prompting elicits reasoning in large language models},
  author={Wei, Jason and Wang, Xuezhi and Schuurmans, Dale and Bosma, Maarten and Xia, Fei and Chi, Ed and Le, Quoc V and Zhou, Denny and others},
  journal={Advances in neural information processing systems},
  volume={35},
  pages={24824--24837},
  year={2022}
}

@article{parmar2024one,
  title={One-step image translation with text-to-image models},
  author={Parmar, Gaurav and Park, Taesung and Narasimhan, Srinivasa and Zhu, Jun-Yan},
  journal={arXiv preprint arXiv:2403.12036},
  year={2024}
}

@inproceedings{yan2021self,
  title={Self-aligned video deraining with transmission-depth consistency},
  author={Yan, Wending and Tan, Robby T and Yang, Wenhan and Dai, Dengxin},
  booktitle={Proceedings of the IEEE/CVF Conference on Computer Vision and Pattern Recognition},
  pages={11966--11976},
  year={2021}
}

@article{wu2025qwen,
  title={Qwen-image technical report},
  author={Wu, Chenfei and Li, Jiahao and Zhou, Jingren and Lin, Junyang and Gao, Kaiyuan and Yan, Kun and Yin, Sheng-ming and Bai, Shuai and Xu, Xiao and Chen, Yilei and others},
  journal={arXiv preprint arXiv:2508.02324},
  year={2025}
}

@article{li2018benchmarking,
  title={Benchmarking single-image dehazing and beyond},
  author={Li, Boyi and Ren, Wenqi and Fu, Dengpan and Tao, Dacheng and Feng, Dan and Zeng, Wenjun and Wang, Zhangyang},
  journal={IEEE transactions on image processing},
  volume={28},
  number={1},
  pages={492--505},
  year={2018},
  publisher={IEEE}
}

@article{zhao2020dehazing,
  title={Dehazing evaluation: Real-world benchmark datasets, criteria, and baselines},
  author={Zhao, Shiyu and Zhang, Lin and Huang, Shuaiyi and Shen, Ying and Zhao, Shengjie},
  journal={IEEE Transactions on Image Processing},
  volume={29},
  pages={6947--6962},
  year={2020},
  publisher={IEEE}
}

@article{yan2022feature,
  title={Feature-aligned video raindrop removal with temporal constraints},
  author={Yan, Wending and Xu, Lu and Yang, Wenhan and Tan, Robby T},
  journal={IEEE Transactions on Image Processing},
  volume={31},
  pages={3440--3448},
  year={2022},
  publisher={IEEE}
}

@inproceedings{esser2024scaling,
  title={Scaling rectified flow transformers for high-resolution image synthesis},
  author={Esser, Patrick and Kulal, Sumith and Blattmann, Andreas and Entezari, Rahim and M{\"u}ller, Jonas and Saini, Harry and Levi, Yam and Lorenz, Dominik and Sauer, Axel and Boesel, Frederic and others},
  booktitle={Forty-first international conference on machine learning},
  year={2024}
}

@InProceedings{chu2022nafssr,
    author    = {Chu, Xiaojie and Chen, Liangyu and Yu, Wenqing},
    title     = {NAFSSR: Stereo Image Super-Resolution Using NAFNet},
    booktitle = {Proceedings of the IEEE/CVF Conference on Computer Vision and Pattern Recognition (CVPR) Workshops},
    month     = {June},
    year      = {2022},
    pages     = {1239-1248}
}

@article{Qwen3-VL,
      title={Qwen3-VL Technical Report}, 
      author={Shuai Bai and Yuxuan Cai and Ruizhe Chen and Keqin Chen and Xionghui Chen and Zesen Cheng and Lianghao Deng and Wei Ding and Chang Gao and Chunjiang Ge and Wenbin Ge and Zhifang Guo and Qidong Huang and Jie Huang and Fei Huang and Binyuan Hui and Shutong Jiang and Zhaohai Li and Mingsheng Li and Mei Li and Kaixin Li and Zicheng Lin and Junyang Lin and Xuejing Liu and Jiawei Liu and Chenglong Liu and Yang Liu and Dayiheng Liu and Shixuan Liu and Dunjie Lu and Ruilin Luo and Chenxu Lv and Rui Men and Lingchen Meng and Xuancheng Ren and Xingzhang Ren and Sibo Song and Yuchong Sun and Jun Tang and Jianhong Tu and Jianqiang Wan and Peng Wang and Pengfei Wang and Qiuyue Wang and Yuxuan Wang and Tianbao Xie and Yiheng Xu and Haiyang Xu and Jin Xu and Zhibo Yang and Mingkun Yang and Jianxin Yang and An Yang and Bowen Yu and Fei Zhang and Hang Zhang and Xi Zhang and Bo Zheng and Humen Zhong and Jingren Zhou and Fan Zhou and Jing Zhou and Yuanzhi Zhu and Ke Zhu},
	  journal={arXiv preprint arXiv:2511.21631},
      year={2025},
}
\bibliographystyle{icml2026}

\newpage
\appendix
\onecolumn
\section{Appendix}
\label{sec:appendix}

This appendix provides additional details that complement the main paper, including
(i) the prompting templates used in the \emph{Reason} phase,
(ii) additional qualitative visualizations,
and (iii) implementation and experimental details that are omitted from the main body
due to space constraints, such as the semi-realworld data generation process.

\subsection{Detailed Reasoning Prompts for the Reason Phase}
\label{sec:appendix_prompt}

The \emph{Reason} phase relies on carefully designed prompting templates to elicit
structured, restoration-oriented diagnostic reasoning from the vision-language model.
Rather than generic image captioning, the prompt explicitly separates
\emph{degradation diagnosis} from \emph{clean-scene reconstruction in words},
ensuring that semantic understanding is disentangled from degradation artifacts.

Given a degraded input image, the model is instructed to perform the following tasks:
(i) identify all present degradation types,
(ii) quantify their severity,
(iii) estimate degradation-specific physical or statistical parameters,
and (iv) reconstruct the underlying clean scene purely at the semantic and geometric level,
without referencing any degradation cues.

A representative prompt used in our experiments is shown below:

\begin{quote}
\small
\textbf{User Prompt:}

\texttt{Please analyze this image and do the following steps, \textbf{step-by-step}: <image>}\\ 
\texttt{\textbf{First}, analyze the degradation effects applied to this image}\\
\texttt{\textbf{Second, based on previous understanding}, give each degradation a severity score, 0 means no degradation and 100 means the worst degradation.}\\
\texttt{\textbf{Third, based on previous understanding}, give degradation-specific parameters accordingly.}\\
\texttt{\textbf{Fourth, based on previous understanding}, describe the underlying clean scene (max 30 words) following the instruction:}\\
\texttt{Provide a detailed semantic and geometric description of this image: <image> (max 30 words):}\\
\texttt{- Main subjects and their shape}\\
\texttt{- Geometry and edges that should be sharp}\\
\texttt{- Texture that should be restored}\\
\texttt{- Regions that should be smooth}\\
\texttt{- Probable true colors of objects}\\
\texttt{- Relative depth ordering}\\
\texttt{- Occlusion relationships}\\
\texttt{- Do NOT describe fog, blur, rain, noise, low contrast, artifacts, or anything related to degradation.}\\
\texttt{-Only reconstruct the scene content in words.}\\
\texttt{\textbf{Finally}, check all your outputs and ensure they are correct and self-consistent.}
\end{quote}

The corresponding model output is constrained to follow a predefined diagnostic format,
including explicit binary indicators for each degradation type, continuous severity
scores, degradation-specific parameter estimates, and a concise clean-scene description.
An example output includes:

\begin{itemize}
    \item Fog degradation: Yes (intensity score)
    \item Motion blur: Yes (intensity score)
    \item Rain streaks: Yes (intensity score)
    \item Gaussian noise: Yes (intensity score)
    \item Detailed degradation parameters (e.g., atmospheric light, transmission, blur kernel statistics)
    \item Clean scene description (no more than 30 words)
\end{itemize}

Importantly, the prompt explicitly prohibits mentioning any degradation effects when
describing the clean scene. This constraint forces the model to reason about the
\emph{latent clean content}—including object geometry, textures, materials, color
priors, and depth ordering—rather than merely paraphrasing visible artifacts.
Such degradation-agnostic scene descriptions provide high-level semantic and geometric
priors that are directly useful for guiding the subsequent restoration process.

This prompt design encourages the model to first \emph{understand} how the image is
degraded and what the scene should look like, before any pixel-level reconstruction
is performed, thereby operationalizing the \textbf{Reason-before-Restore} principle.
\subsection{Semi-realworld Data Generation Details}
\label{sec:appendix_data}

To train both the Reason and Restore components, we construct a semi-realworld
degradation dataset using a procedural degradation pipeline that explicitly
models fog, motion blur, rain streaks, and sensor noise.

For each clean image, the presence of each degradation type is random.
This design produces diverse mixed-degradation scenarios while maintaining
precise access to ground-truth clean images and degradation parameters.

\paragraph{Fog Degradation.}
Fog is simulated using a physically motivated atmospheric scattering model.
An atmospheric light vector $A \in \mathbb{R}^3$ is sampled uniformly from
$[0.7, 1.0]^3$, representing global illumination and color bias.
A depth-dependent transmission map $t(x)$ is generated by exponentiating a
precomputed relative depth prior $t_0(x)$:
\[
t(x) = t_0(x)^{\beta}, \quad \beta \sim \mathcal{U}(0.8, 3.0),
\]
followed by clamping to $[0.05, 1.0]$ for numerical stability.
The foggy image is then synthesized as
$I = J \cdot t + A \cdot (1 - t)$.

To facilitate structured reasoning supervision, we record the mean
transmission $t_{\text{mean}}$ 
as one of fog parameters.
The fog severity score is computed from the mean transmission as
\[
s_{\text{fog}} = \frac{1 - t_{\text{mean}}}{1 - 0.05} \times 99 + 1,
\]
yielding a continuous score in the range $[1,100]$ that monotonically reflects
visibility degradation.

\paragraph{Motion Blur (Shake).}
Motion blur is generated using a Gaussian-weighted random shake kernel that
simulates hand-held camera jitter.
A random walk trajectory with a randomly sampled number of steps is constructed,
and a 1D Gaussian temporal weighting is applied along the trajectory to emphasize
the middle portion of the motion.
The resulting 2D kernel is normalized and applied to the image via convolution.

From the generated kernel, three interpretable parameters are extracted:
(i) the dominant blur direction (via eigen-decomposition of the weighted
second-moment matrix),
(ii) the effective blur length (square root of the largest eigenvalue),
and (iii) the kernel energy (L2 norm).
The shake severity score is computed from the root-mean-square radius of the kernel:
\[
s_{\text{shake}} = \frac{r_{\text{rms}}}{r_{\text{max}}} \times 99 + 1,
\]
where $r_{\text{max}}$ denotes the theoretical maximum kernel radius.

\paragraph{Rain Streaks.}
Rain streaks are synthesized by first sampling sparse random seed points over the
image according to a density parameter, followed by directional motion blur to
elongate seeds into streaks.
The blur direction is determined by a slant angle, and the streak appearance is
controlled by blur length, width, opacity, and color.
Rain color is tied to the atmospheric light to ensure physical consistency.

Rain severity is quantified by computing the spatial coverage ratio of visible
rain streaks (pixels exceeding a fixed intensity threshold).
The rain score is defined as
\[
s_{\text{rain}} = \text{coverage} \times 99 + 1,
\]
which directly reflects the proportion of the image affected by rain artifacts.

\paragraph{Gaussian Readout Noise.}
Sensor noise is modeled as signal-dependent Gaussian readout noise.
For each pixel with intensity $x \in [0,255]$, the noise standard deviation is
computed as
\[
\sigma(x) = k \cdot x + b,
\]
where $k \sim \mathcal{U}(0, 0.3)$ and $b \sim \mathcal{U}(-20, 20)$.
The resulting noise is added in normalized intensity space and clipped to valid
ranges.

The noise severity score is derived from the mean normalized noise variance:
\[
s_{\text{noise}} = \frac{\bar{\sigma}}{0.05} \times 99 + 1,
\]
ensuring a consistent $[1,100]$ scale across samples.

\paragraph{Dataset Outputs.}
For each image, we store the final degraded image along with intermediate results
after each degradation stage.
All sampled parameters, degradation presence flags, and severity scores are logged
and used to generate structured diagnostic annotations.
These annotations serve as supervision targets for the Reason phase and as
conditioning signals and reinforcement rewards for the Restore phase, enabling
tight alignment between degradation diagnosis and restoration behavior.

\subsection{Theoretical Foundation of Fidelity-based Policy Distribution Proxy}
To provide a rigorous foundation for the proposed Reinforcement Learning (RL) framework, we interpret the restoration model through the lens of Energy-Based Models (EBMs). This section demonstrates that using a fidelity-based surrogate for the policy distribution is mathematically consistent with maximum likelihood estimation.

\paragraph{Energy-Based Formulation}
Formally, we define the policy distribution $\pi_\theta(\hat{x} | y, c)$ over the restored image space as a Gibbs distribution:
\begin{equation}
    \pi_\theta(\hat{x} | y, c) = \frac{1}{Z(\theta)} \exp\left( -E(\hat{x}, x_{gt}) \right)
\end{equation}
where $E(\hat{x}, x_{gt})$ represents the energy function measuring the discrepancy between the restored image $\hat{x}$ and the ground truth $x_{gt}$, and $Z(\theta) = \int \exp(-E(\hat{x}, x_{gt})) d\hat{x}$ is the partition function (normalization constant).

\paragraph{Equivalence to Gaussian MLE}
By defining the energy function as the scaled $\ell_2$ loss (Mean Squared Error):
\begin{equation}
    E(\hat{x}, x_{gt}) = \frac{1}{2\sigma^2} \|\hat{x} - x_{gt}\|_2^2
\end{equation}
the policy distribution $\pi_\theta$ simplifies to an isotropic Gaussian distribution:
\begin{equation}
    \pi_\theta(\hat{x} | y, c) \propto \exp\left( -\frac{\|\hat{x} - x_{gt}\|_2^2}{2\sigma^2} \right) \sim \mathcal{N}(x_{gt}, \sigma^2 \mathbf{I})
\end{equation}

Under this formulation, minimizing the MSE during the restoration process is mathematically equivalent to Maximum Likelihood Estimation (MLE) under a Gaussian prior:
\begin{equation}
    \arg\max_\theta \log \pi_\theta(\hat{x} | y, c) \equiv \arg\min_\theta \|\hat{x} - x_{gt}\|_2^2
\end{equation}

\paragraph{Application to GRPO}
In the context of Group Relative Policy Optimization (GRPO), computing the exact log-probability $\log \pi_\theta$ for high-dimensional image tensors is computationally expensive. However, the above equivalence allows us to utilize the fidelity-based softmax distribution as a differentiable surrogate:
\begin{equation}
    \log P_\theta(\hat{x}_i) = \frac{\exp(-\|\hat{x}_i - x_{gt}\|_2^2 / \tau)}{\sum_{j=1}^G \exp(-\|\hat{x}_j - x_{gt}\|_2^2 / \tau)}
\end{equation}
where $G$ is the group size and $\tau$ is a temperature hyperparameter. This ensures that the RL optimization objective aligns the pixel-level fidelity (represented by the energy surface) with the high-level diagnostic rewards provided by the Reasoner, grounding the model's policy refinement in both low-level reconstruction accuracy and high-level semantic consistency.
\subsection{Additional Qualitative Results}
\label{sec:appendix_visual}
\begin{figure*}[htbp]
    \centering
    \begin{subfigure}{0.11\textwidth}
        \centering
        \textbf{Input}
    \end{subfigure}
    \begin{subfigure}{0.11\textwidth}
        \centering
        \textbf{3D}
    \end{subfigure}
    \begin{subfigure}{0.11\textwidth}
        \centering
        \textbf{FoundIR}
    \end{subfigure}
    \begin{subfigure}{0.11\textwidth}
        \centering
        \textbf{Img2Img-Turbo}
    \end{subfigure}
    \begin{subfigure}{0.11\textwidth}
        \centering
        \textbf{Stable Diffusion3}
    \end{subfigure}
    \begin{subfigure}{0.11\textwidth}
        \centering
        \textbf{Qwen3-Image}
    \end{subfigure}
    \begin{subfigure}{0.11\textwidth}
        \centering
        \textbf{(Ours) R\&R}
    \end{subfigure}
    \begin{subfigure}{0.11\textwidth}
        \centering
        \textbf{GT}
    \end{subfigure}
    \vspace{2mm}

    \includegraphics[width=0.11\textwidth]{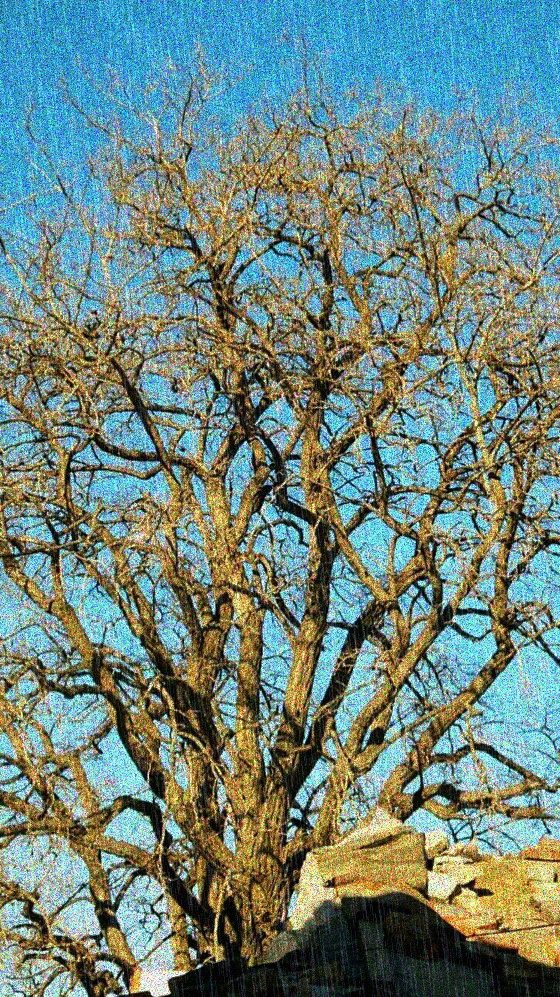}
    \includegraphics[width=0.11\textwidth]{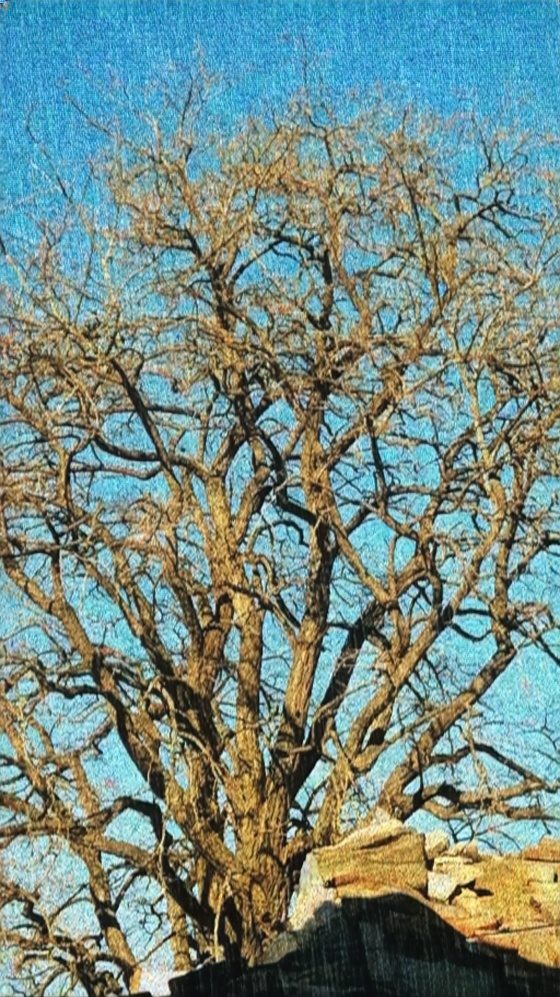}
     \includegraphics[width=0.11\textwidth]{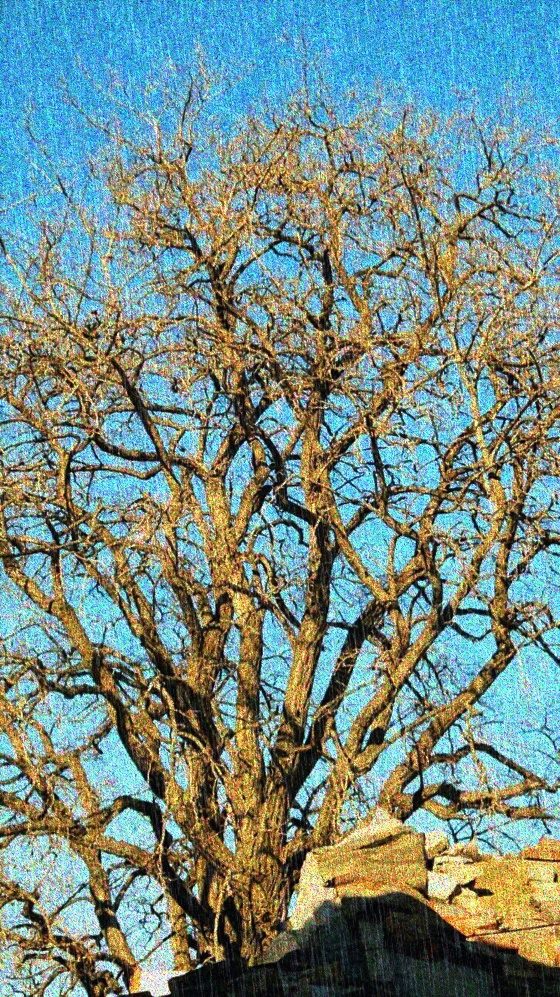}
    \includegraphics[width=0.11\textwidth]{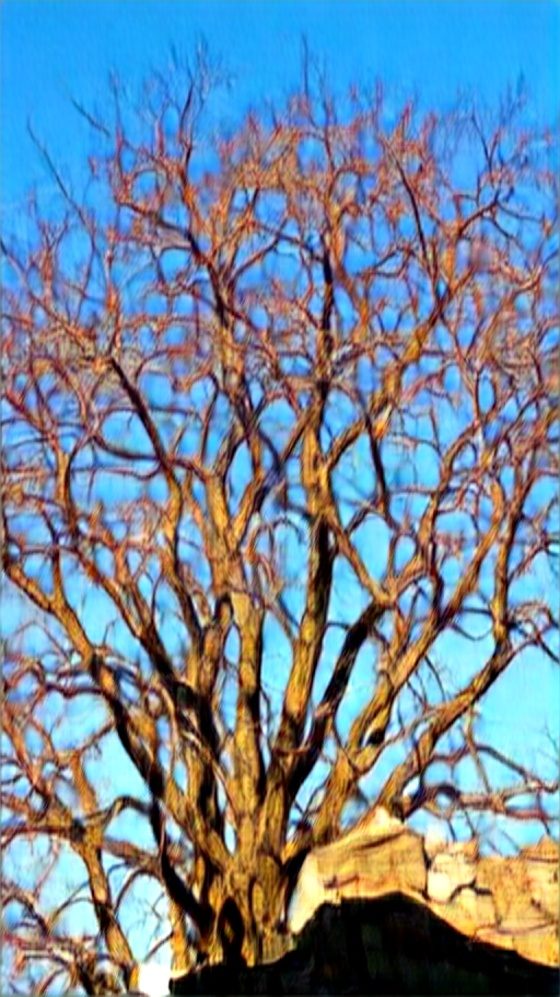}
    \includegraphics[width=0.11\textwidth]{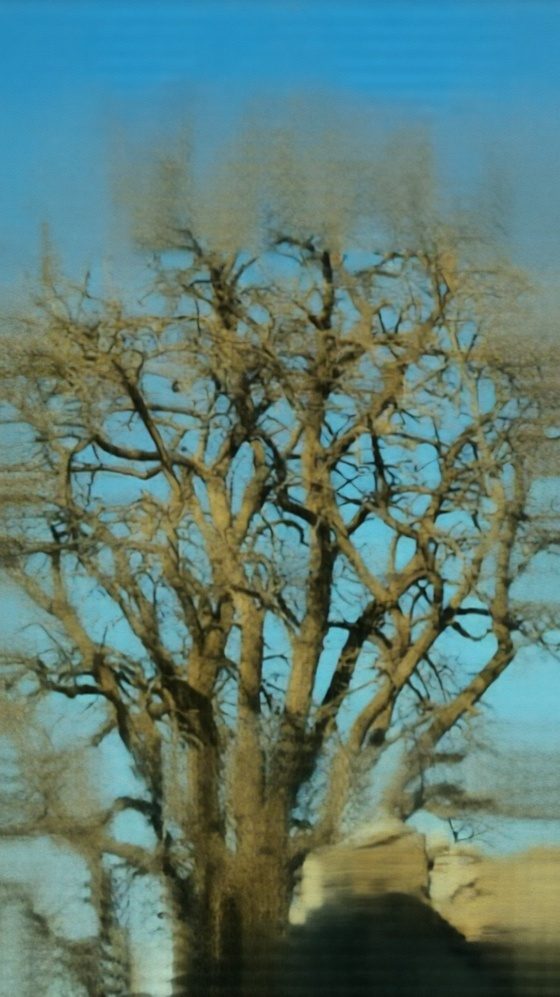}
    \includegraphics[width=0.11\textwidth]{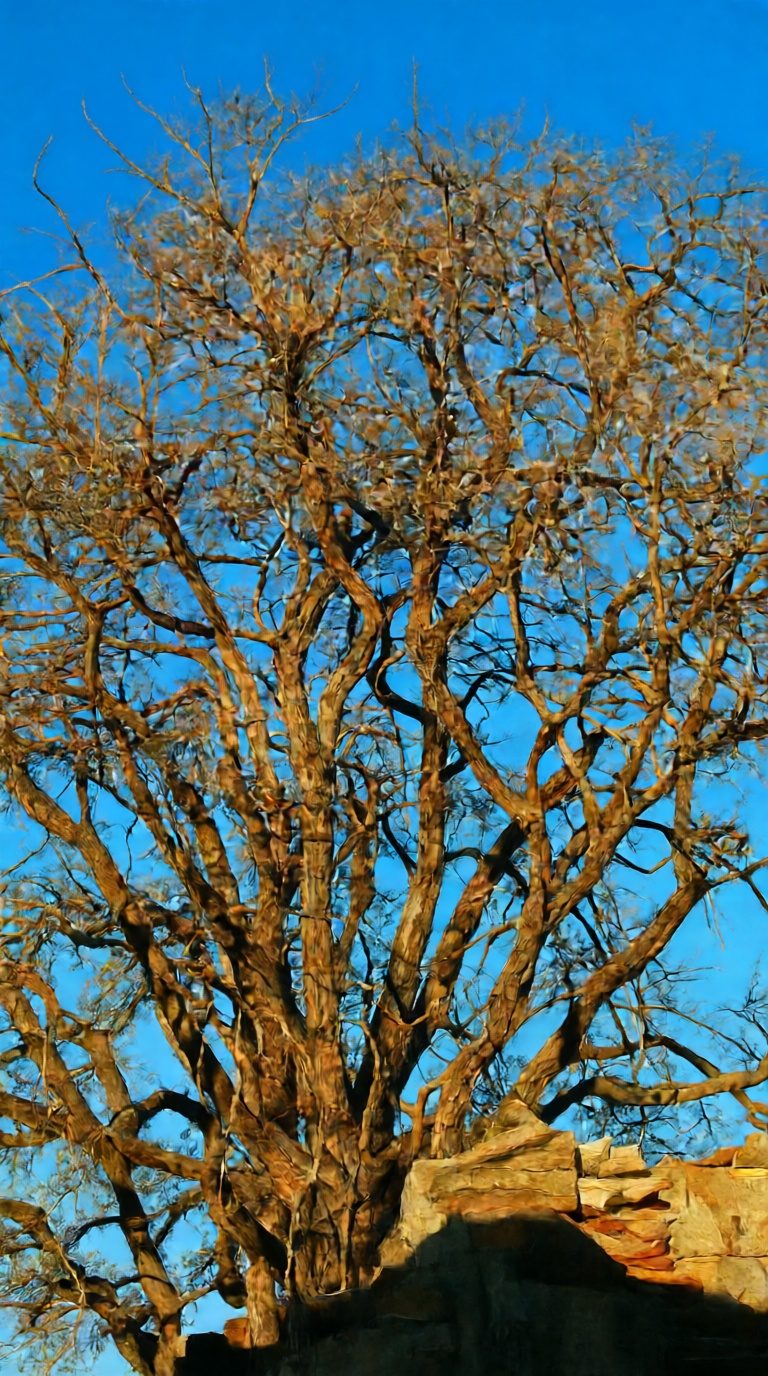}
    \includegraphics[width=0.11\textwidth]{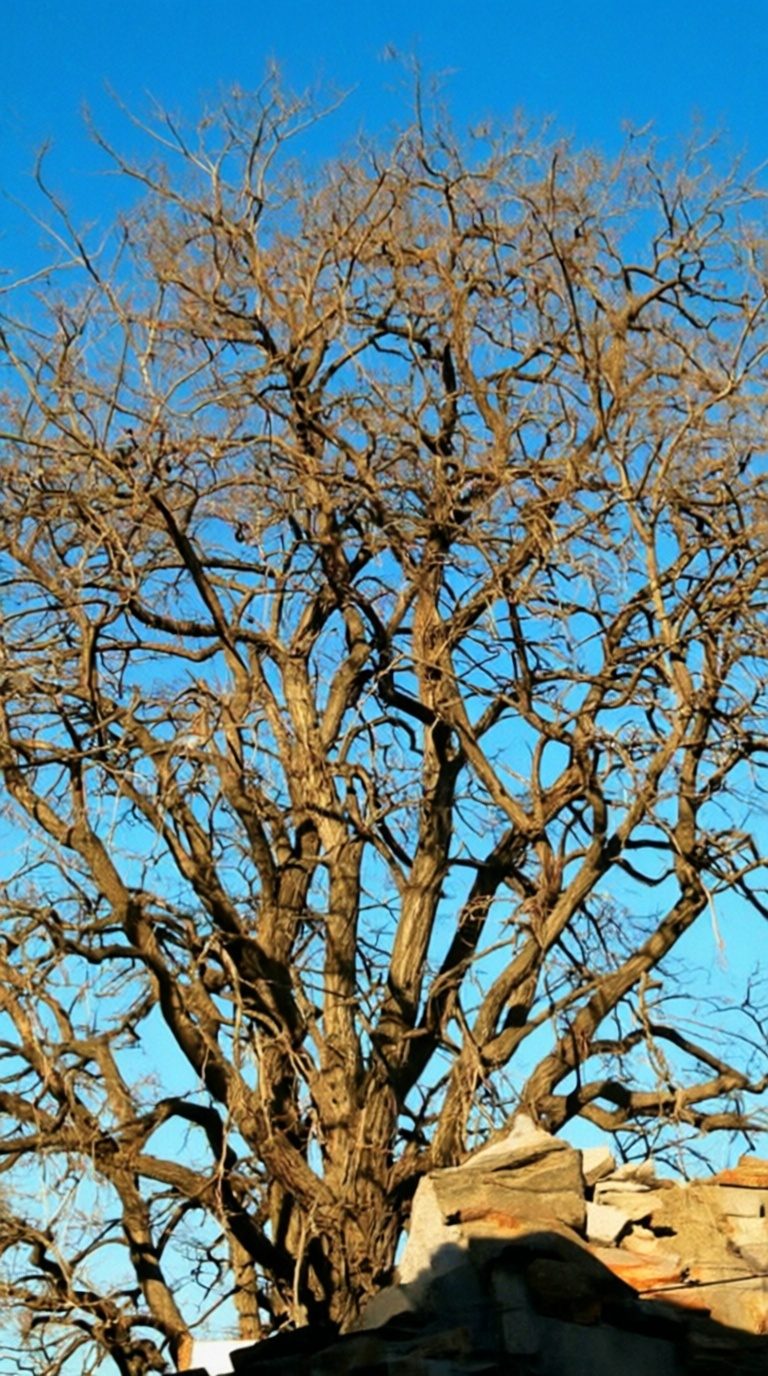}
    \includegraphics[width=0.11\textwidth]{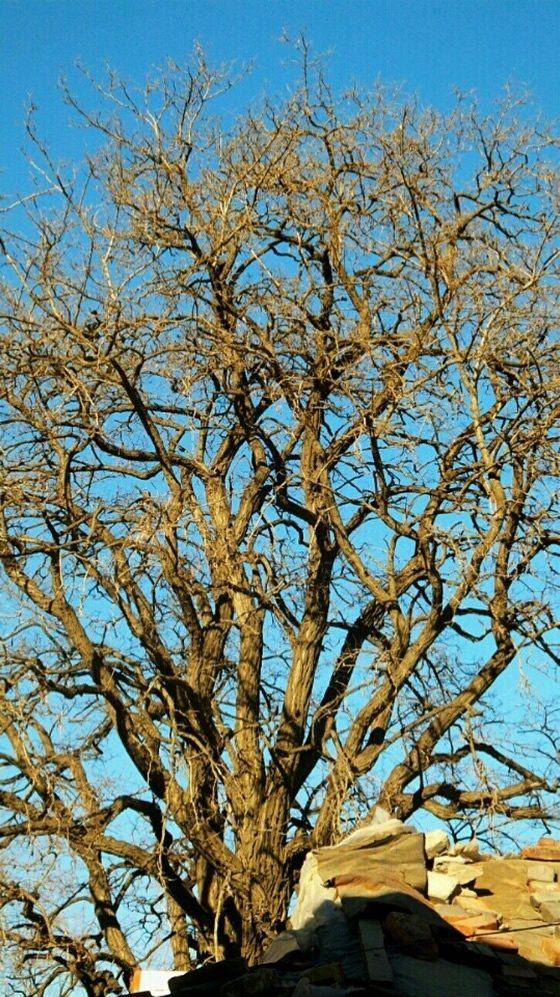}
    
    \includegraphics[width=0.11\textwidth]{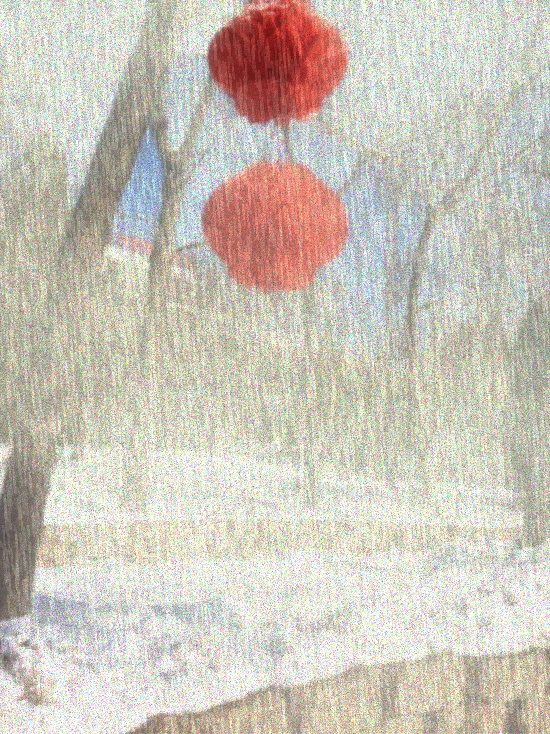}
    \includegraphics[width=0.11\textwidth]{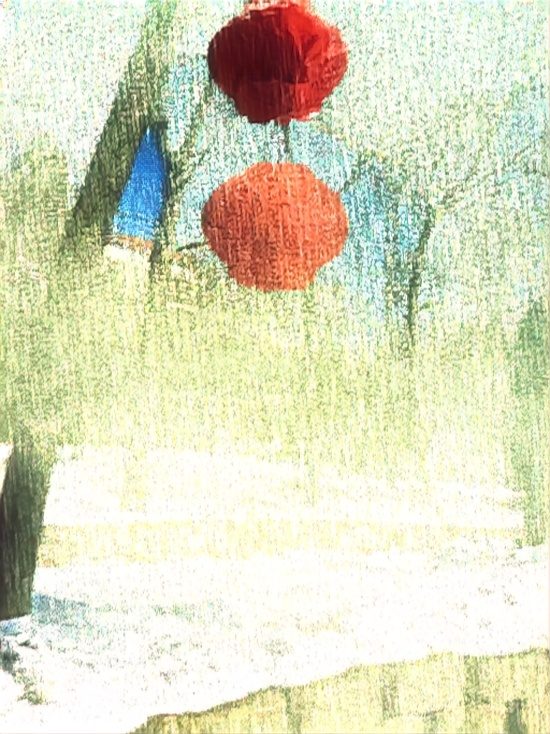}
     \includegraphics[width=0.11\textwidth]{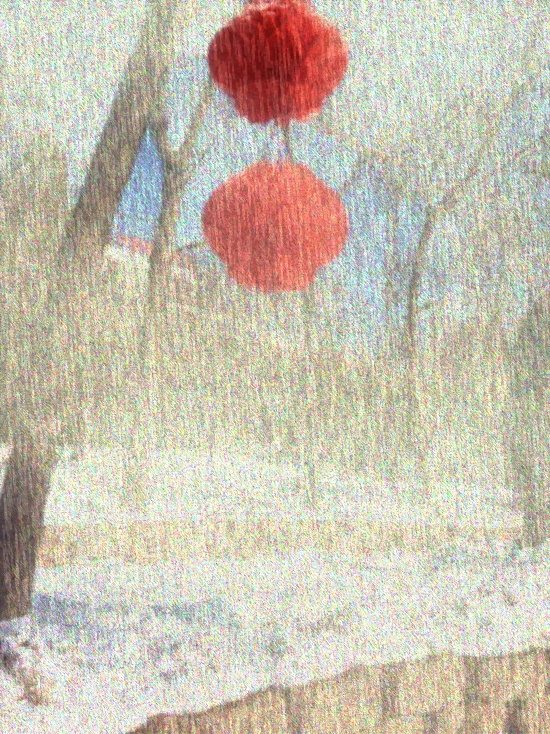}
    \includegraphics[width=0.11\textwidth]{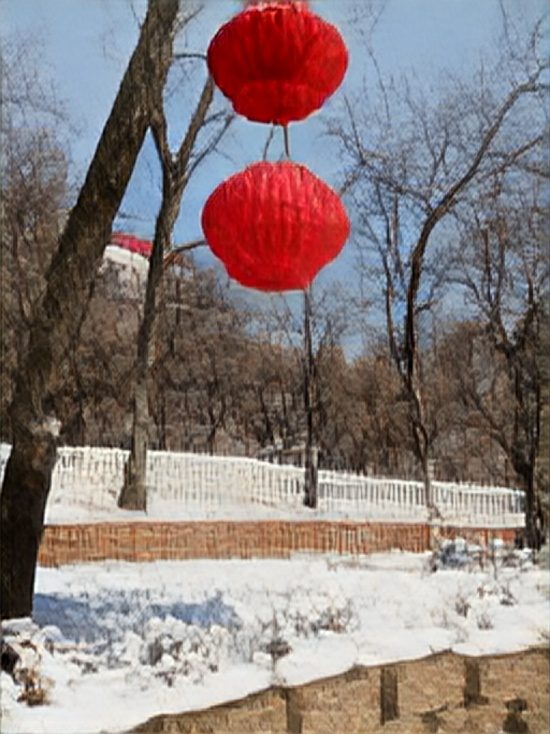}
    \includegraphics[width=0.11\textwidth]{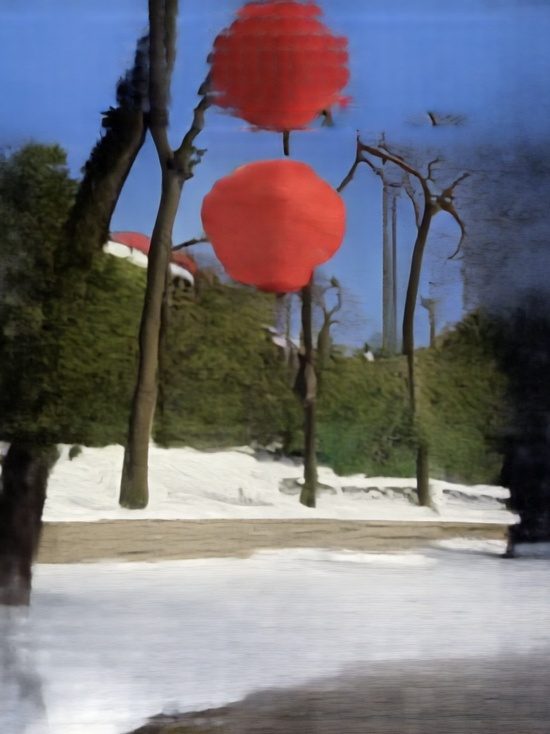}
    \includegraphics[width=0.11\textwidth]{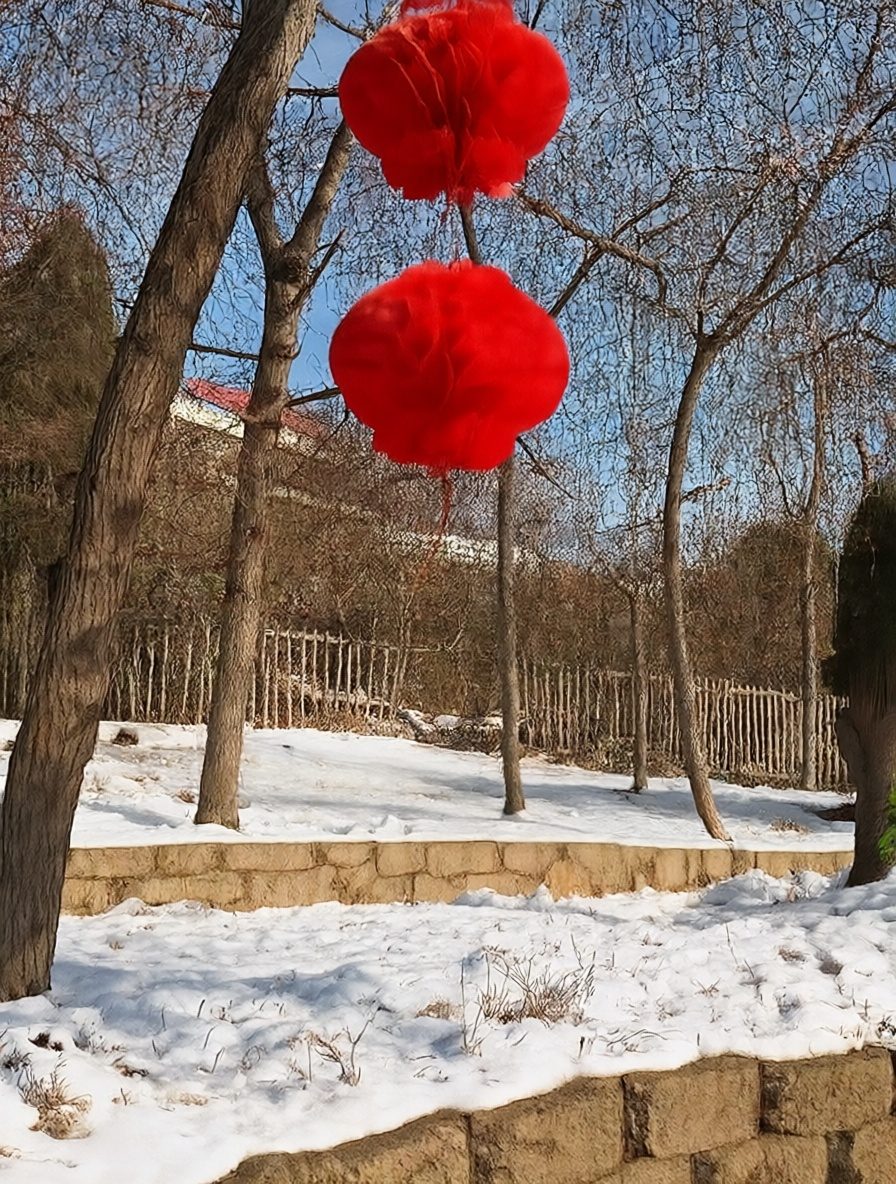}
    \includegraphics[width=0.11\textwidth]{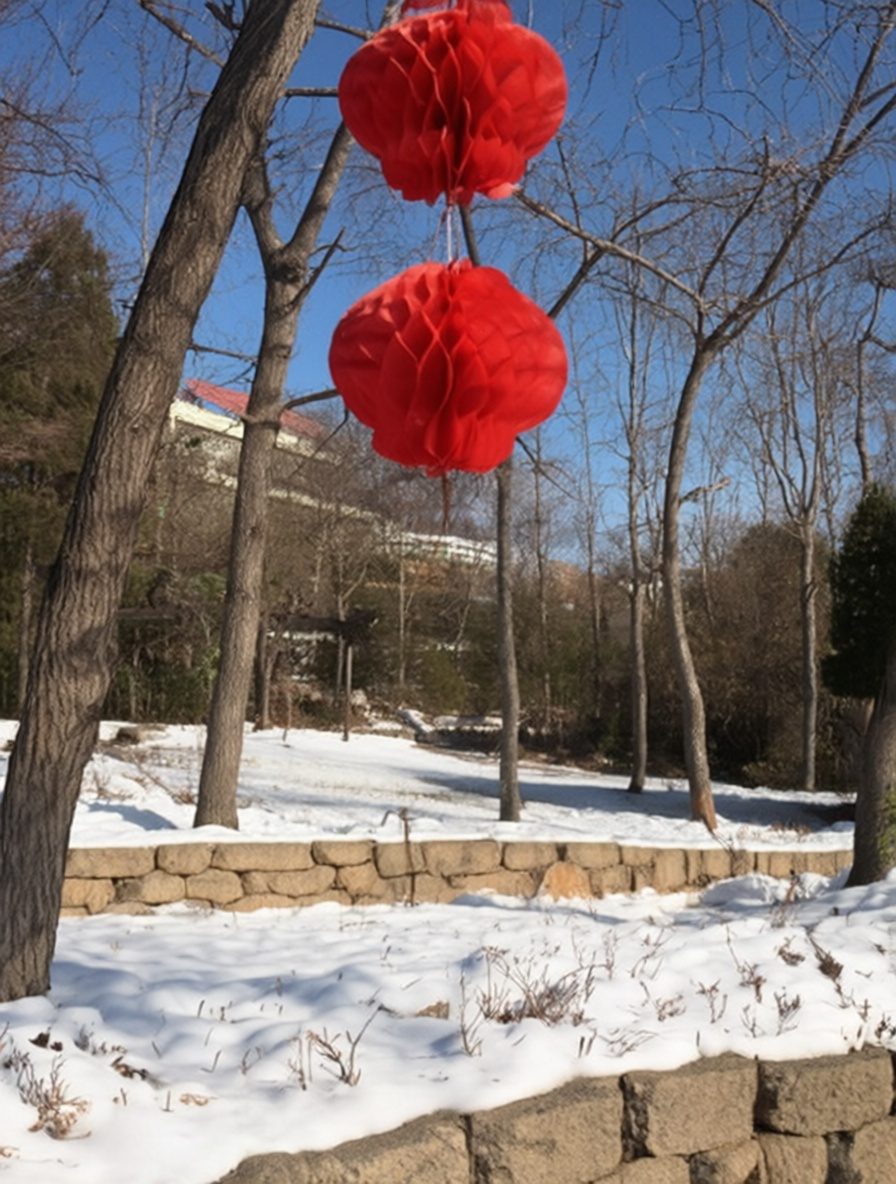}
    \includegraphics[width=0.11\textwidth]{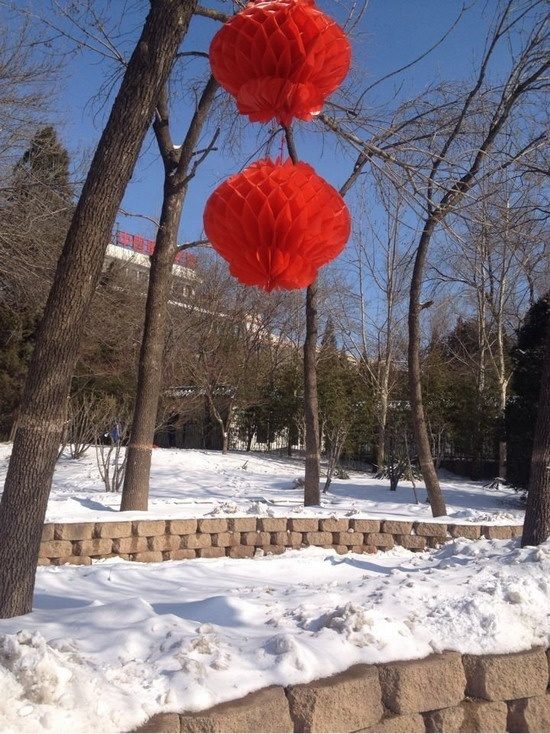}  
    

    
    \includegraphics[width=0.11\textwidth]{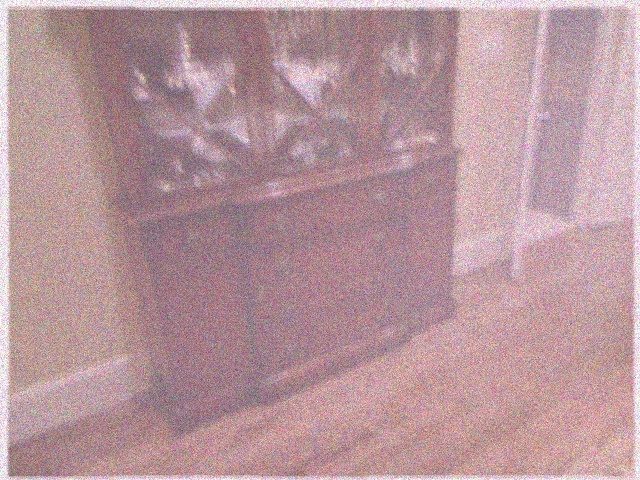}
    \includegraphics[width=0.11\textwidth]{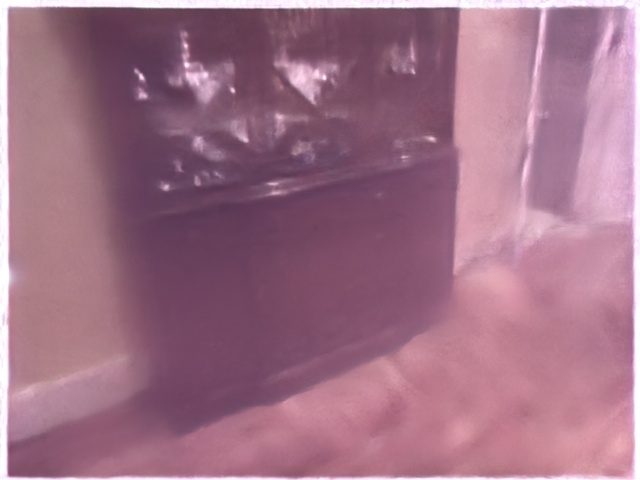}
    \includegraphics[width=0.11\textwidth]{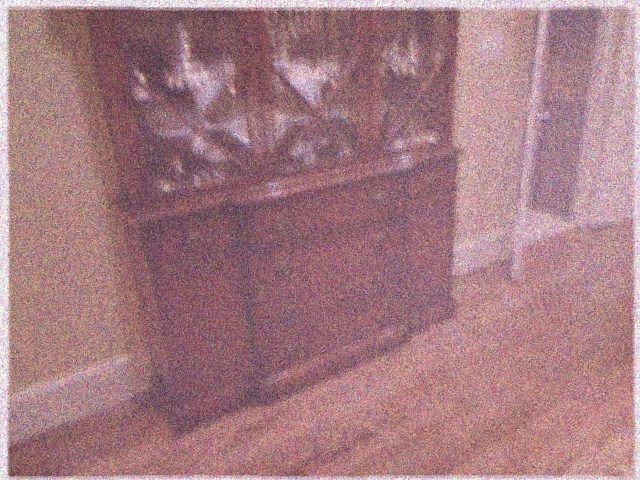}    
    \includegraphics[width=0.11\textwidth]{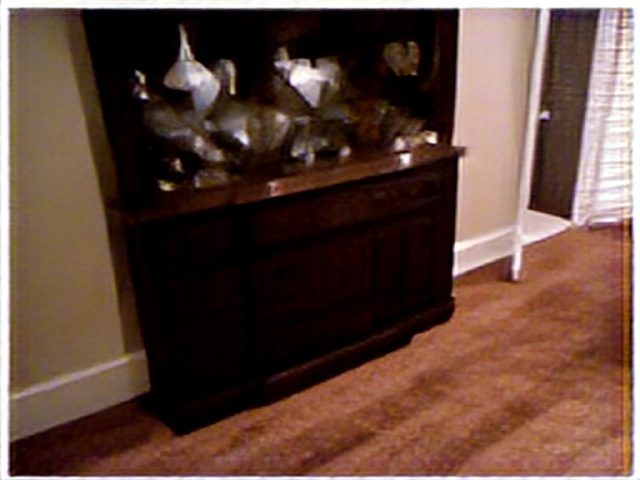}
    \includegraphics[width=0.11\textwidth]{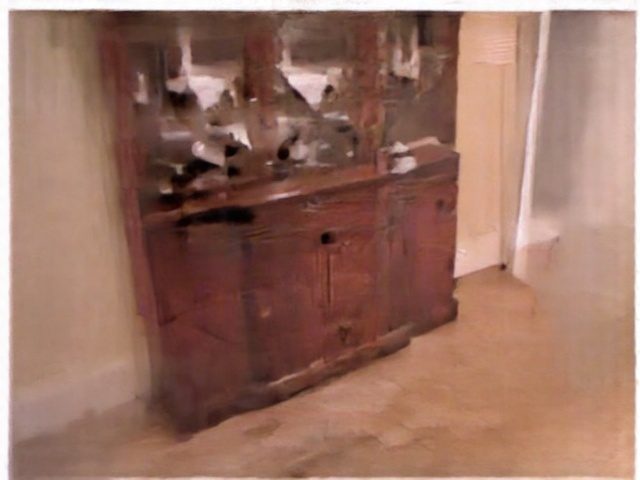}
    \includegraphics[width=0.11\textwidth]{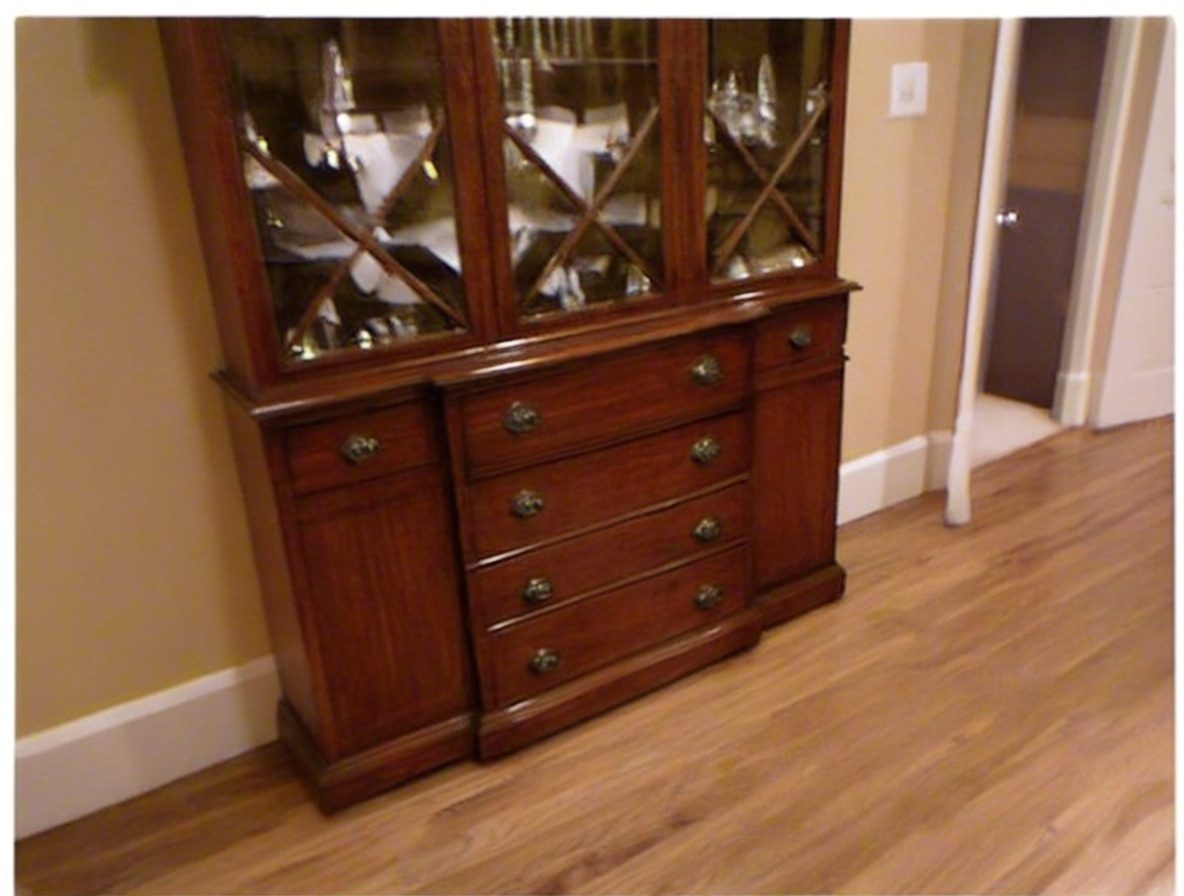}
    \includegraphics[width=0.11\textwidth]{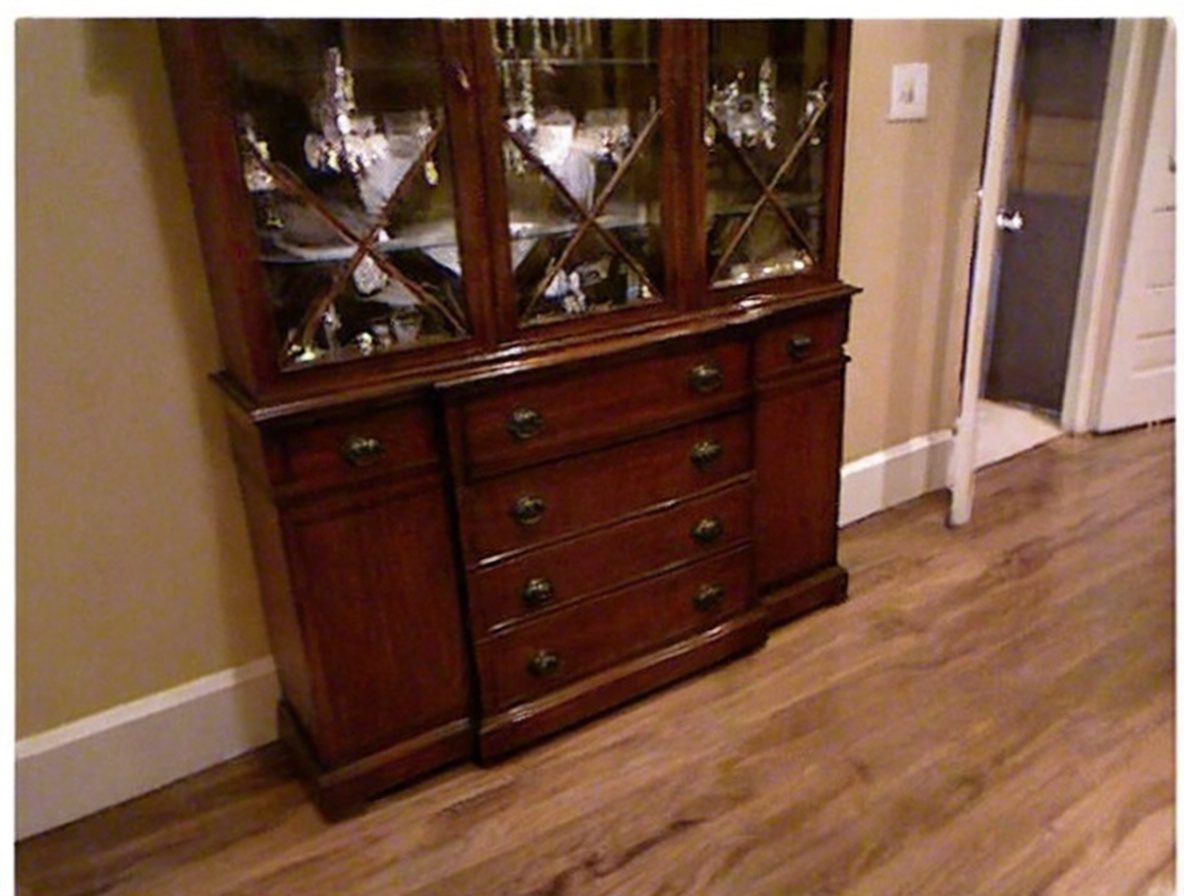}
    \includegraphics[width=0.11\textwidth]{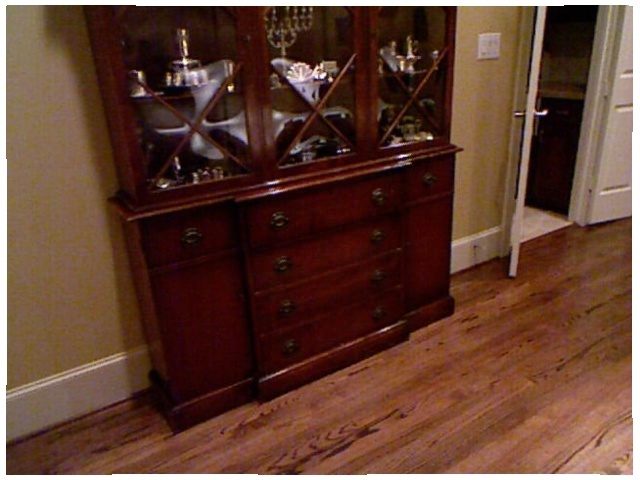}

    \includegraphics[width=0.11\textwidth]{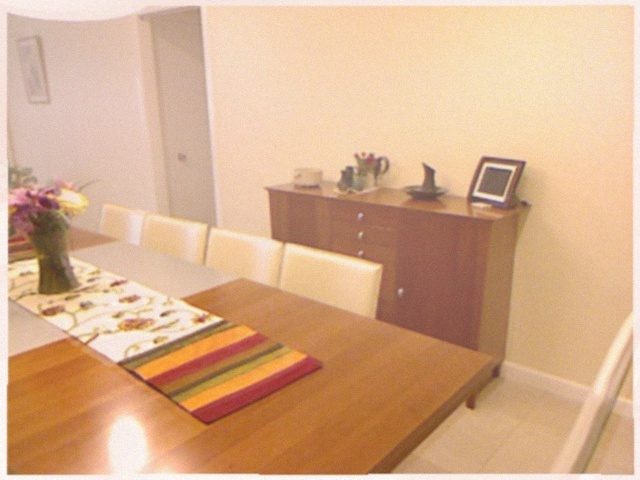}
    \includegraphics[width=0.11\textwidth]{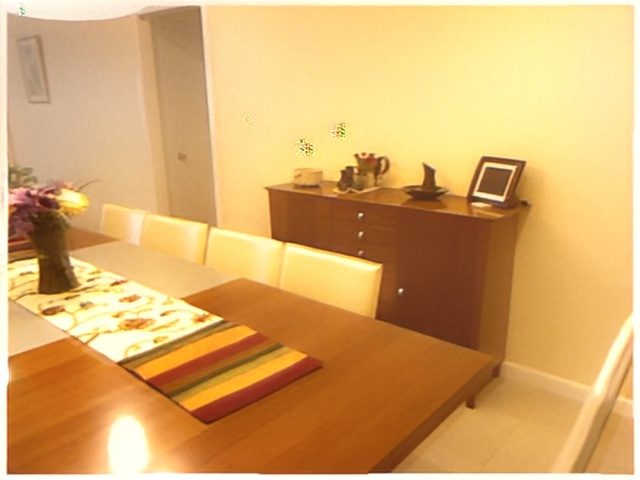}
    \includegraphics[width=0.11\textwidth]{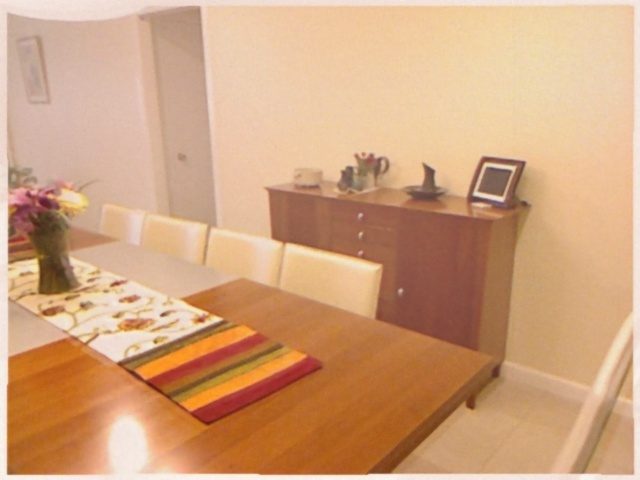}    
    \includegraphics[width=0.11\textwidth]{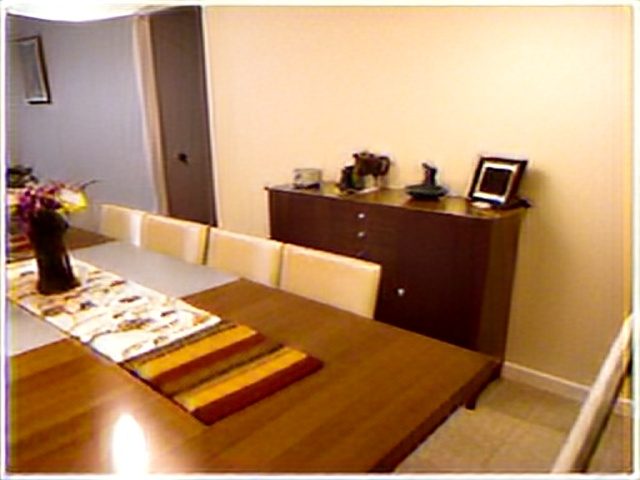}
    \includegraphics[width=0.11\textwidth]{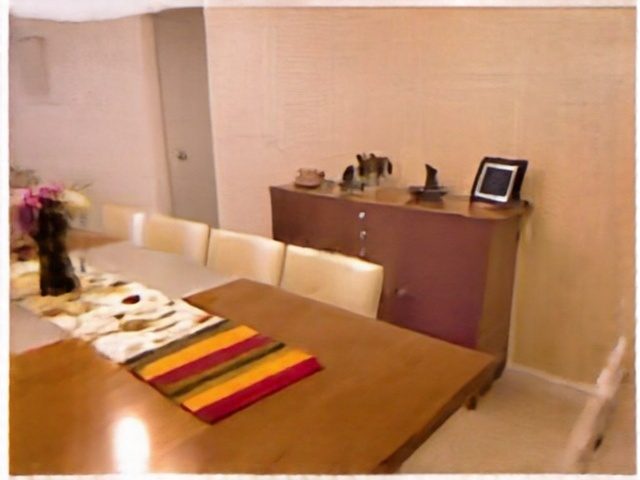}
    \includegraphics[width=0.11\textwidth]{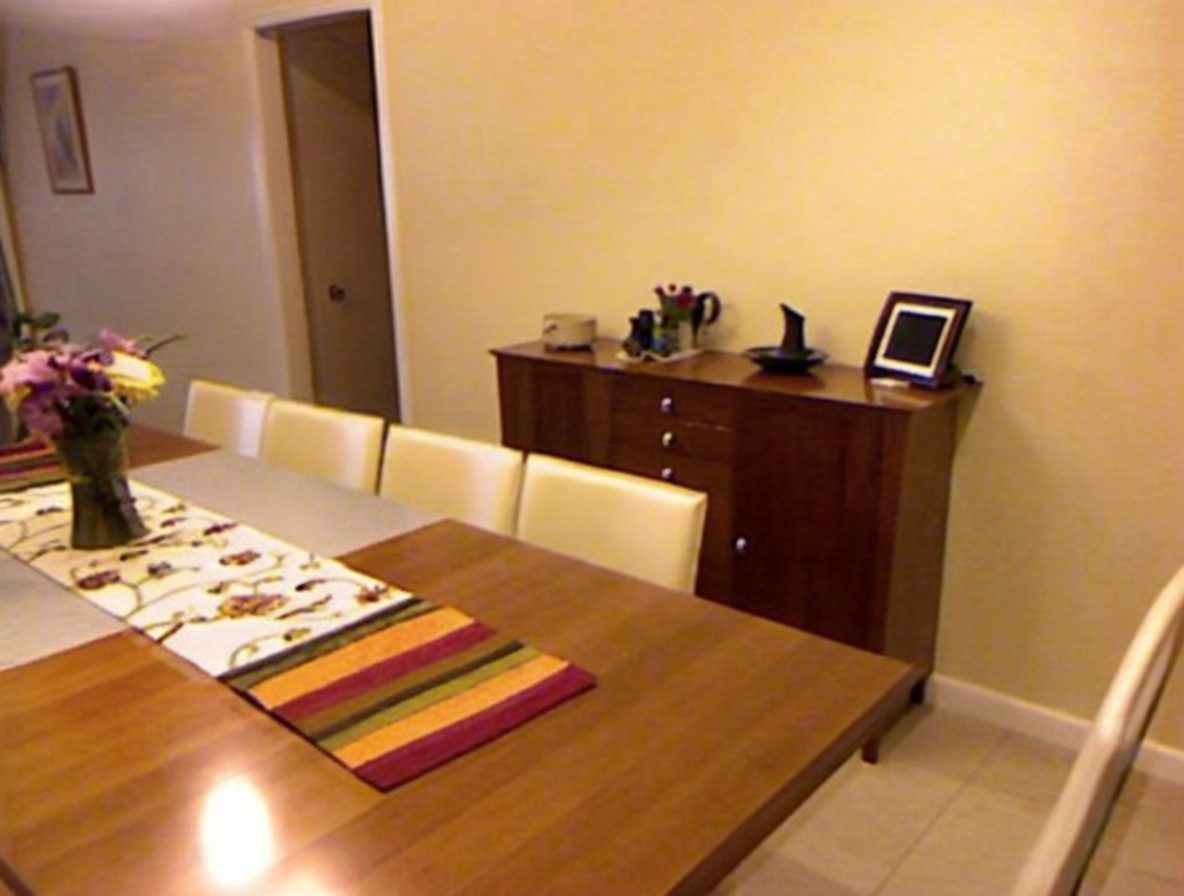}
    \includegraphics[width=0.11\textwidth]{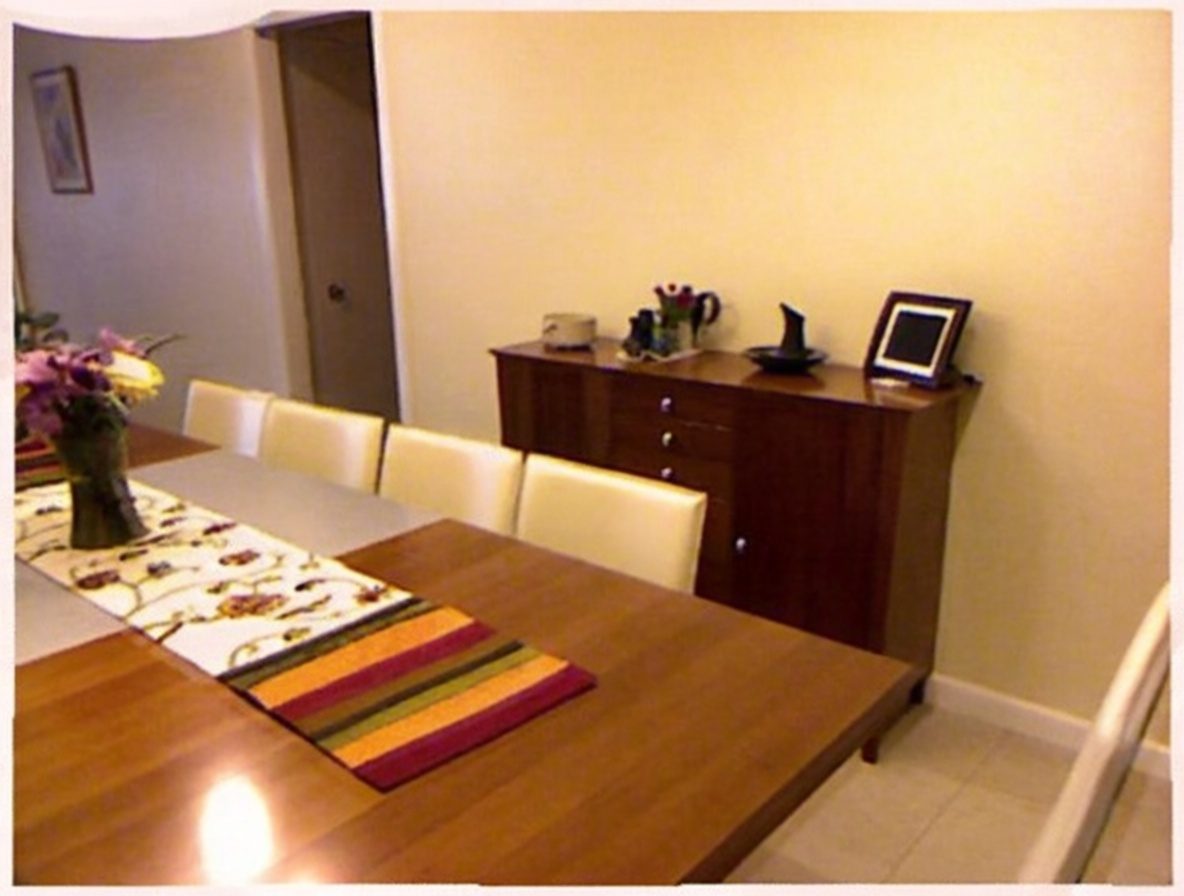}
    \includegraphics[width=0.11\textwidth]{figures/RESIDE/RESIDE_test_gt_jpg/1436.jpg}
    
    \caption{Additional qualitative comparisons on challenging synthetic data}
    \label{fig:appendix_visual_synt}
\end{figure*}

\begin{figure*}[htbp]
    \centering

    \begin{subfigure}{0.12\textwidth}
        \centering
        \textbf{Input}
    \end{subfigure}
    \begin{subfigure}{0.12\textwidth}
        \centering
        \textbf{3D}
    \end{subfigure}
    \begin{subfigure}{0.12\textwidth}
        \centering
        \textbf{FoundIR}
    \end{subfigure}    
    \begin{subfigure}{0.12\textwidth}
        \centering
        \textbf{Img2Img-Turbo}
    \end{subfigure}
    \begin{subfigure}{0.12\textwidth}
        \centering
        \textbf{Stable Diffusion3}
    \end{subfigure}
    \begin{subfigure}{0.12\textwidth}
        \centering
        \textbf{Qwen3-Image}
    \end{subfigure}
    \begin{subfigure}{0.12\textwidth}
        \centering
        \textbf{(Ours) R\&R}
    \end{subfigure}
    \vspace{2mm}

    \includegraphics[width=0.12\textwidth]{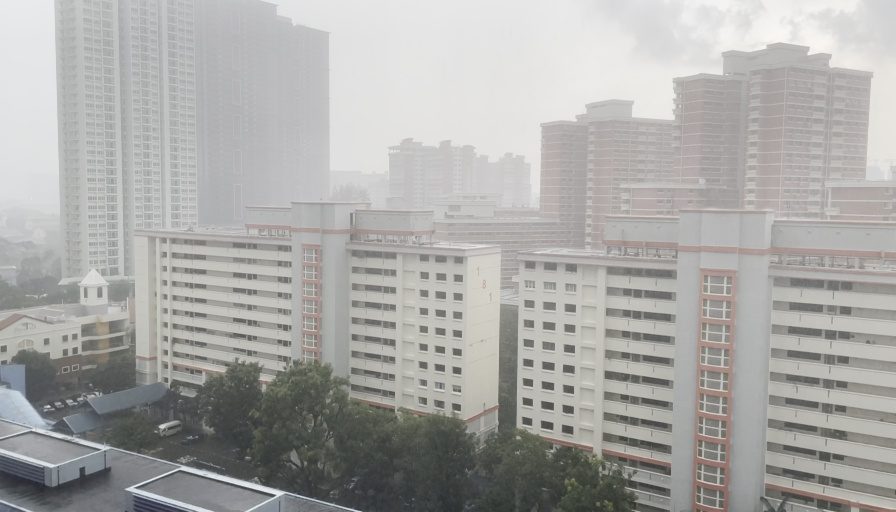}
    \includegraphics[width=0.12\textwidth]{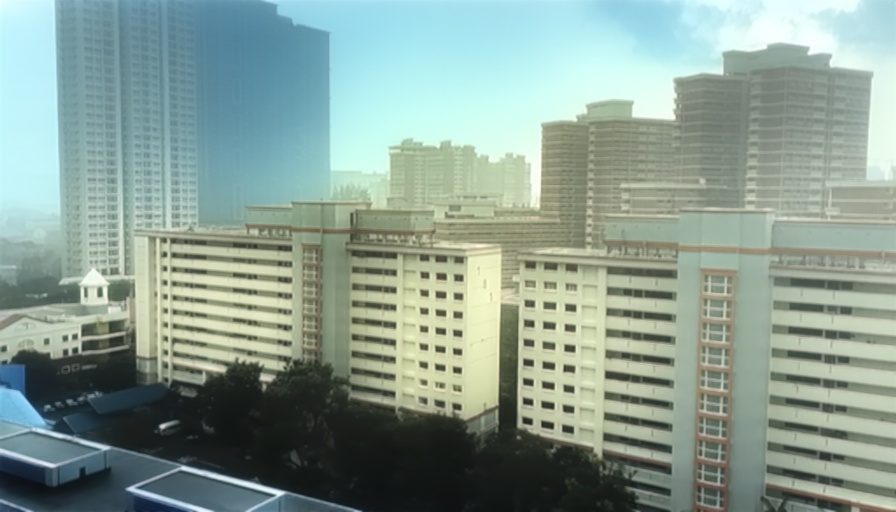}
    \includegraphics[width=0.12\textwidth]{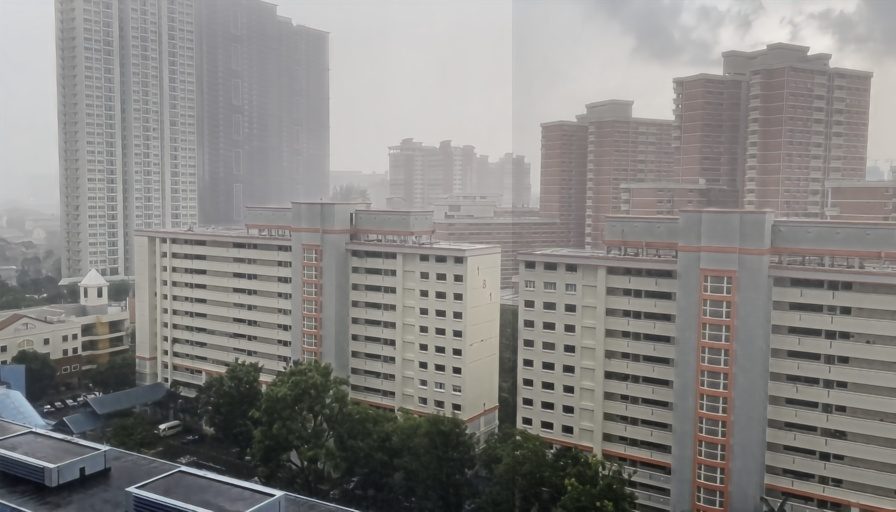}
    \includegraphics[width=0.12\textwidth]{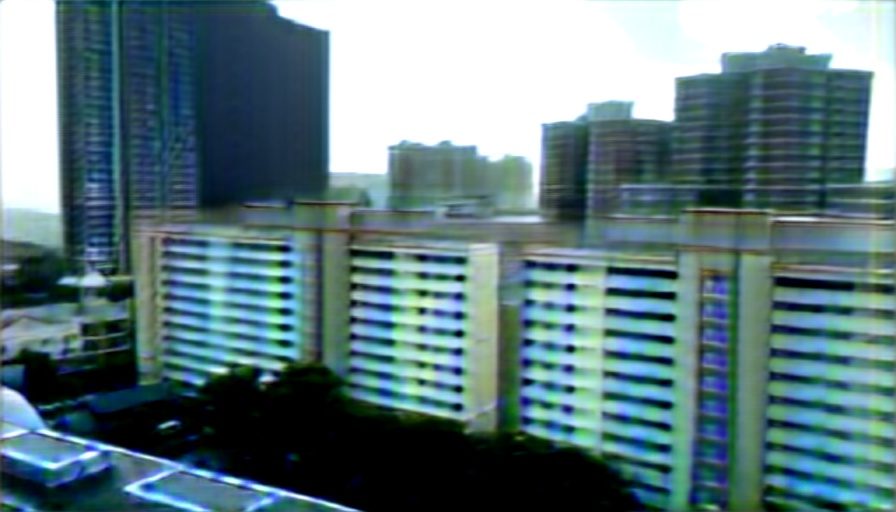}
    \includegraphics[width=0.12\textwidth]{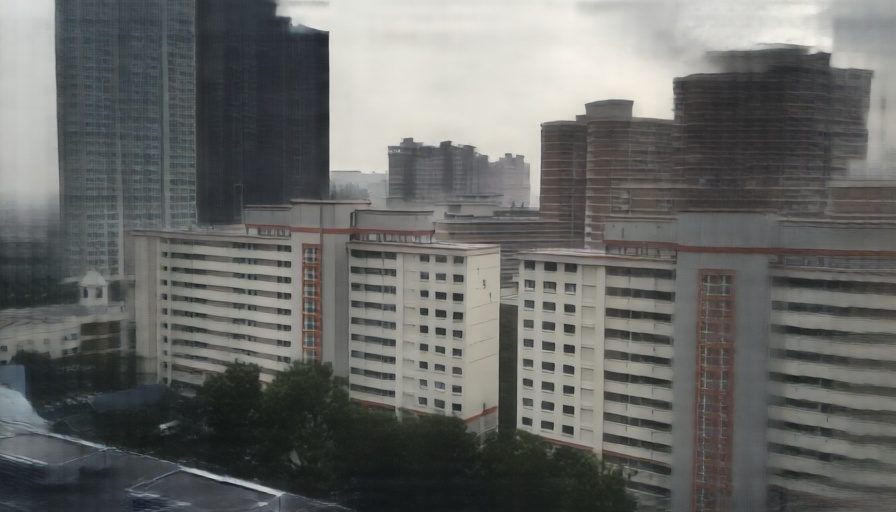}
    \includegraphics[width=0.12\textwidth]{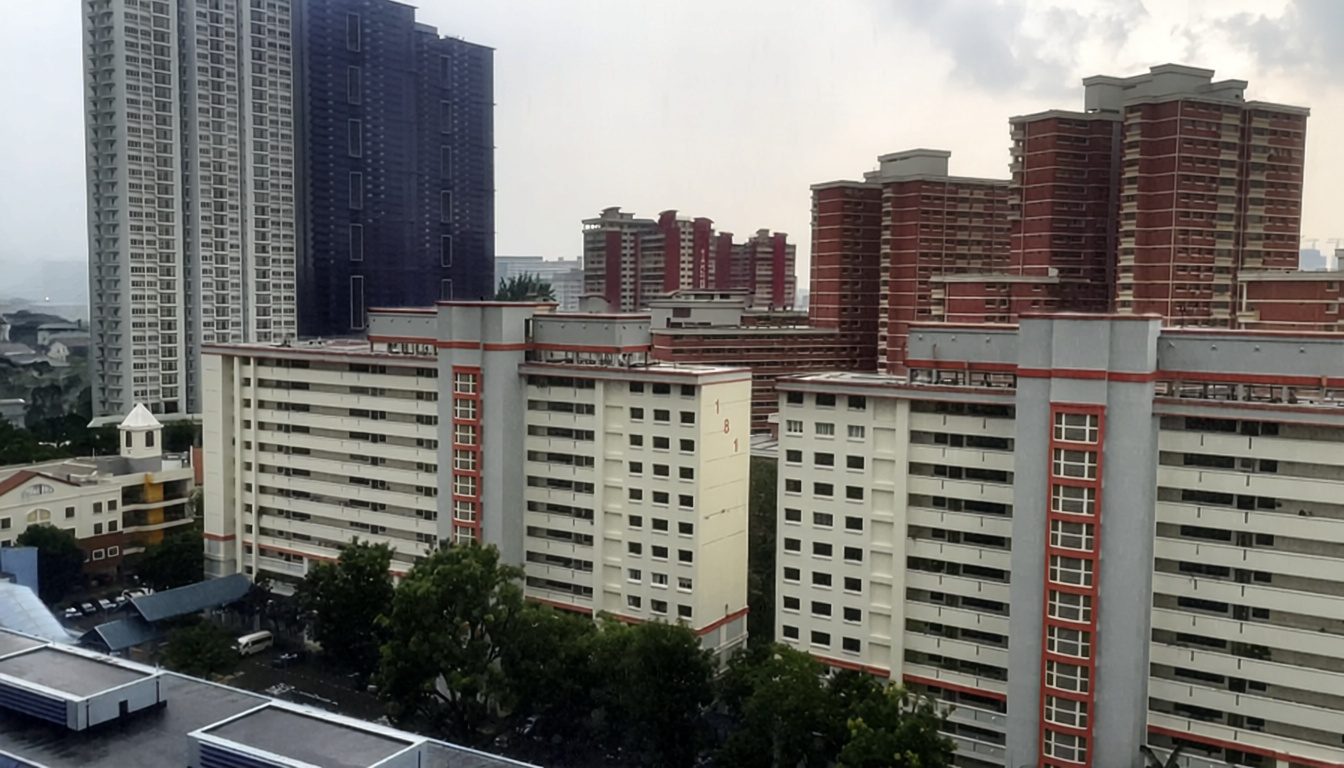}
    \includegraphics[width=0.12\textwidth]{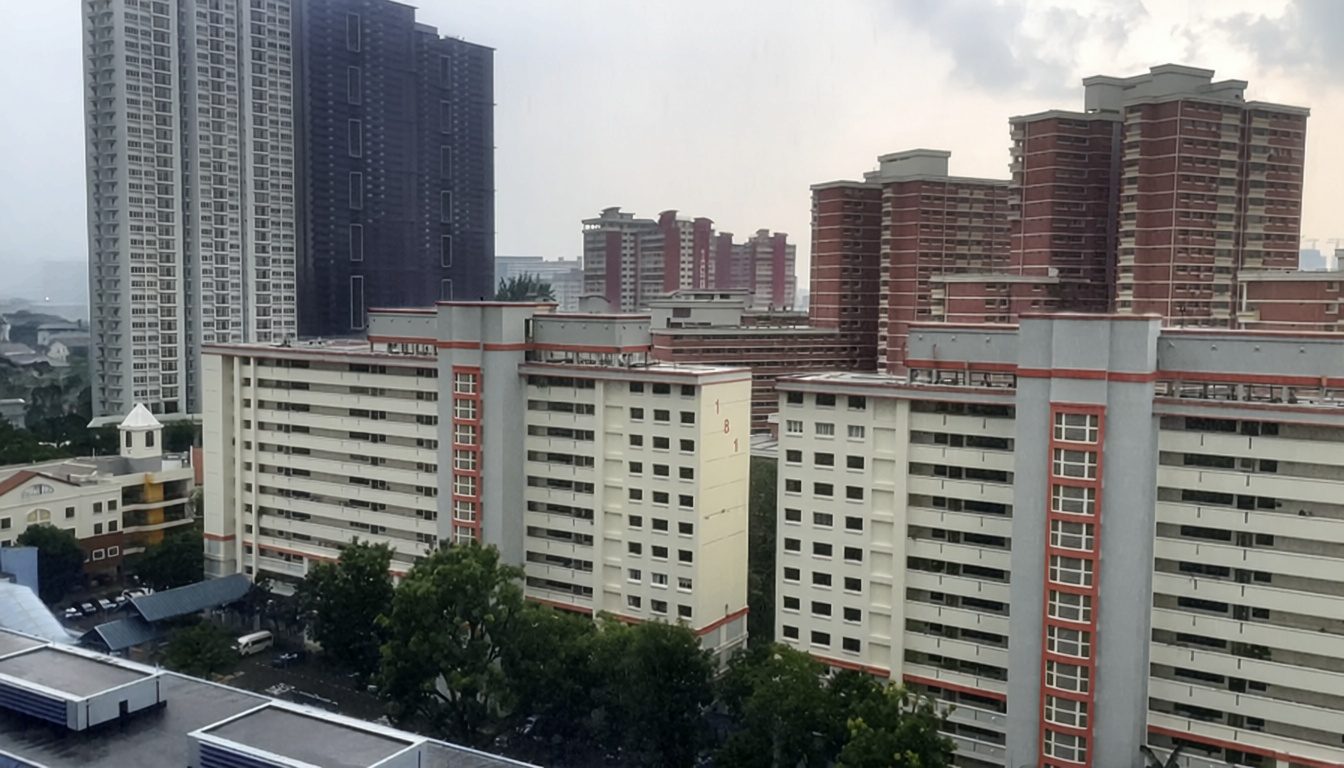}    

    \includegraphics[width=0.12\textwidth]{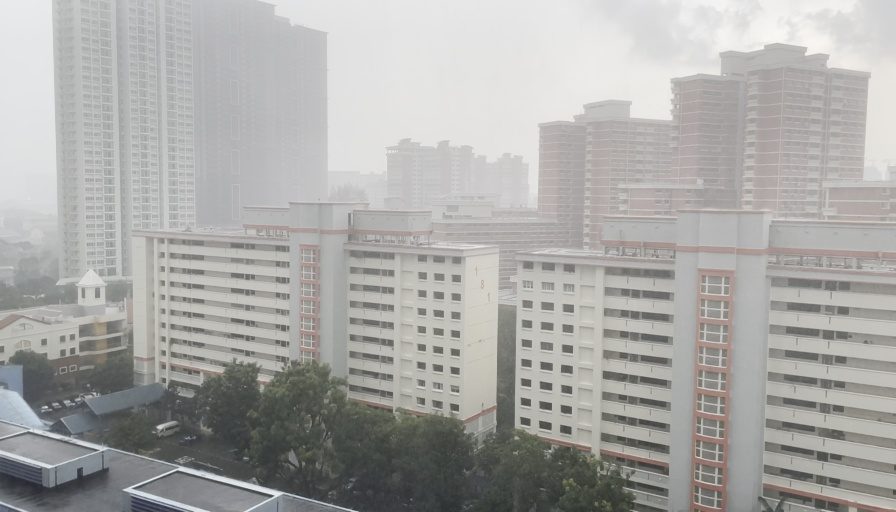}
    \includegraphics[width=0.12\textwidth]{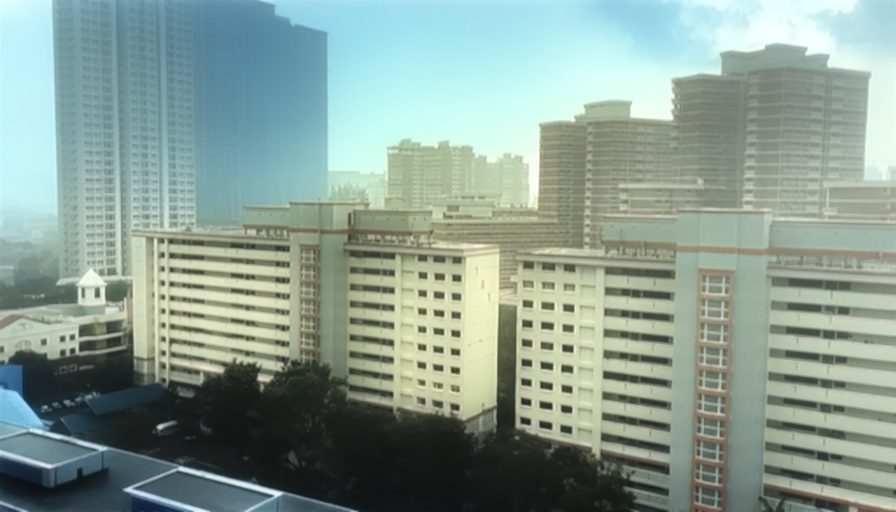}
    \includegraphics[width=0.12\textwidth]{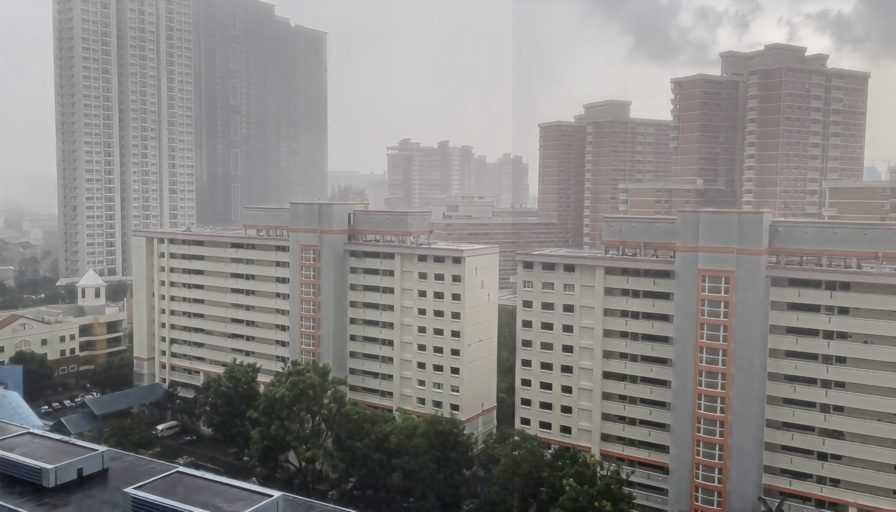}
    \includegraphics[width=0.12\textwidth]{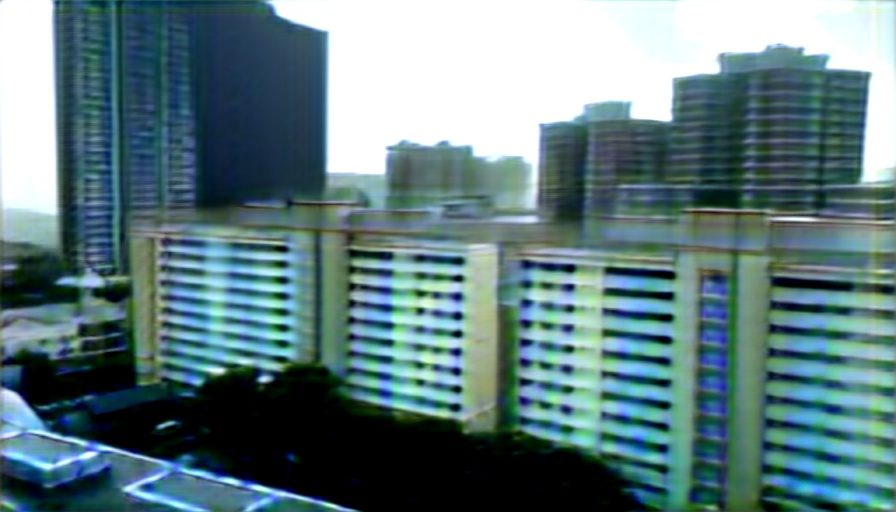}
    \includegraphics[width=0.12\textwidth]{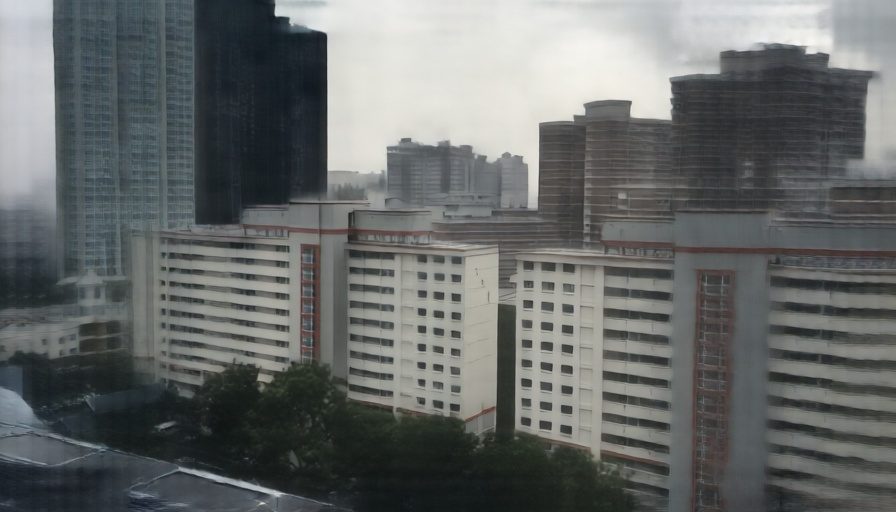}
    \includegraphics[width=0.12\textwidth]{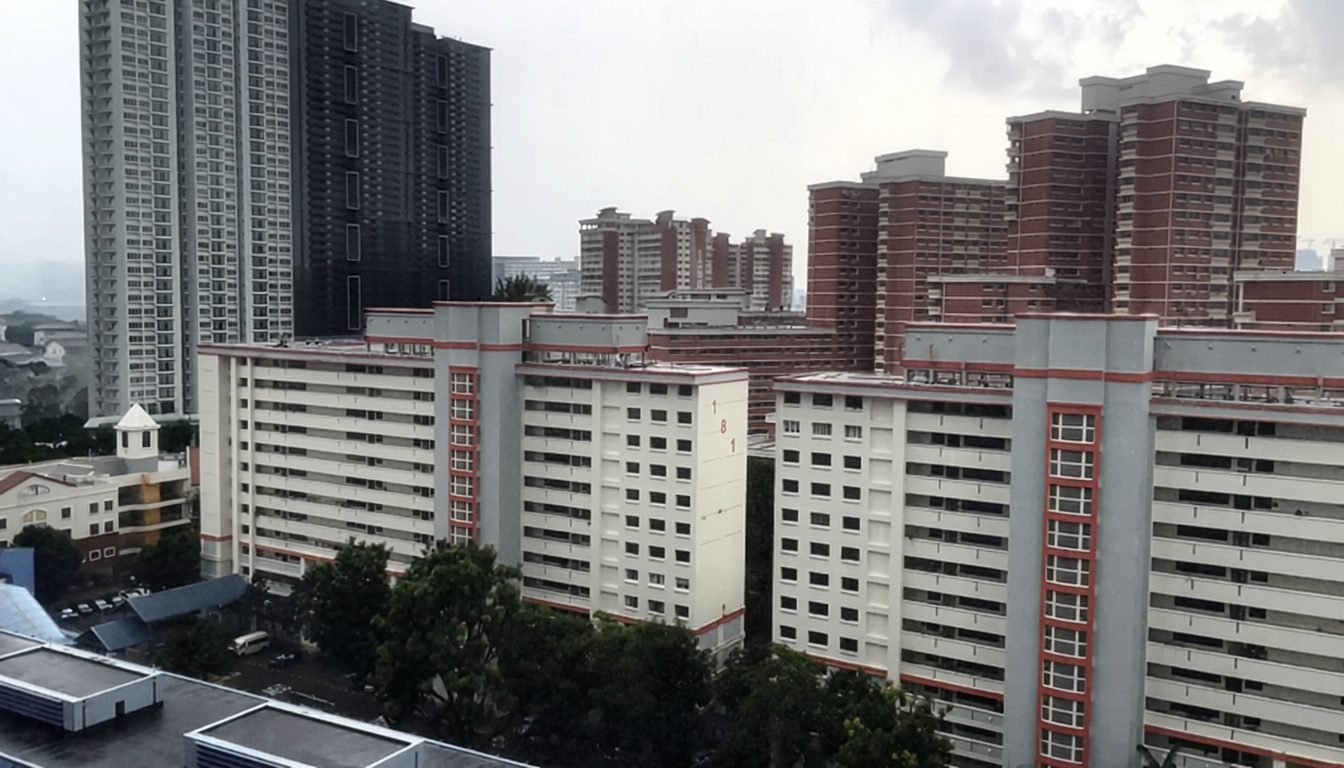}
    \includegraphics[width=0.12\textwidth]{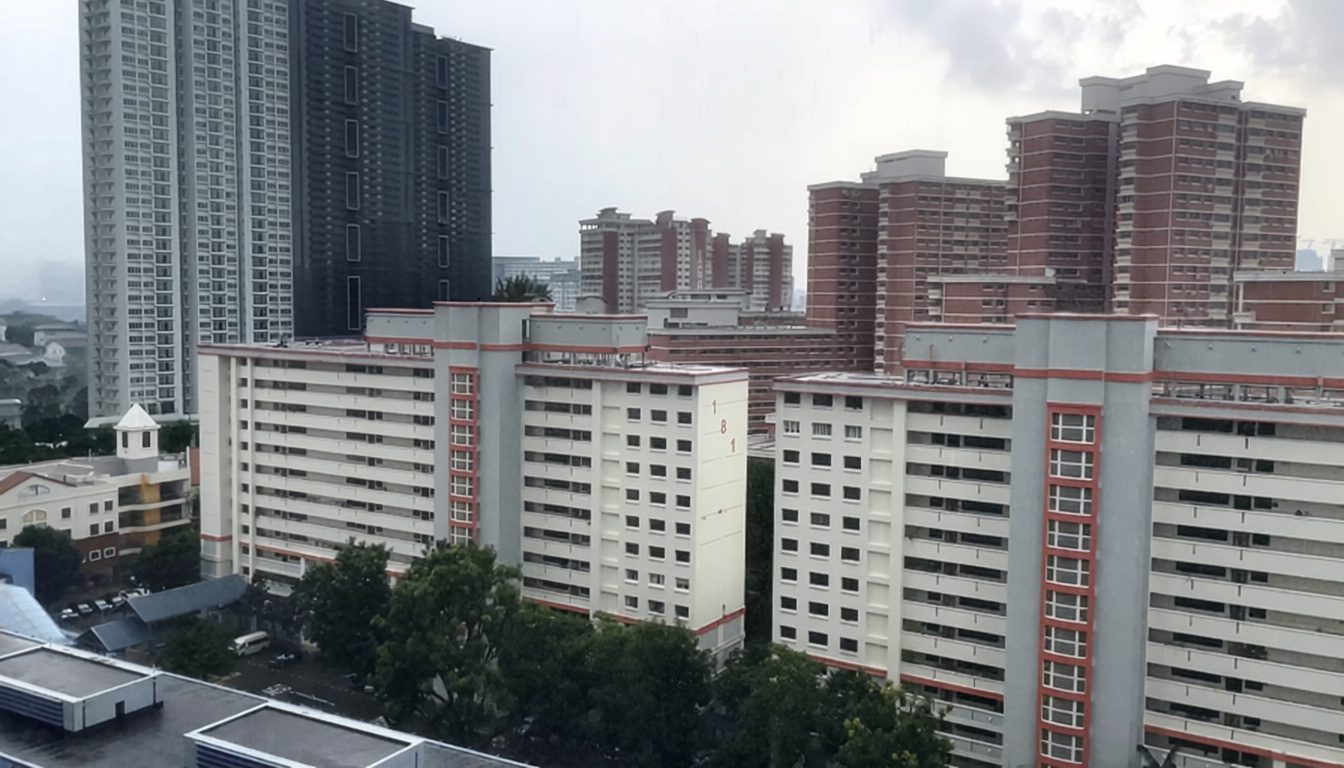}        


    \includegraphics[width=0.12\textwidth]{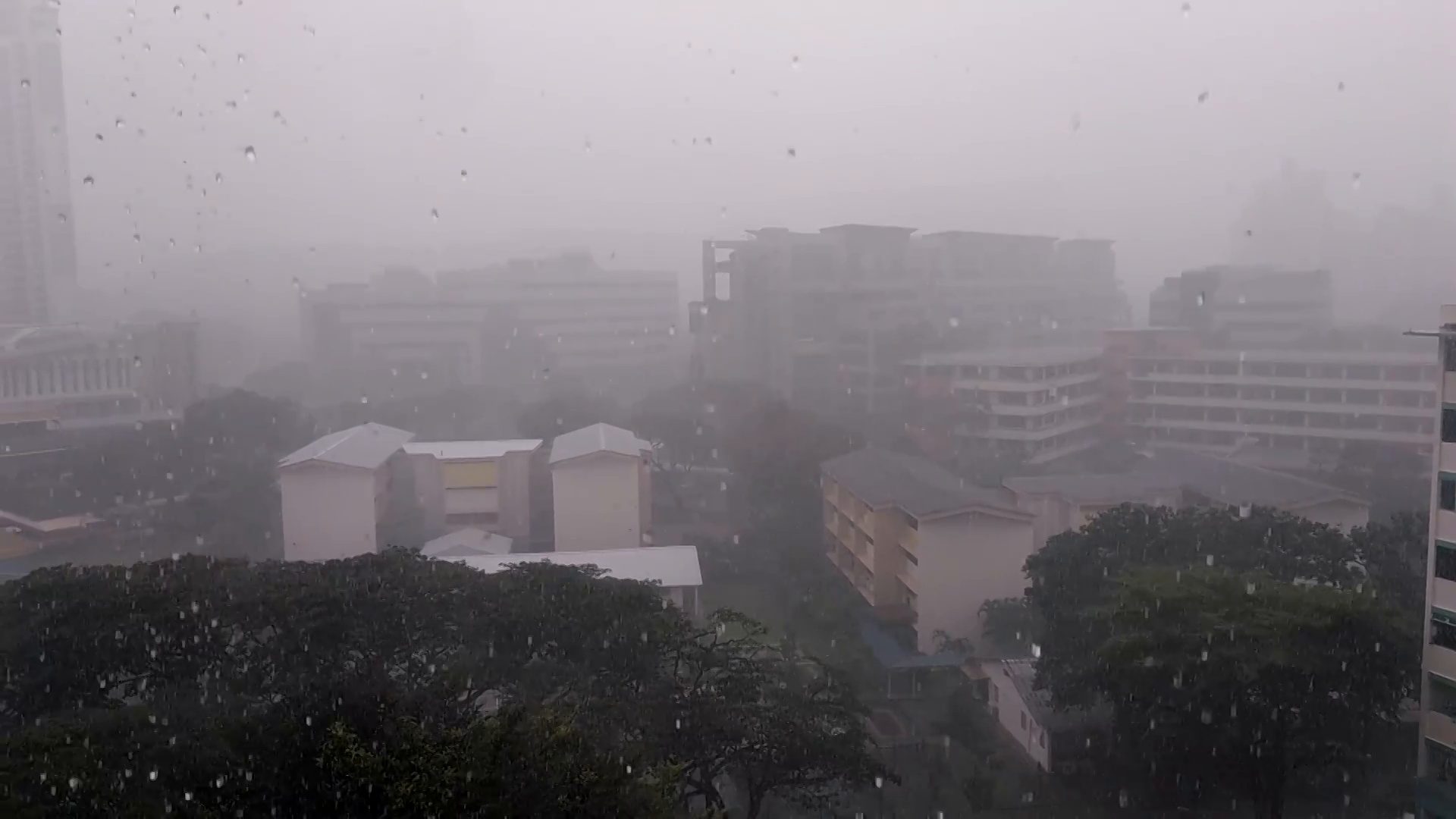}
    \includegraphics[width=0.12\textwidth]{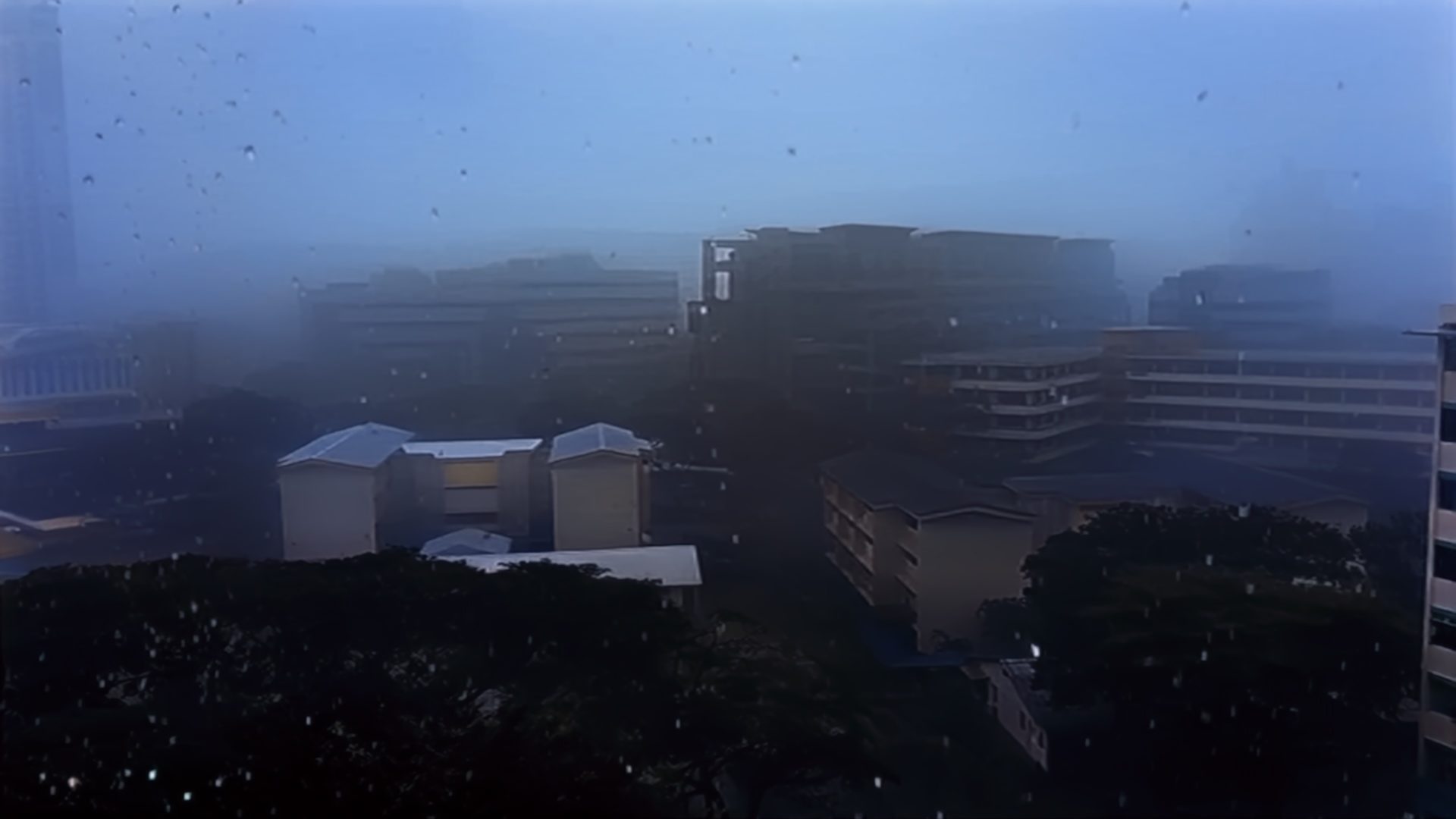}
    \includegraphics[width=0.12\textwidth]{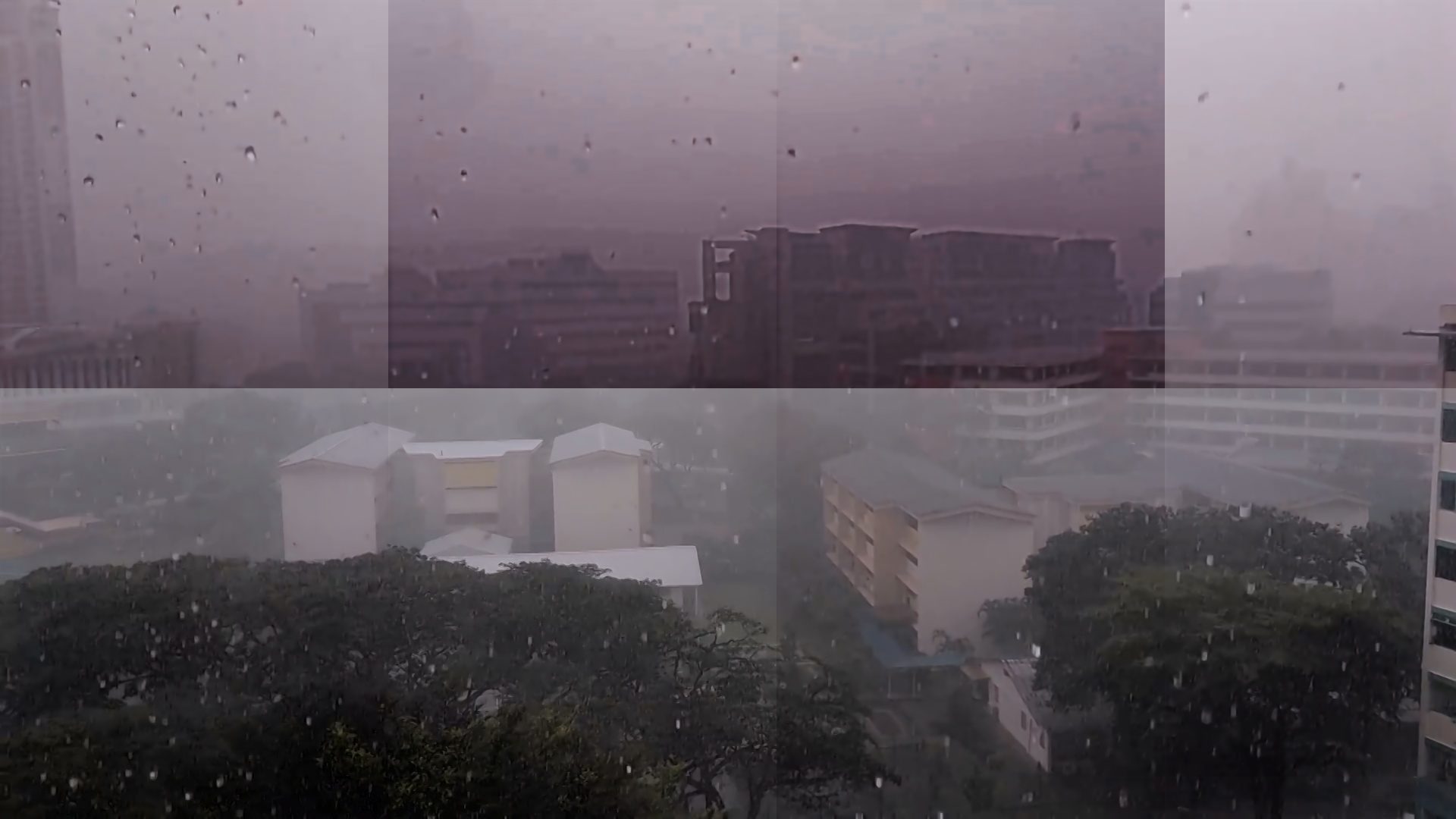}
    \includegraphics[width=0.12\textwidth]{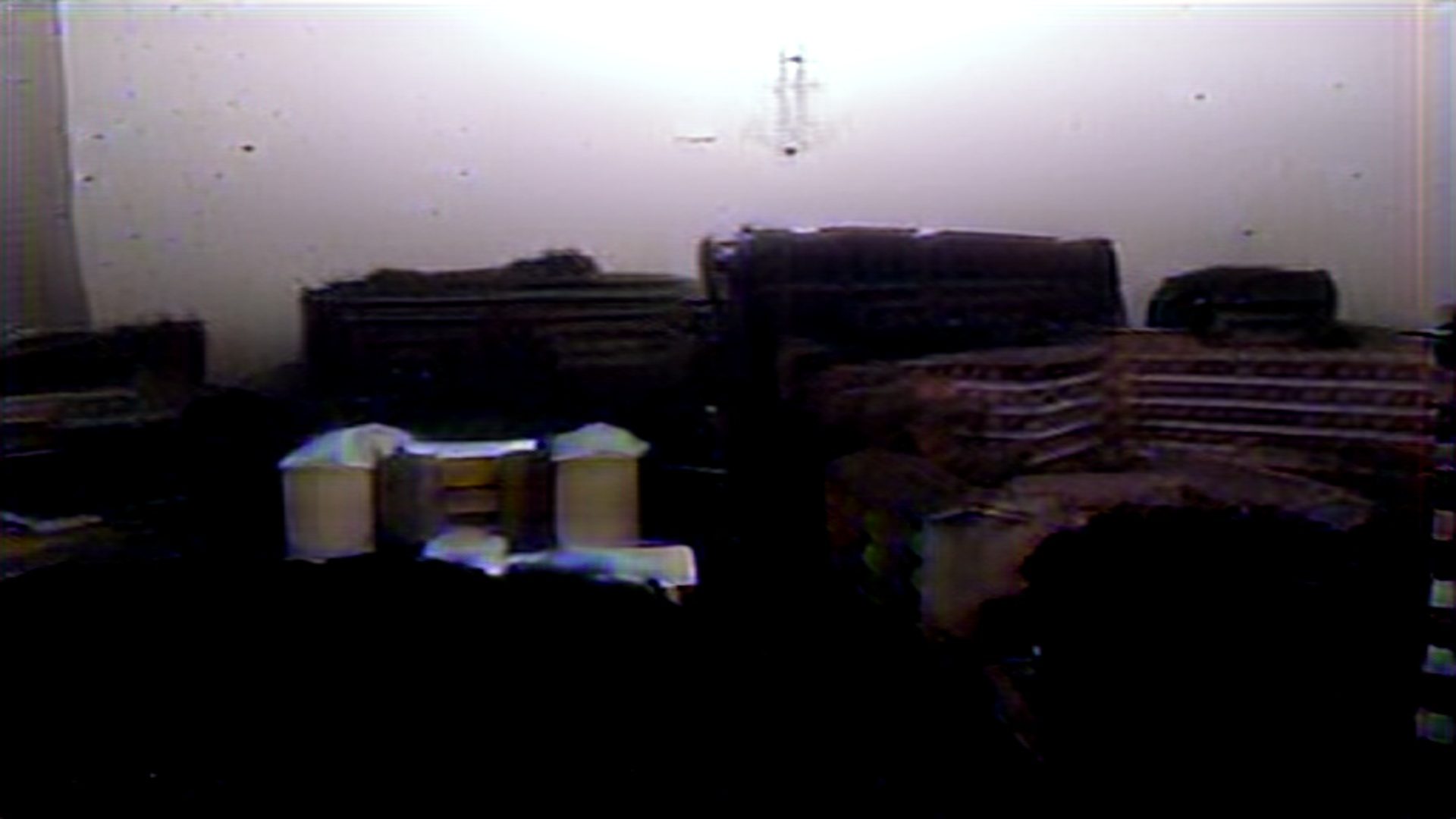}
    \includegraphics[width=0.12\textwidth]{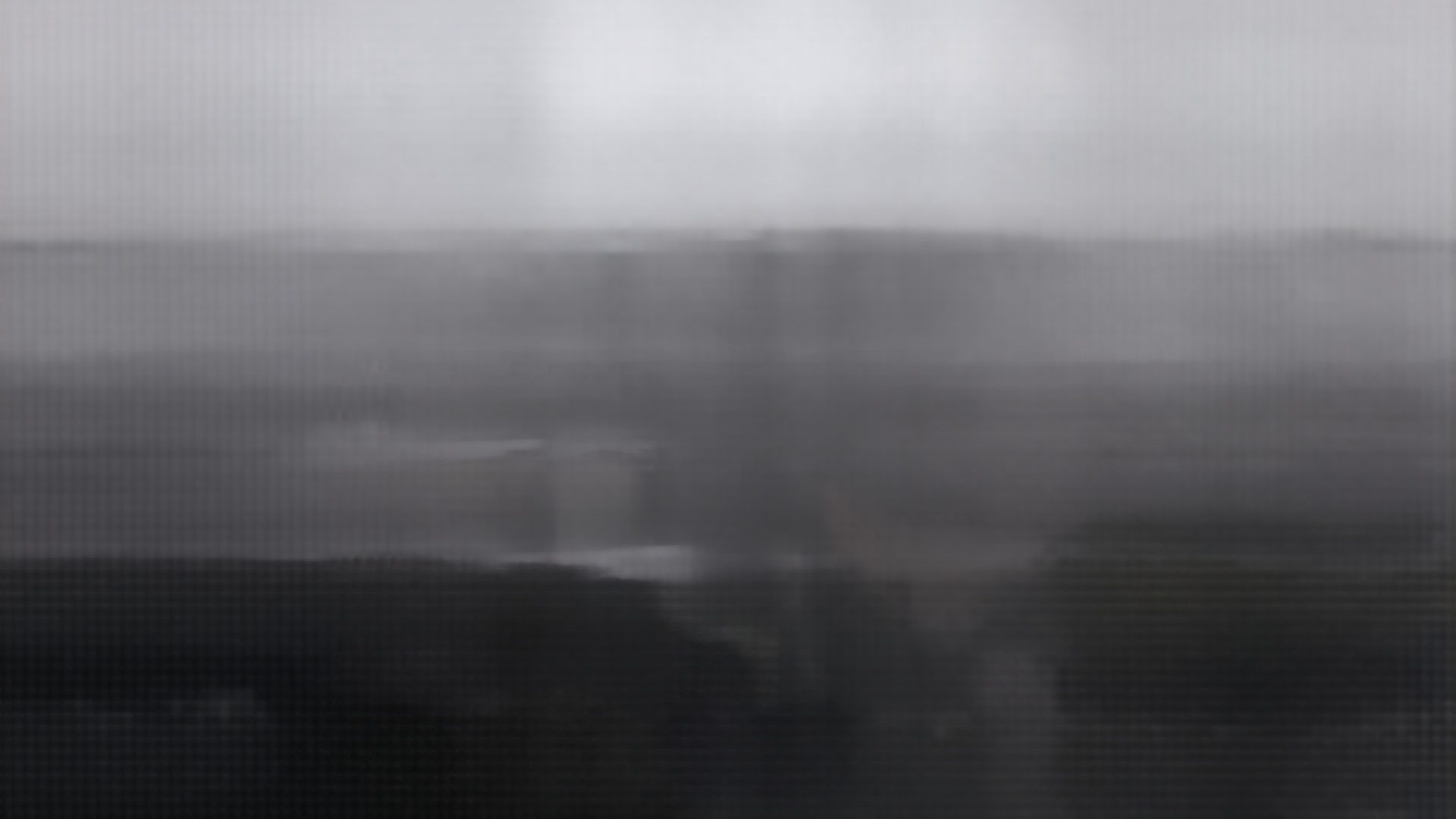}
    \includegraphics[width=0.12\textwidth]{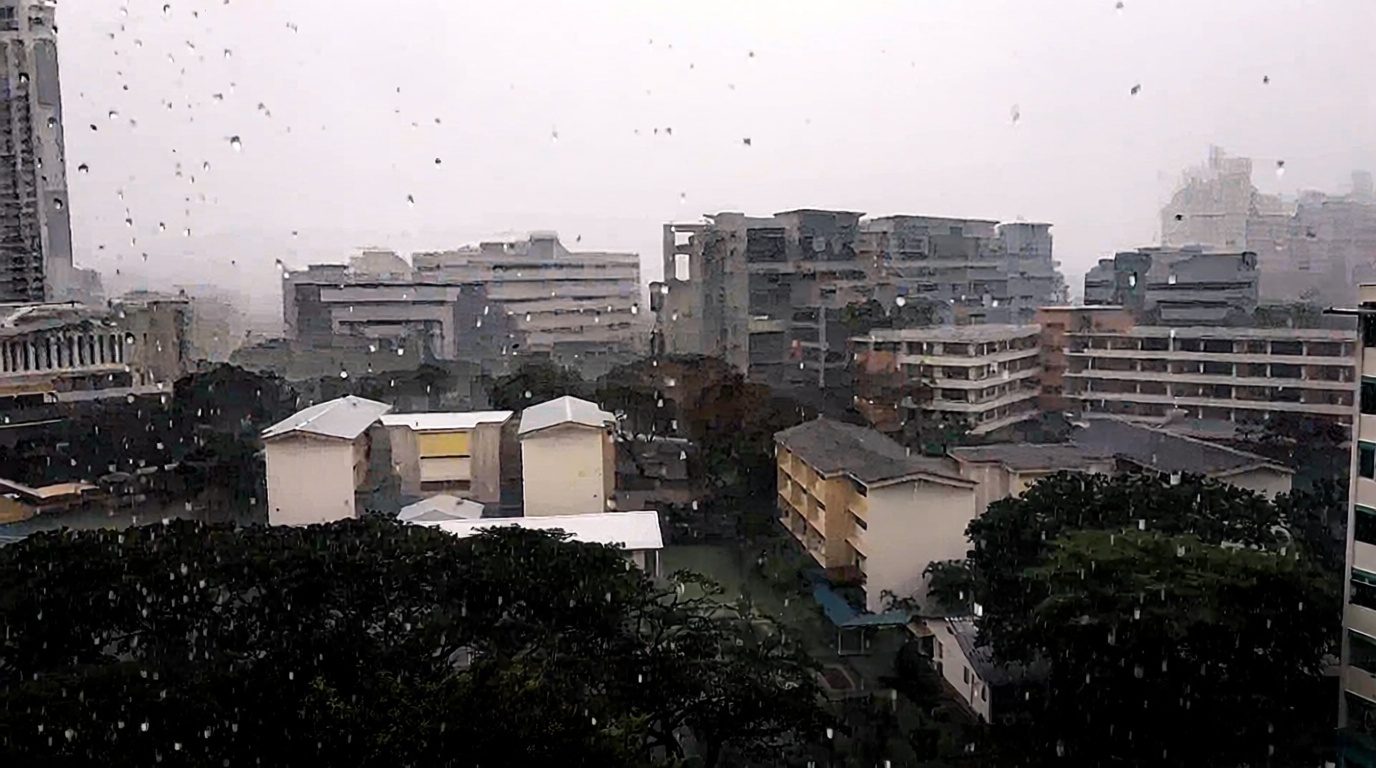}
    \includegraphics[width=0.12\textwidth]{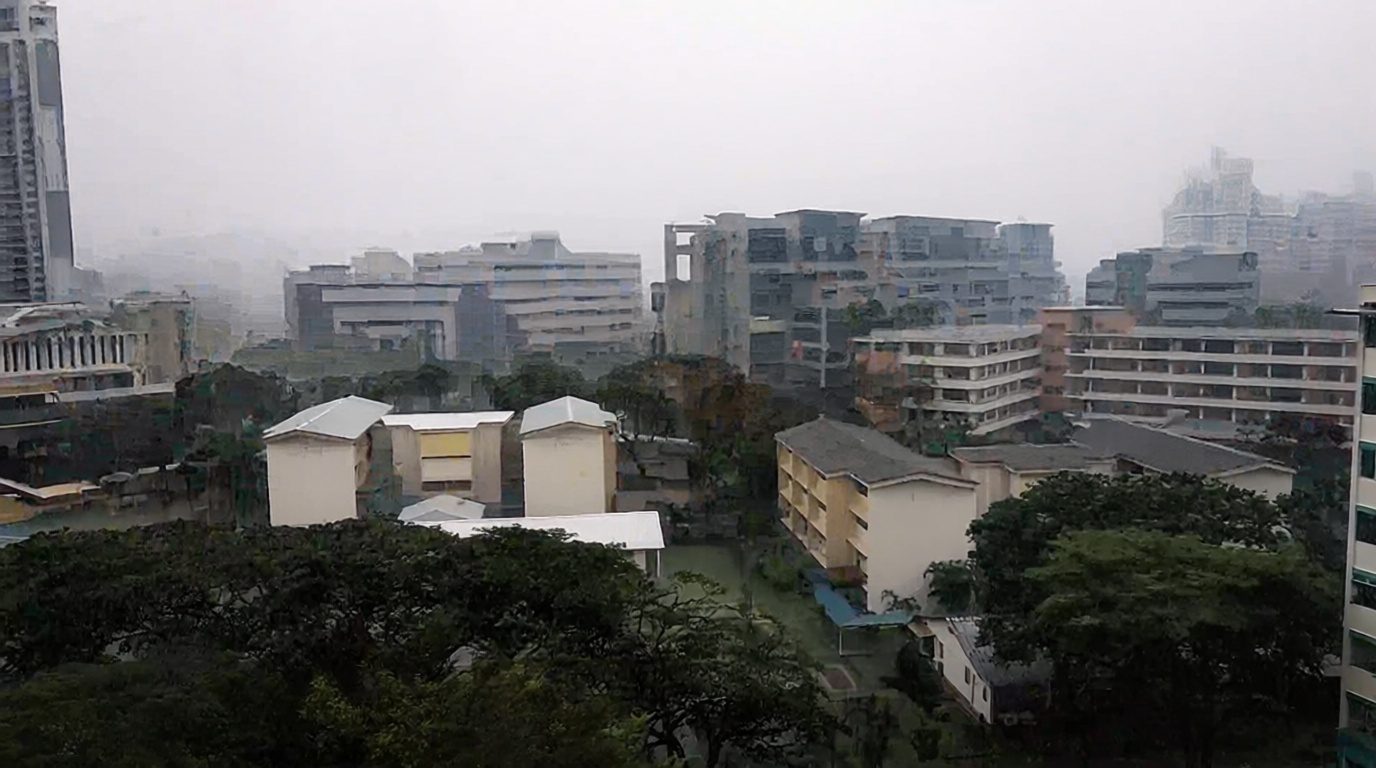}    

    \includegraphics[width=0.12\textwidth]{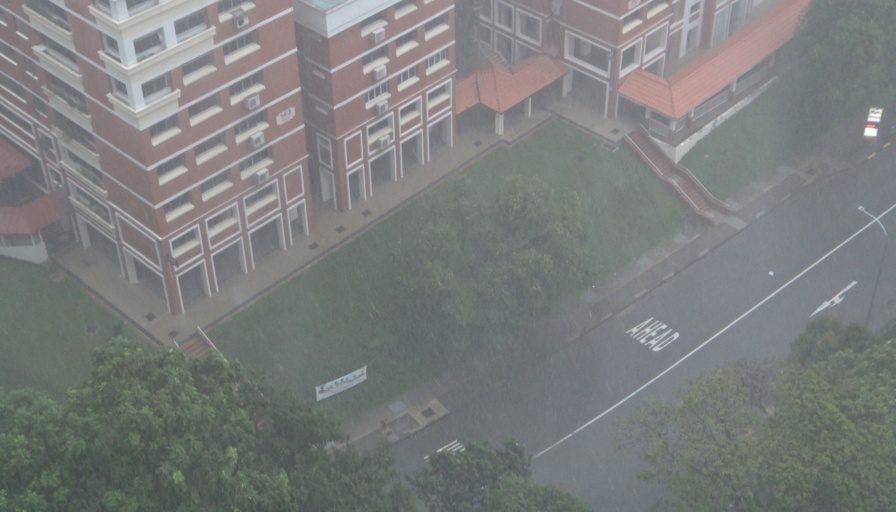}
    \includegraphics[width=0.12\textwidth]{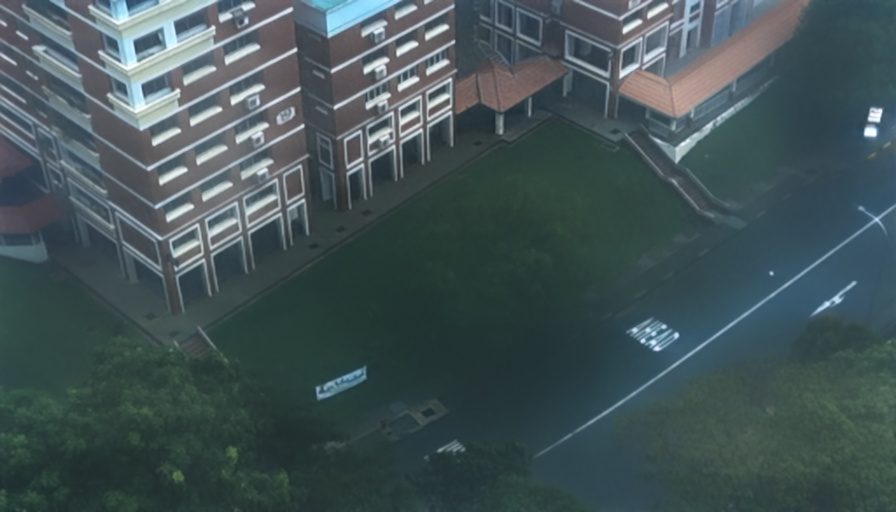}
    \includegraphics[width=0.12\textwidth]{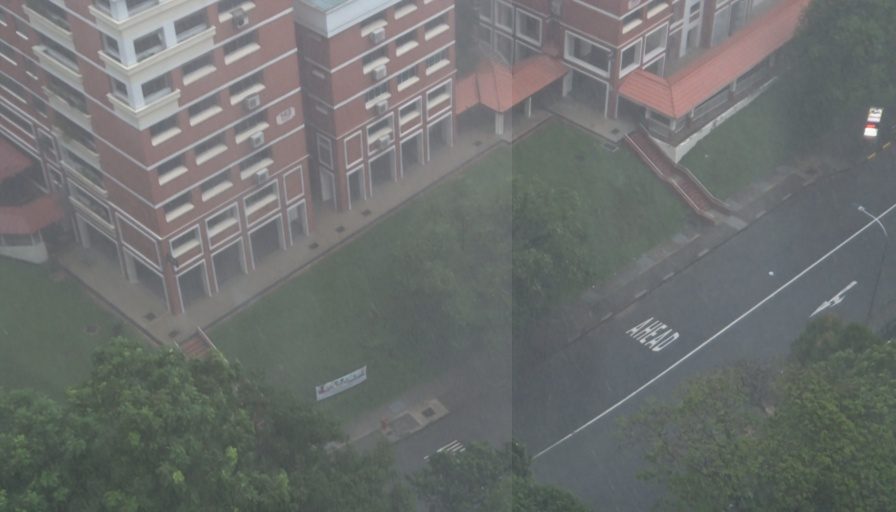}
    \includegraphics[width=0.12\textwidth]{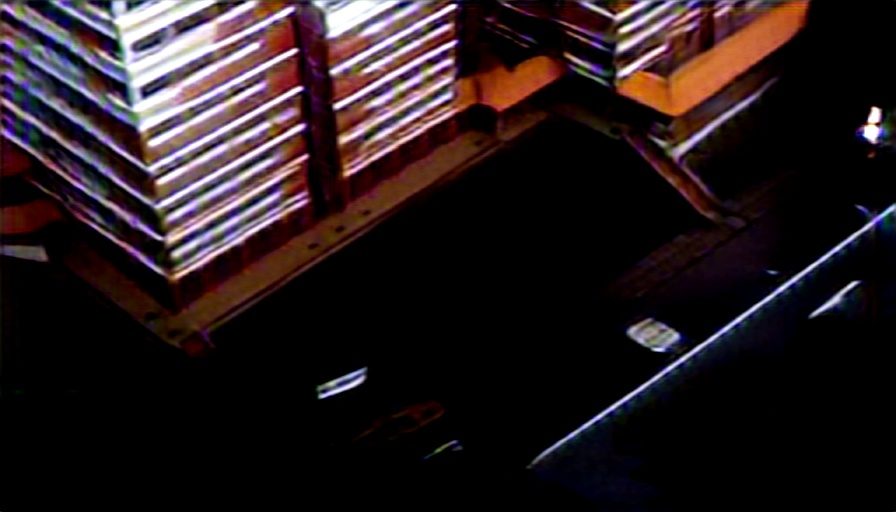}
    \includegraphics[width=0.12\textwidth]{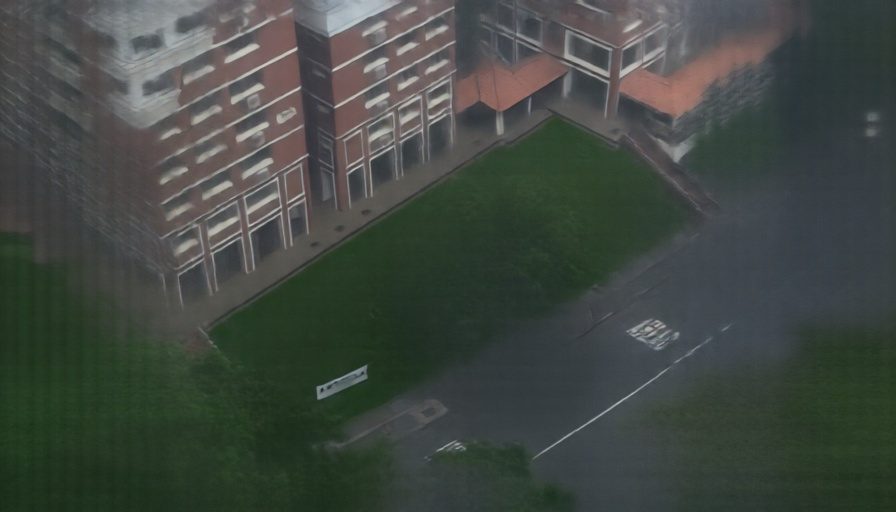}
    \includegraphics[width=0.12\textwidth]{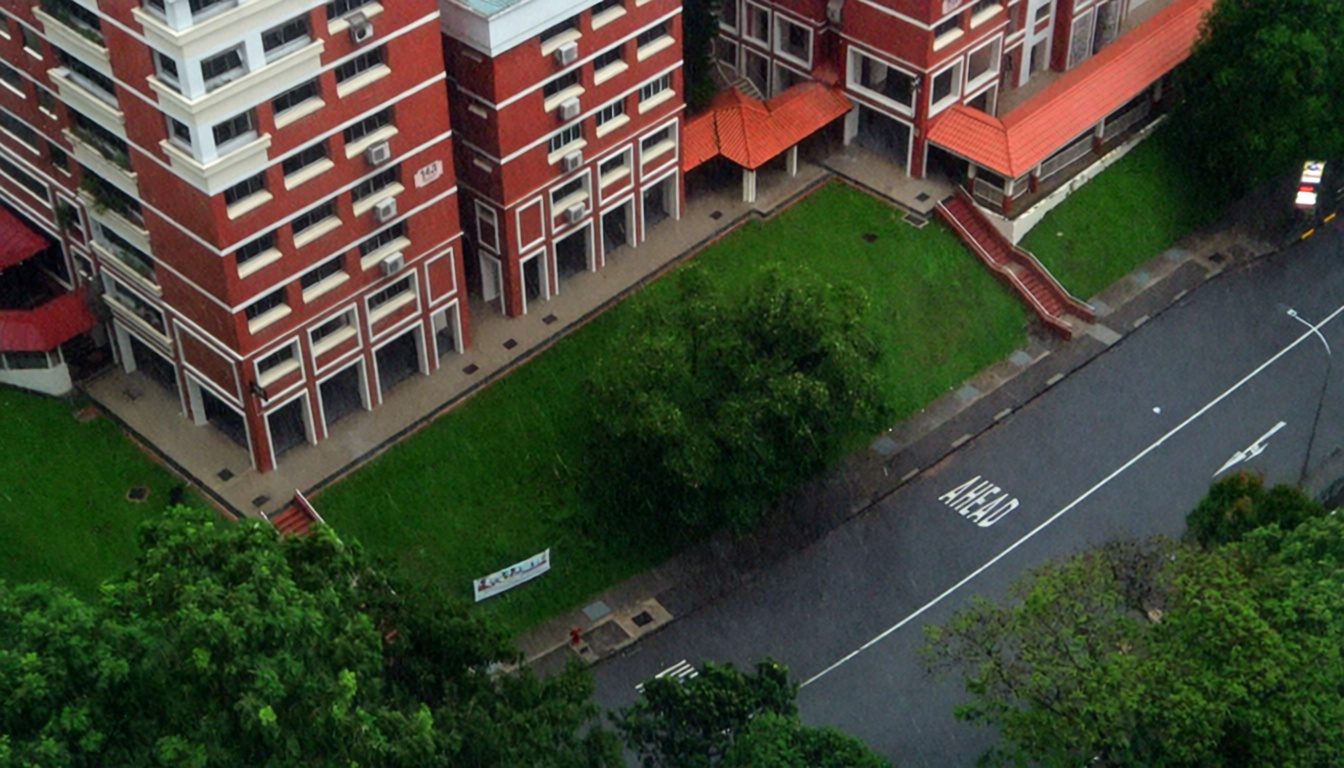}
    \includegraphics[width=0.12\textwidth]{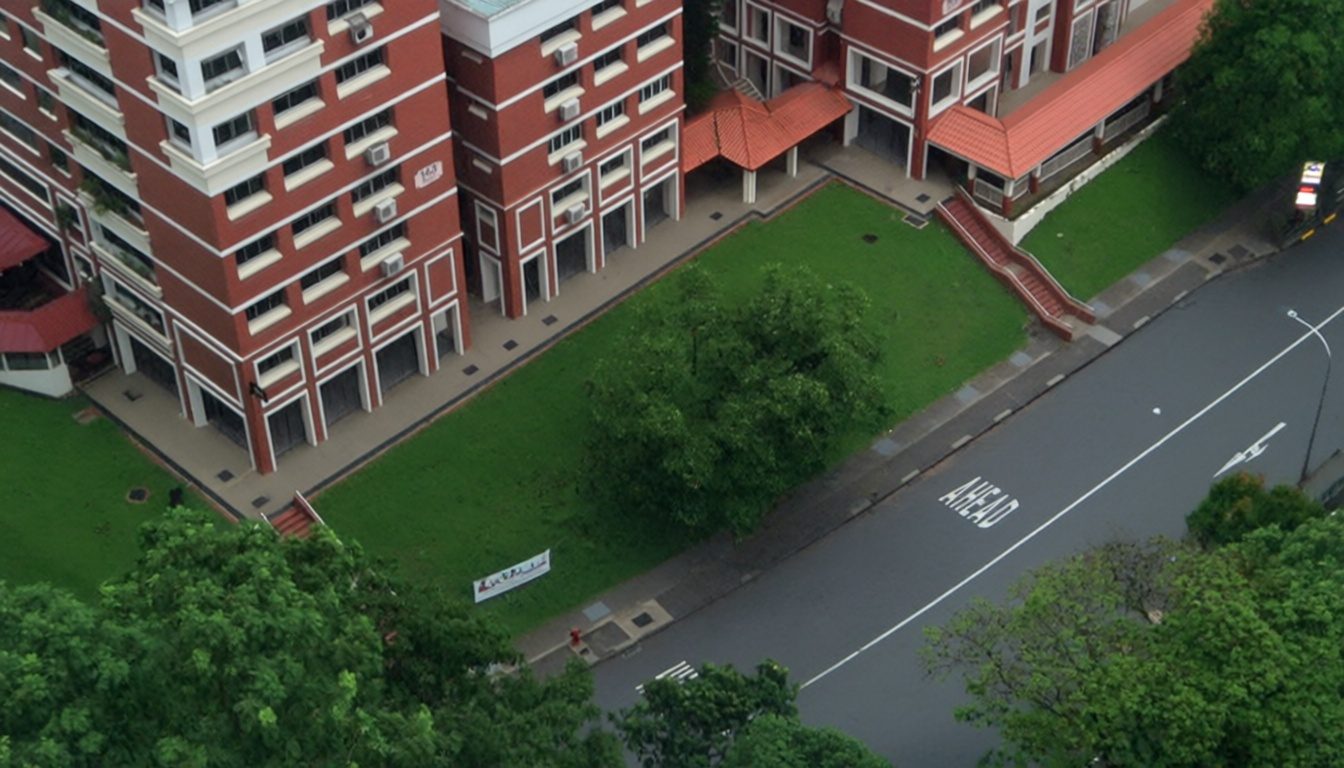}    

    \caption{Additional qualitative comparisons on challenging real-world data}
    \label{fig:appendix_visual_real}
\end{figure*}

Figure~\ref{fig:appendix_visual_synt} and  Figure~\ref{fig:appendix_visual_real}  present additional qualitative comparisons on both
synthetic and real-world images involving extreme fog, heavy noise, dense rain streaks, and strong blur that are underrepresented in
standard benchmarks. Across these cases, our proposed R\&R consistently produces more stable geometric
structures and improved color consistency, while baseline methods often exhibit structure collapse, excessive smoothing, or hallucinate details under severe degradations.

\end{document}